\documentclass{arxivtmpl}

\def\1{\bm{1}}

\DeclareMathAlphabet{\mathsfit}{\encodingdefault}{\sfdefault}{m}{sl}
\SetMathAlphabet{\mathsfit}{bold}{\encodingdefault}{\sfdefault}{bx}{n}

\tcbuselibrary{skins,breakable}
\newtcolorbox{findingbox}{enhanced,breakable,frame hidden,colback=gred!6,borderline west={3pt}{0pt}{gred},sharp corners,boxsep=0pt,left=12pt,right=10pt,top=8pt,bottom=8pt,before skip=12pt,after skip=12pt}
\newtcolorbox{notebox}{enhanced,breakable,frame hidden,colback=gblue!6,borderline west={3pt}{0pt}{gblue},sharp corners,boxsep=0pt,left=12pt,right=10pt,top=8pt,bottom=8pt,before skip=12pt,after skip=12pt}
\newtcolorbox{tipbox}{enhanced,breakable,frame hidden,colback=ggreen!6,borderline west={3pt}{0pt}{ggreen},sharp corners,boxsep=0pt,left=12pt,right=10pt,top=8pt,bottom=8pt,before skip=12pt,after skip=12pt}
\newtcolorbox{warnbox}{enhanced,breakable,frame hidden,colback=gyellow!6,borderline west={3pt}{0pt}{gyellow},sharp corners,boxsep=0pt,left=12pt,right=10pt,top=8pt,bottom=8pt,before skip=12pt,after skip=12pt}

\tcbuselibrary{listings}
\newcommand{\boxfont}{\fontsize{9}{11}\selectfont}
\newtcblisting{promptbox}[1][]{%
  enhanced, listing only, breakable, frame hidden, width=\linewidth,
  colback=gblue!6, borderline west={3pt}{0pt}{gblue},
  sharp corners, boxsep=0pt,
  left=12pt, right=10pt, top=8pt, bottom=8pt, before skip=12pt, after skip=12pt,
  before upper={\def\promptlabel{#1}\ifx\promptlabel\empty\else{\sffamily\bfseries\color{gblue}#1}\par\smallskip\fi},   %
  listing options={basicstyle=\ttfamily\boxfont, escapeinside={(*}{*)}},
}

\newtcolorbox{transcriptbox}{%
  enhanced, breakable, frame hidden,
  colback=ggreen!6, borderline west={3pt}{0pt}{ggreen},
  sharp corners, boxsep=0pt, fontupper=\ttfamily\boxfont, halign upper=flush left,   %
  left=12pt, right=10pt, top=6pt, bottom=8pt, before skip=12pt, after skip=12pt,
}

\newcommand{\teaserfigure}[2]{%
  \begingroup\centering\captionsetup{type=figure}
  #1\par\vspace{0.4em}\caption{#2}%
  \par\endgroup\vspace{1em}
}

\hypersetup{
    colorlinks=true,
    linkcolor=linkcol,
    citecolor=linkcol,
    urlcolor=linkcol,
    pdfborderstyle={},
    pdfborder={0 0 0}
}
\renewcommand\Affilfont{\sffamily\mdseries}

\makeatletter
\newcommand\cardaffil[2]{\@namedef{card@inst@#1}{#2}}
\newcount\card@n \newcount\card@cols \newdimen\card@wd
\newcommand\carda[3]{%
  \parbox[t]{\card@wd}{\raggedright\rightskip=6pt plus 1fil\sffamily
    {\fontsize{10}{12}\selectfont\strut#3\ifx\relax#2\relax\else\smash{\textsuperscript{\sffamily\color{ggrey}#2}}\fi\par}%
    {\fontsize{8}{10}\selectfont\color{ggrey}\strut\ifx\relax#1\relax\else\@nameuse{card@inst@#1}\fi\par}}%
  \global\advance\card@n 1
  \ifnum\card@n=\card@cols \global\card@n=0 \par\vspace{5pt}\noindent\fi}
\newcommand\authorcards[2]{\par{\card@cols=#1 \card@wd=\dimexpr\linewidth/#1\relax \global\card@n=0
  \setlength\parskip{0pt}\setlength\lineskip{0pt}\setlength\lineskiplimit{0pt}\setlength\parindent{0pt}\noindent#2\par}}
\makeatother

\colorlet{perfhue}{gblue}
\colorlet{perf0}{white}
\colorlet{perf10}{perfhue!12}
\colorlet{perf20}{perfhue!22}
\colorlet{perf30}{perfhue!33}
\colorlet{perf40}{perfhue!44}
\colorlet{perf50}{perfhue!55}
\colorlet{perf60}{perfhue!66}
\colorlet{perf70}{perfhue!77}
\colorlet{perf80}{perfhue!88}

\newcommand{\RN}[1]{%
	\textup{\lowercase\expandafter{\it \romannumeral#1}}%
}

\newcommand{\cmark}{\ding{51}}%
\newcommand{\xmark}{\ding{55}}%

\SetKwInput{KwInput}{Input}
\SetKwInput{KwOutput}{Output}
\DontPrintSemicolon

\SetKwComment{Comment}{\color{ggreen}\# }{}

\SetKwProg{Function}{def}{:}{}

\SetKwProg{For}{for}{:}{}
\SetKwProg{If}{if}{:}{}

\usepackage{booktabs}
\usepackage{multirow}
\usepackage{array}
\usepackage{graphicx}

\usepackage{capt-of}

\usepackage{array,longtable}
\usepackage{placeins}
\usepackage{capt-of}
\usepackage{xurl}
\IfFileExists{fontawesome5.sty}{\usepackage{fontawesome5}}{\providecommand\faGlobe{}\providecommand\faGithub{}}
\usetikzlibrary{arrows.meta,positioning,calc,fit}

\definecolor{cmarkgreen}{RGB}{34,139,34}
\definecolor{xmarkred}{RGB}{200,30,30}
\renewcommand{\cmark}{\textcolor{cmarkgreen}{\ding{51}}}
\renewcommand{\xmark}{\textcolor{xmarkred}{\ding{55}}}

\newcolumntype{Y}{>{\raggedright\arraybackslash}X}
\newcommand{\benchmark}{\text{GameSWE}}

\newcommand{\UncGroups}{60}
\newcommand{\UncB}{10{,}000}
\newcommand{\UncSpearmanMean}{0.81}
\newcommand{\UncSpearmanMeanCI}{[0.71, 0.82]}

\newcommand{\UncPairsSignificant}{16}
\newcommand{\UncPairsTotal}{21}

\newcommand{\ntasks}{150}
\newcommand{\nconstruct}{50}
\newcommand{\nrepair}{100}
\newcommand{\nhosts}{12}

\let\iclrabstract\abstract
\renewcommand{\abstract}{\iclrabstract
{\sffamily\noindent
\begin{minipage}[t]{0.5\linewidth}{\small\bfseries\faGlobe\enspace Website}\par\vspace{1pt}{\small\href{https://touchstone-ai-lab.github.io/bench/game-swe}{\textcolor{gblue}{touchstone-ai-lab.github.io/bench/game-swe}}}\end{minipage}%
\begin{minipage}[t]{0.5\linewidth}{\small\bfseries\faGithub\enspace Code}\par\vspace{1pt}{\small\href{https://github.com/Touchstone-AI-Lab/game-swe}{\textcolor{gblue}{github.com/Touchstone-AI-Lab/game-swe}}}\end{minipage}\par}
\vspace{0.9\baselineskip}
{\sffamily\fontseries{eb}\normalsize\selectfont Abstract}\par\vspace{0.15\baselineskip}\small}

\title{GameSWE: Evaluating Coding Agents in Live Game Runtimes}

\cardaffil{1}{University of Michigan}
\cardaffil{2}{Northwestern University}
\cardaffil{3}{Princeton University}
\cardaffil{4}{UIUC}
\cardaffil{5}{OpenAI}
\cardaffil{6}{Carnegie Mellon University}
\cardaffil{7}{UCLA}
\cardaffil{8}{MIT}
\cardaffil{9}{Georgia Institute of Technology}
\cardaffil{10}{University of Pennsylvania}
\cardaffil{11}{Johns Hopkins University}
\cardaffil{12}{Northern Arizona University}
\cardaffil{13}{The Ohio State University}
\cardaffil{14}{UT Austin}
\cardaffil{15}{UC Berkeley}
\cardaffil{16}{Northeastern University}
\cardaffil{17}{UCSB}
\cardaffil{18}{UW--Madison}
\cardaffil{19}{Cornell University}
\cardaffil{20}{NTU}
\cardaffil{21}{Touchstone AI}
\cardaffil{22}{Stanford University}
\cardaffil{23}{Boston University}
\cardaffil{24}{University of Washington}
\cardaffil{25}{FutureMLS}
\cardaffil{1+2}{UMich, Touchstone AI}
\cardaffil{2+3}{UPenn, Touchstone AI}
\newcommand\equalauthors{%
  \carda{1+2}{*}{Hui~Shen}%
  \carda{2}{*}{Zhihan~Liu}%
  \carda{21}{*}{Xingzhi~Li}%
  \carda{1}{*}{Sihan~Xu}%
}
\newcommand\otherauthors{%
  \carda{2}{}{Yifei~Zuo}%
  \carda{3}{}{Wenhao~Chai}%
  \carda{4}{}{Jisen~Li}%
  \carda{5}{}{Zhuoran~Yin}%
  \carda{6}{}{Changyi~Yang}%
  \carda{7}{}{Chenyang~Zhao}%
   \carda{15}{}{Wenjie~Qu}%
  \carda{8}{}{Xianbang~Wang}%
  \carda{9}{}{Xuesong~Ye}%
  \carda{10}{}{Yunta~Hsieh}%
  \carda{11}{}{Jiaxuan~Luo}%
  \carda{21}{}{Zhuhan~Bao}%
  \carda{9}{}{Zhengshen~He}%
  \carda{8}{}{Yunge~Wen}%
  \carda{21}{}{Jianing~Ma}%
  \carda{21}{}{Guobin~Zhang}%
  \carda{13}{}{Xin~Wang}%
  \carda{13}{}{Zhongwei~Wan}%
  \carda{6}{}{Hangrui~Cao}%
  \carda{4}{}{Yuqing~Jian}%
  \carda{22}{}{Shatong~Zhu}%
  \carda{8}{}{Yifeng~Liu}%
  \carda{1}{}{Xueshen~Liu}%
  \carda{1}{}{Minkyoung~Cho}%
  \carda{1}{}{Boyuan~Zheng}%
  \carda{1}{}{Hei~Ting~(Una)~Chan}%
  \carda{1}{}{Zesen~Zhao}%
  \carda{13}{}{Jingxuan~Zhang}%
  \carda{}{}{Hengyuan~Zhang}%
  \carda{}{}{Zunhai~Su}%
  \carda{}{}{Jing~Xiong}%
  \carda{}{}{Chaofan~Tao}%
  \carda{14}{}{Ziheng~Zhang}%
  \carda{9}{}{Kejing~Xia}%
  \carda{2}{}{Xing~Wei}%
  \carda{1}{}{Yuchen~Jiang}%
  \carda{21}{}{Rui~Hu}%
  \carda{15}{}{Xutao~Ma}%
  \carda{16}{}{Zhen~Qi}%
  \carda{17}{}{Zhen~Zhang}%
  \carda{23}{}{Ping~Zhang}%
  \carda{9}{}{Yichi~Chen}%
  \carda{21}{}{Simo~Zhang}%
  \carda{21}{}{Mengyao~Wen}%
  \carda{21}{}{Xin~Xu}%
  \carda{18}{}{Jiawei~Lu}%
  \carda{1}{}{Yifan~Zhang}%
  \carda{18}{}{Deyang~Luorong}%
  \carda{10}{}{Ziyu~Liu}%
  \carda{14}{}{Jirong~Yang}%
  \carda{24}{}{Ruoke~Zhang}%
  \carda{1}{}{Chenwei~Wu}%
  \carda{}{}{Minchen~Luo}%
  \carda{}{}{Yuanshuo~Lu}%
  \carda{22}{}{Xiaoyin~Qu}%
  \carda{19}{}{Qiao~Xiao}%
  \carda{21}{}{Xinyuan~Dai}%
  \carda{21}{}{Sangyu~Pan}%
  \carda{1}{}{Muhammad~Khaquan}%
  \carda{6}{}{Haizhong~Zheng}%
  \carda{8}{}{Rui~Sun}%
  \carda{25}{}{Zhongzhu~Zhou}%
  \carda{21}{}{Yuwei~Fang}%
  \carda{8}{}{Tianfu~Fu}%
  \carda{13}{}{Kai~Zhang}%
  \carda{}{}{Ngai~Wong}%
  \carda{13}{}{Mi~Zhang}%
  \carda{2}{}{Zhaoran~Wang}%
}

\hypersetup{pdftitle={GameSWE: Evaluating Coding Agents in Live Game Runtimes},pdfauthor={Hui Shen et al.}}

\graphicspath{{./}{../}}

\begin{document}

\makeatletter\gdef\@author{}\makeatother
\maketitle

\thispagestyle{firstpagestyle}

\begin{abstract}
Coding agents are increasingly evaluated on repository-level tasks, where a change is judged by the repository's own tests. Yet much software runs inside a live host program, where correctness depends on runtime state, lifecycle events, and third-party components loaded alongside it. We introduce \benchmark{}, a benchmark of \ntasks{} repair and construction tasks built from open-source mods for \nhosts{} popular games across five programming languages. Each submission is compiled, loaded into the real game or its dedicated server together with the pinned third-party mods it depends on, and judged by fail-to-pass and pass-to-pass tests executed in the live runtime, without mocks or LLM judges. Across seven frontier and open-weight models, the best agent resolves 69.0\% of tasks while three models remain at or below 25.0\%, and even the best agent's per-game rate ranges from 40.0\% to 93.3\%. Our analysis shows that 70.0\% of unresolved attempts end with the agent submitting its work as complete, and in at least 31.0\% the agent reports that its own checks pass, yet the submission fails in the runtime, with failures concentrated at the boundary between code and host: interfaces the ecosystem does not provide, host conventions beyond the code, and logic not connected to the game's lifecycle. \benchmark{} provides a reproducible testbed for coding agents whose code must work where it runs.

\end{abstract}

\def\hasteaser{}%
\teaserfigure{\includegraphics[width=\linewidth]{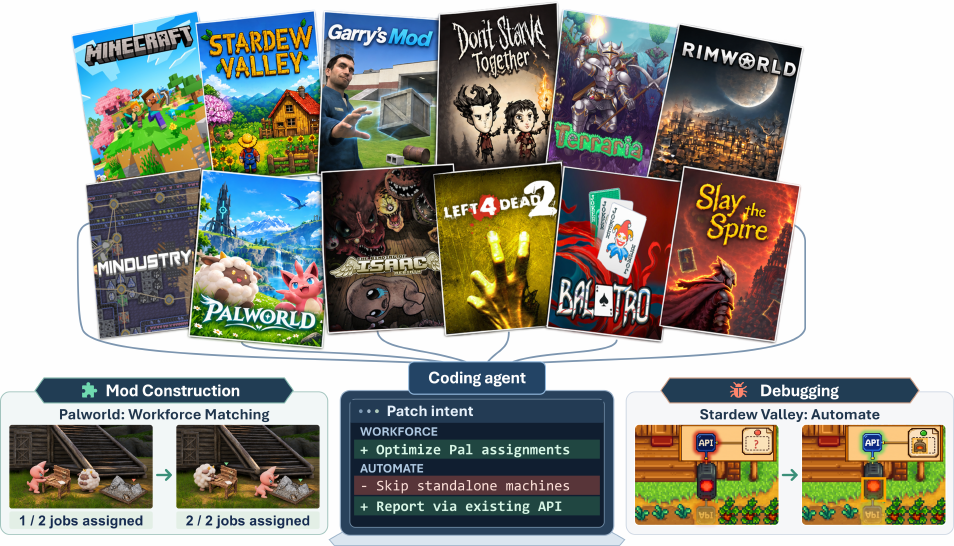}}{\textbf{GameSWE.} Mod repair and construction tasks in the benchmark. \label{fig:teaser}}

\clearpage
\section*{Author list}
{\setlength{\parindent}{0pt}\raggedright\sffamily
\vspace{0.5em}\authorcards{4}{\equalauthors}\vspace{1em}
\authorcards{4}{\otherauthors}\vspace{0.8em}
{\fontsize{8.5}{11.5}\selectfont\color{ggrey}Zhuoran Yin contributed to this work independently and in personal capacity, not as part of his role at OpenAI.\par}}
\vspace{1.2em}

\clearpage
\section{Introduction}
\label{sec:intro}

Coding agents have advanced at a remarkable pace, progressing from completing standalone functions~\citep{humaneval} to resolving real-world GitHub issues~\citep{swebench, sweagent, openhands}, and they are now part of everyday development workflows~\citep{agenticmuch, cliagents}. Benchmarks in the SWE-bench family measure this progress in a common way: the agent edits a repository, and the edit is judged by tests from that same repository, run in an isolated container~\citep{swebench, swebenchmultimodal, swepolybench, swebenchpro}. This protocol has made agent evaluation realistic and reproducible, but it judges a change only by what the repository's tests exercise.

Much software, however, does not run on its own. Plugins, extensions, and mods run inside a \emph{live host program} that their authors do not own, and whether they work depends on what happens in that program: whether the host loads them, which interfaces and globals it actually exposes, how its state evolves through loading, saving, death and respawn, restarts, and reloads, and which third-party components run beside them. Repository tests and mocks capture these conditions only as far as their authors anticipate them. A change can therefore look correct in its repository and still fail where it runs.

\begin{figure}[htbp]
\centering
\includegraphics[width=\linewidth]{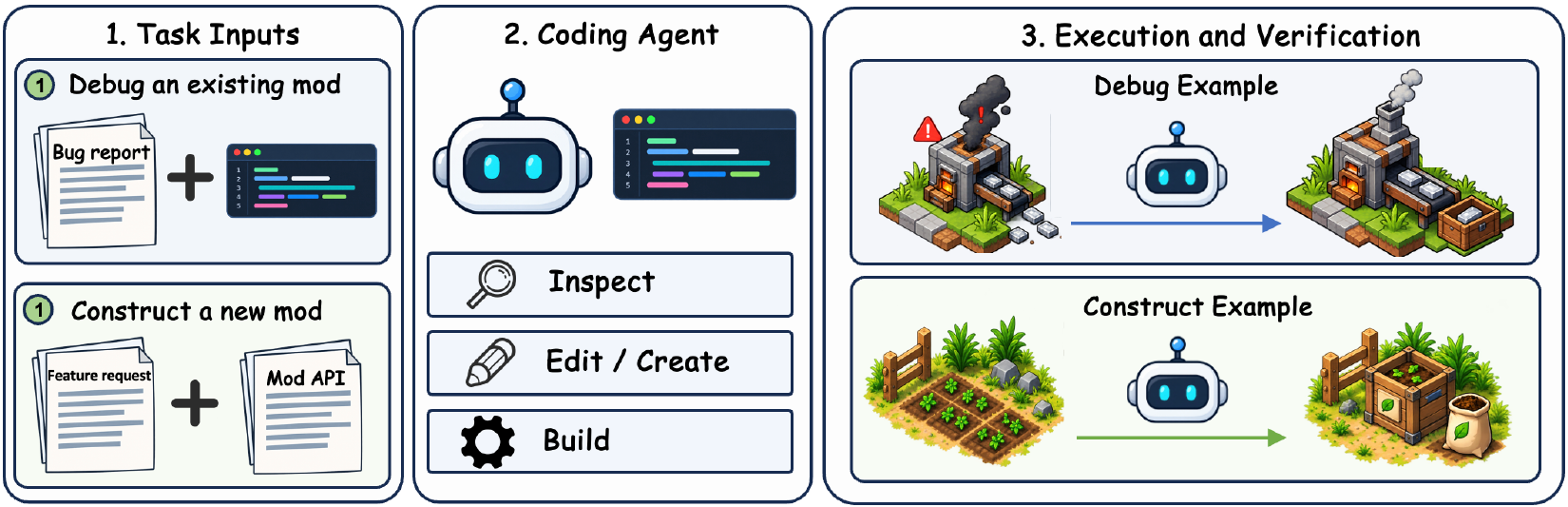}
\caption{\textbf{Overview of \benchmark{}.} Repair tasks provide bug reports and an existing mod; construction tasks provide a feature request and the game interfaces it builds on. The agent inspects and edits the code without access to the game. The live-runtime harness then builds the submission, loads it into the real game or dedicated server together with the pinned third-party mods it depends on, and runs fail-to-pass and pass-to-pass tests inside the running game (Section~\ref{sec:verification}). The scenes illustrate the two task types and are not recorded agent outputs.}
\label{fig:overview}
\end{figure}

Game modding provides a natural testbed for this gap. Mods extend released software through host interfaces, run alongside independently developed components, and can be evaluated directly in the running game. In-game fixtures exercise stateful events such as save/load, respawn, and server restart, making the host runtime an authoritative oracle for both functionality and ecosystem compatibility.

\begin{figure}[htbp]
\centering
\includegraphics[width=\linewidth]{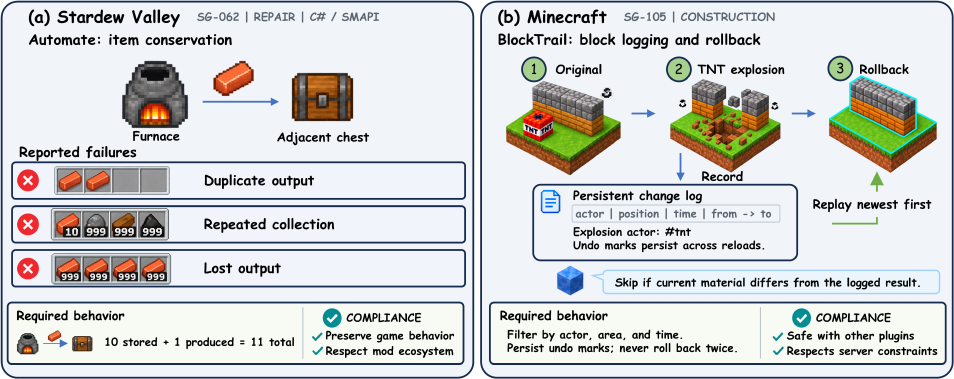}
\caption{\textbf{Two representative tasks from \benchmark{}.} (a) A Stardew Valley repair task: the Automate mod must stop duplicating, re-collecting, and losing machine output while keeping its existing behavior. (b) A Minecraft construction task: a new Paper plugin must log block changes and roll them back, with undo marks that persist across reloads.}
\label{fig:example}
\end{figure}

We introduce \benchmark{}, a benchmark of \ntasks{} tasks built from open-source mods for \nhosts{} popular games, including Minecraft, Stardew Valley, and Garry's Mod, with code in Lua, Java, C\#, SourcePawn, and Squirrel (Figure~\ref{fig:overview}). In the \nrepair{} \emph{repair} tasks, the agent receives an existing mod with inserted bugs and the corresponding bug reports; in the \nconstruct{} \emph{construction} tasks, it receives a feature request and the game interfaces to build on. Following SWE-bench, each task is judged by \emph{fail-to-pass} (F2P) tests for the requested fix or feature and, where the initial code already provides working functionality, by \emph{pass-to-pass} (P2P) tests for that functionality. Unlike repository-level benchmarks, these tests run inside the live game, with the pinned third-party mods that the target mod depends on loaded alongside the submission. Figure~\ref{fig:example} shows two representative tasks.

We evaluate seven frontier and open-weight models as coding agents: Claude Opus 5.5~\citep{claudeopus55}, GPT-6 Astra~\citep{gpt6astra}, Kimi K3.1 Pro~\citep{kimik31pro}, DeepSeek-V4.1-Flash~\citep{deepseekv41flash}, MiniMax M3.1 Flash~\citep{minimaxm31flash}, GLM 5.3 Flash~\citep{glm53flash}, and Qwen 3.8 27B~\citep{qwen38}. Our main findings are:
\begin{itemize}
    \item \textbf{Live-runtime tasks are far from solved.} The best agent, Claude Opus 5.5, resolves 69.0\% of tasks, and three of the seven models stay at or below 25.0\%. Every model resolves fewer construction tasks than repairs.
    \item \textbf{Passing the agent's own checks does not mean working in the game.} 70.0\% of the unresolved attempts end with the agent submitting its work as complete, and in at least 31.0\% the agent reports that its own checks or tests pass, yet the submission fails once loaded into the game.
    \item \textbf{Failures concentrate at the boundary between code and host.} Unresolved submissions most often use ecosystem interfaces that do not exist, violate the host's packaging conventions, or implement the requested logic without connecting it to the game's full lifecycle. These failures surface only when the game executes the code: 427 of the 617 unresolved attempts are loaded into the game and fail there, at load time or in its tests.
\end{itemize}

\clearpage
\setcounter{tocdepth}{2}
\begingroup
\hypersetup{linkcolor=black}
\tableofcontents
\endgroup
\clearpage

\section{Related Work}
\label{sec:related_work}

\paragraph{Benchmarks for coding agents.}
Evaluation of code generation has moved from function-level synthesis in HumanEval~\citep{humaneval} to repository-level issue resolution in SWE-bench~\citep{swebench}, which judges a patch by fail-to-pass tests for the requested change and pass-to-pass tests for previously working behavior. Later benchmarks extend this protocol along several axes: SWE-bench Multimodal adds visual JavaScript domains~\citep{swebenchmultimodal}, SWE-PolyBench spans several languages and task types~\citep{swepolybench}, SWE-Bench Pro targets long-horizon changes in enterprise-grade repositories~\citep{swebenchpro}, FeatureBench evaluates feature development~\citep{featurebench}, and SWE-Lancer grades freelance tasks with end-to-end browser tests~\citep{swelancer}. \benchmark{} adopts the fail-to-pass and pass-to-pass distinction, but changes where the tests run. In these benchmarks, a change is judged by tests of the repository or application that the change itself belongs to. In \benchmark{}, the submission extends a released program that it does not own, and it is judged inside that program's live runtime, together with third-party components loaded alongside it.

\paragraph{Game development benchmarks.}
A growing line of work evaluates coding agents on game development. GameDevBench builds tasks from Godot tutorials and checks gameplay, graphics, and UI features in the engine~\citep{gamedevbench}, and GameEngineBench asks agents to implement gameplay systems, networking, and persistence in Unreal Engine projects~\citep{gameenginebench}. GameCraft-Bench~\citep{gamecraftbench} and WebGameBench~\citep{webgamebench} evaluate complete games by interacting with them at runtime, while GameXpert-Bench covers the generation, repair, and iterative improvement of browser games with regression checks~\citep{gamexpertbench}. In these benchmarks, the agent develops a game project itself. Closest to our work, GameForgeBench includes Minecraft Paper plugins among Godot, C++, and browser projects, and verifies some plugin tasks on a live Paper server~\citep{gameforgebench}. \benchmark{} differs in three respects (Table~\ref{tab:closest-work}): it is built entirely around extending released games and spans \nhosts{} of them, every submission is verified in the live game runtime, and the runtime includes independently developed mods, which we find in no public GameForgeBench task.

\par\addvspace{10pt} 
\noindent
\begin{minipage}{\columnwidth}
\centering
\scriptsize
\setlength{\tabcolsep}{1.5pt}
\renewcommand{\arraystretch}{1.16}

\captionof{table}{
\textbf{Comparison with related benchmarks.}
Host: extending released software;
Live: in-runtime verification;
3rd party: external components.
$\circ$ denotes partial support.
}
\label{tab:closest-work}
\label{tab:related-comparison}

\begin{tabularx}{\linewidth}{@{}
  >{\raggedright\arraybackslash}p{0.31\linewidth}
  *{3}{>{\centering\arraybackslash}m{0.085\linewidth}}
  >{\raggedright\arraybackslash}X
  @{}}

\toprule

\textbf{Benchmark}
& \textbf{Host}
& \shortstack{\textbf{Live}\\\textbf{tests}}
& \shortstack{\textbf{3rd}\\\textbf{party}}
& \textbf{Capability evaluated} \\

\midrule

SWE-bench
& \xmark & \xmark & \xmark
& Resolve repository issues \\

FeatureBench
& \xmark & \xmark & \xmark
& Implement repository features \\

GameDevBench
& \xmark & \xmark & \xmark
& Godot gameplay, graphics, UI \\

GameEngineBench
& \xmark & \xmark & \xmark
& Unreal C++ gameplay systems \\

GameCraft-Bench
& \xmark & \xmark & \xmark
& Build complete Godot games \\

GameXpert-Bench
& \xmark & \xmark & \xmark
& Generate, repair, improve games \\

GameForge
& \textcolor{orange!85!black}{$\circ$}
& \textcolor{orange!85!black}{$\circ$}
& \xmark
& Edit and repair game projects \\

\midrule

\textbf{\benchmark{} (ours)}
& \cmark & \cmark & \cmark
& \textbf{Repair and extend mods of \nhosts{} games} \\

\bottomrule
\end{tabularx}

\end{minipage}
\par\medskip

\FloatBarrier

\section{\benchmark{}}
\label{sec:benchmark}

\subsection{Task formulation}
\label{sec:formulation}

Each task in \benchmark{} asks an agent to change a mod for a released game. The agent receives a task description $d$, initial code $c_0$, and documentation of the game interfaces and tools it may use, and it produces modified code $c$. As in SWE-bench~\citep{swebench}, the submission is judged by two sets of tests: \emph{fail-to-pass} (F2P) tests $T_{\mathrm{F2P}}$, which check the requested fix or feature and fail on $c_0$, and \emph{pass-to-pass} (P2P) tests $T_{\mathrm{P2P}}$, which check functionality that already works with $c_0$. $T_{\mathrm{P2P}}$ is empty when $c_0$ provides no such functionality, as for 49 of the \nconstruct{} construction tasks and 8 of the \nrepair{} repair tasks. Unlike SWE-bench, both sets run in an environment $E$ consisting of the live game runtime and a set $M$ of pinned third-party mods, and they are withheld from the agent. \benchmark{} includes two task types (Figure~\ref{fig:overview}). In \emph{repair} tasks, $c_0$ is an existing mod with inserted bugs and $d$ consists of bug reports. In \emph{construction} tasks, $c_0$ is an empty mod project or a skeleton and $d$ is a feature request written in terms of the interfaces the game already provides.

\subsection{Live-runtime verification}
\label{sec:verification}

\paragraph{Harness.}
For each task, a live-runtime harness evaluates the submission in four stages. It (1) \emph{builds} $c$ with the game's modding toolchain, (2) \emph{loads} it into the real game or its dedicated server together with the third-party mods in $M$, (3) \emph{executes} the tests inside the running game, and (4) \emph{tears down} the session and checks that it terminates cleanly. Neither the game nor $M$ is mocked; a submission sees exactly the interfaces, globals, and load order that the shipped game provides.

\paragraph{Tests as in-game fixtures.}
Each test starts from a specified game state, performs a sequence of in-game actions, and checks the resulting state; we call the state and actions the test's \emph{fixture}. Fixtures drive the game through the same events that players and servers produce, such as loading a save, placing a machine next to a chest, killing and respawning a player, reloading a plugin's configuration, or restarting the server with saved data.

\paragraph{Outcomes and metric.}
A test \emph{passes}, \emph{fails}, or is \emph{unobserved} when execution stops before it runs, for example because the mod crashes; we keep unobserved tests separate so that a single crash does not count as many independent failures. A task is \emph{resolved} only if the submission builds and loads, every F2P and P2P test passes with $M$ loaded, and the session terminates cleanly. We report the resolution rate, and we record F2P and P2P outcomes separately to distinguish an incomplete change from one that breaks existing functionality. Each task pins the versions of the game, the target mod, the mods in $M$, and the harness, so that results are reproducible.

\subsection{Task construction and validation}
\label{sec:construction}
\label{sec:data-annotation}

\paragraph{Sources.}
All tasks are built from publicly available, open-source mods. A repair task starts from a working version of a mod: we either revert a fix from the mod's history or introduce a new defect, and the working version serves as the reference solution. We also identify latent defects in upstream implementations and construct reference fixes against explicitly specified behavioral contracts. A construction task specifies a feature through interfaces that the game or its modding ecosystem documents, and we write its reference implementation.

\paragraph{Tests.}
All tests are written by hand. Each reported bug or requested feature is covered by at least one F2P test. P2P tests cover functionality that already works with the unmodified game and initial code, including interactions with the mods in $M$; a construction task that starts from an empty project has no such baseline and contains F2P tests only.

\paragraph{Validation.}
We initially collect 250 candidate tasks, comprising 150 construction and 100 debugging tasks. A task enters \benchmark{} only after three checks in the live runtime. First, the reference solution passes all tests with $M$ loaded. Second, on the initial code, every F2P test fails and every P2P test passes. Third, a human reviewer plays through the task in the game and confirms that the description matches what happens in the game and that the tests check it. Tasks that fail any check are revised and validated again, or discarded. From the 250 candidates, we retain \nrepair{} repair and \nconstruct{} construction tasks, which constitute \benchmark{}.

\paragraph{Difficulty calibration.}
We calibrate difficulty within each task type across the 12 games. For repair tasks, annotators match the scope of reference patches and reported bugs; for construction tasks, they match the complexity of reference implementations, including code changes and host interfaces. Tasks with substantially different difficulty are revised or excluded.

\subsection{Benchmark features}
\label{sec:features}

\begin{figure}[!ht]
\centering
\includegraphics[width=\linewidth]{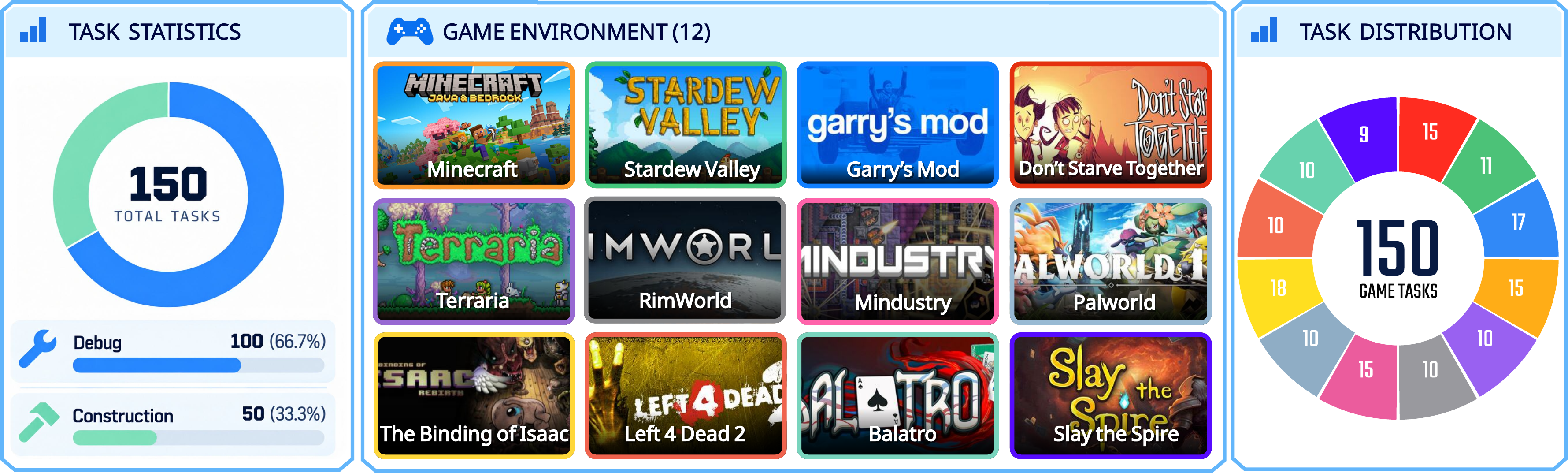}
\caption{\textbf{Task composition and game coverage.} Left: the \ntasks{} tasks comprise \nrepair{} repairs and \nconstruct{} constructions. Middle: the \nhosts{} games; equal-sized tiles do not encode task counts. Right: the number of tasks per game, from 9 to 18.}
\label{fig:inventory}
\end{figure}

\benchmark{} contains \ntasks{} tasks, \nrepair{} repairs and \nconstruct{} constructions, with 9 to 18 tasks per game (Figure~\ref{fig:inventory}). A repair task starts from a median of 8{,}918 non-blank lines of mod source, and its reference solution changes a median of 54 lines; a construction task starts from an empty project or a skeleton of at most 15 lines, and its reference adds a median of 379.5 lines. Tasks have a median of 11 F2P and 4 P2P tests. For comparison, SWE-bench tasks come from codebases of 438K lines on average, and their gold patches edit 32.8 lines in 1.7 files with 9.1 F2P tests~\citep{swebench}. The mods in \benchmark{} are much smaller than SWE-bench repositories, but the reference changes are larger, 221.4 lines in 2.9 files on average, and each task has 17.4 F2P tests on average, all of which run in the live game. Together, these tasks give \benchmark{} the following properties.

\paragraph{Stateful and lifecycle-rich.}
Fixtures exercise the game's lifecycle, not just isolated functions. In a Minecraft repair, reloading an edited configuration must register new and renamed commands, remove obsolete ones, and keep unchanged ones working. In a Garry's Mod construction, equipment profiles must survive a player's death and respawn and a server restart.

\paragraph{Shared runtime with third-party mods.}
Mods for the same game are written by different authors without coordination and often modify the same game objects~\citep{perfectgame}. Where the target mod depends on independently developed mods, the submission runs with them loaded (20 tasks), and some tasks involve them directly: in a Stardew Valley repair, the Automate mod must return missing machine reports through its existing API, because the overlay mod that consumes the API cannot reach a newly added function.

\subsection{A Living Benchmark}
\label{sec:living_benchmark}

\benchmark{} is designed as a living benchmark that continuously expands its coverage of software ecosystems. The 12 games and 150 tasks presented in this paper constitute \textbf{GameSWE v1}, rather than a fixed or exhaustive test suite. Future releases will incorporate additional games, introducing new programming languages, modding frameworks, host interfaces, and runtime behaviors. Each addition will follow the same task construction and live-runtime validation procedures described above. To preserve reproducibility, we maintain versioned benchmark snapshots with pinned game, mod, test, and harness versions. Results are reported against specific versions, as expanding game coverage can change the aggregate score. This design allows \benchmark{} to track coding-agent capabilities across an increasingly diverse range of real-world software ecosystems.

\section{Experiments}
\label{sec:experiments}

\subsection{Experimental setup}
\label{sec:protocol}

\paragraph{Models.}
We evaluate seven models as coding agents: Claude Opus 5.5~\citep{claudeopus55}, GPT-6 Astra~\citep{gpt6astra}, Kimi K3.1 Pro~\citep{kimik31pro}, DeepSeek-V4.1-Flash~\citep{deepseekv41flash}, MiniMax M3.1 Flash~\citep{minimaxm31flash}, GLM 5.3 Flash~\citep{glm53flash}, and Qwen 3.8 27B~\citep{qwen38}. They cover proprietary and open-weight models across a wide range of sizes. Reasoning or thinking is enabled wherever the endpoint offers it, at the high effort setting for Claude Opus 5.5, GPT-6 Astra, and DeepSeek-V4.1-Flash, and MiniMax M3.1 Flash is served in NVFP4 precision.

\paragraph{Agent setup.}
All models run in the same agent scaffold, with text and shell tools. Network access is restricted to the model endpoint. The agent can inspect and edit the task's code and run shell commands in its workspace, but it has no access to the game runtime, the third-party mods, or the tests. After the agent finishes or reaches a limit, the harness evaluates its final code as described in Section~\ref{sec:verification}. This setting deliberately measures how well an agent can produce code that works in the live runtime from the code and the task description alone. 

\paragraph{Metrics.}
We report the resolution rate defined in Section~\ref{sec:verification}. For each game, we compute the rate over its tasks; the overall score averages the per-game rates, so that each game contributes equally regardless of its number of tasks.

\subsection{Main results}
\label{sec:results}

Table~\ref{tab:main_results} reports the resolution rate of each model in each game.

\begin{table*}[t]
\centering
\scriptsize
\setlength{\tabcolsep}{2.2pt}
\renewcommand{\arraystretch}{1.15}

\newcommand{\osmodelicon}[2][1.0em]{%
  \IfFileExists{figures/icons/#2}{%
    \raisebox{-0.18\height}{%
      \includegraphics[height=#1]{figures/icons/#2}%
    }\,%
  }{}%
}

\caption{
\textbf{Main results across the 12 games in \benchmark{}.}
Each game has three columns: the resolution rate (\%) over all of its tasks (\emph{All}), over its repair tasks (\emph{Rep.}), and over its construction tasks (\emph{Con.}); -- marks a game without tasks of that kind.
A task is resolved when its submission builds, loads, and passes all fail-to-pass and pass-to-pass tests in the live runtime.
In the \textbf{Total} columns, \emph{All} averages the 12 per-game rates, weighting each game equally, and \emph{Rep.} and \emph{Con.} pool the \nrepair{} repair and \nconstruct{} construction tasks; their shading runs from red (lowest) to green (highest).
Models are sorted by Total, and each model makes one attempt per task.
Among models, the best result in each column is in \textbf{bold} and the second best is \underline{underlined}.
}
\label{tab:main_results}

\ifdefined\mainresultsbox\else\newsavebox{\mainresultsbox}\fi
\sbox{\mainresultsbox}{%
\begin{tabular}{@{}lccccccccccccccccccccc}
\toprule
\multirow{3}{*}{\textbf{Model / Agent}} & \multicolumn{21}{c}{\textbf{Resolution Rate (\%) $\uparrow$}} \\
\cmidrule(l){2-22}
 & \multicolumn{3}{c}{\textbf{\begin{tabular}[b]{@{}c@{}}Mindustry\end{tabular}}} & \multicolumn{3}{c}{\textbf{\begin{tabular}[b]{@{}c@{}}Palworld\end{tabular}}} & \multicolumn{3}{c}{\textbf{\begin{tabular}[b]{@{}c@{}}The Binding\\of Isaac\end{tabular}}} & \multicolumn{3}{c}{\textbf{\begin{tabular}[b]{@{}c@{}}Left 4\\Dead 2\end{tabular}}} & \multicolumn{3}{c}{\textbf{\begin{tabular}[b]{@{}c@{}}Balatro\end{tabular}}} & \multicolumn{3}{c}{\textbf{\begin{tabular}[b]{@{}c@{}}Slay the\\Spire\end{tabular}}} & \multicolumn{3}{c}{\textbf{\begin{tabular}[b]{@{}c@{}}Total\end{tabular}}} \\
\cmidrule(lr){2-4} \cmidrule(lr){5-7} \cmidrule(lr){8-10} \cmidrule(lr){11-13} \cmidrule(lr){14-16} \cmidrule(lr){17-19} \cmidrule(lr){20-22}
 & All & Rep. & Con. & All & Rep. & Con. & All & Rep. & Con. & All & Rep. & Con. & All & Rep. & Con. & All & Rep. & Con. & All & Rep. & Con. \\
\midrule
\osmodelicon{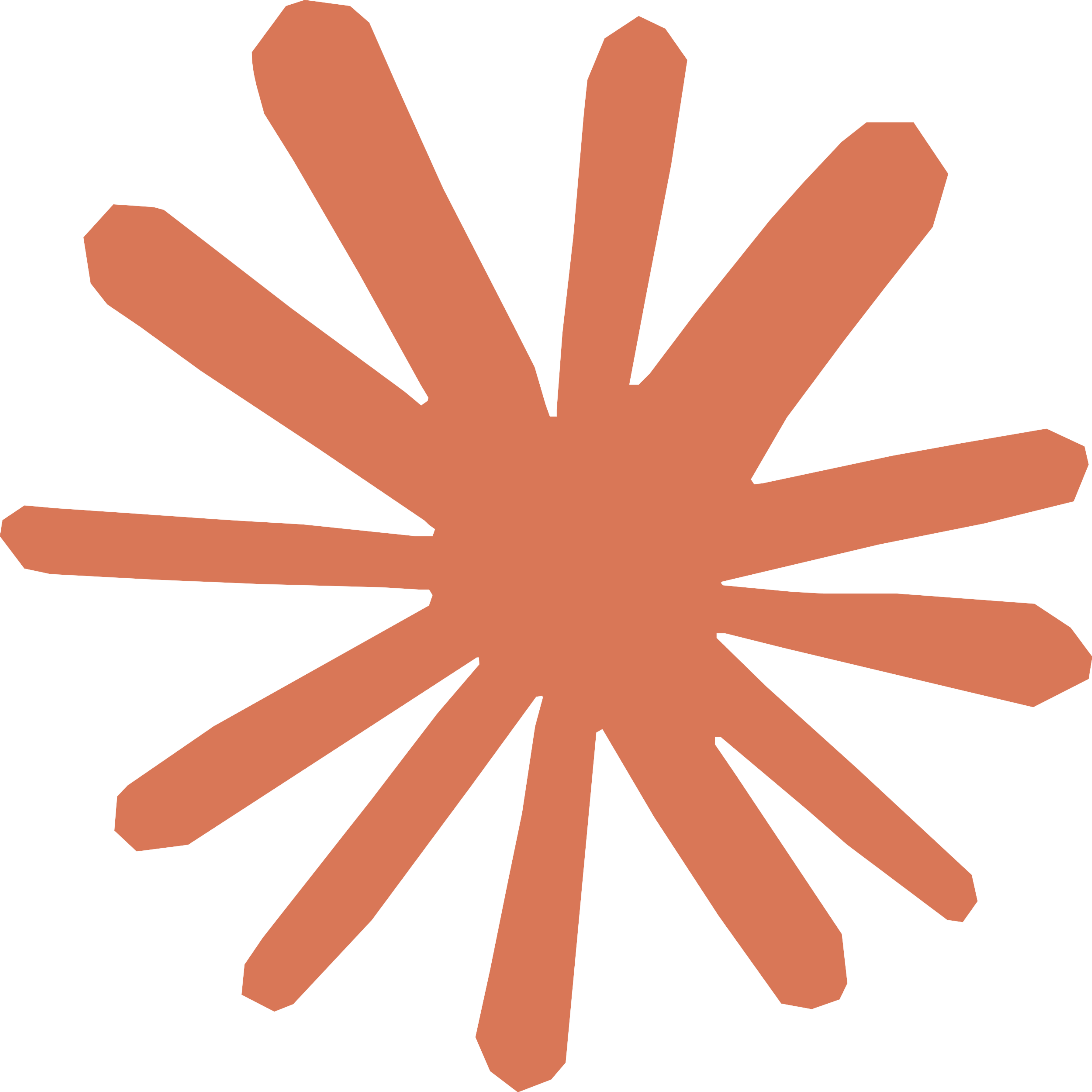}Claude Opus 5.5 & \textbf{73.3} & \textbf{80.0} & \textbf{70.0} & \textbf{40.0} & \textcolor{gray}{--} & \textbf{40.0} & \textbf{61.1} & \textbf{70.0} & \textbf{50.0} & \textbf{70.0} & \textbf{70.0} & \textcolor{gray}{--} & \underline{90.0} & \underline{90.0} & \textcolor{gray}{--} & \textbf{77.8} & \textbf{77.8} & \textcolor{gray}{--} & \cellcolor[HTML]{D9EFDF}\textbf{69.0} & \cellcolor[HTML]{D9EFDF}\textbf{75.0} & \cellcolor[HTML]{D9EFDF}\textbf{56.0} \\
\osmodelicon{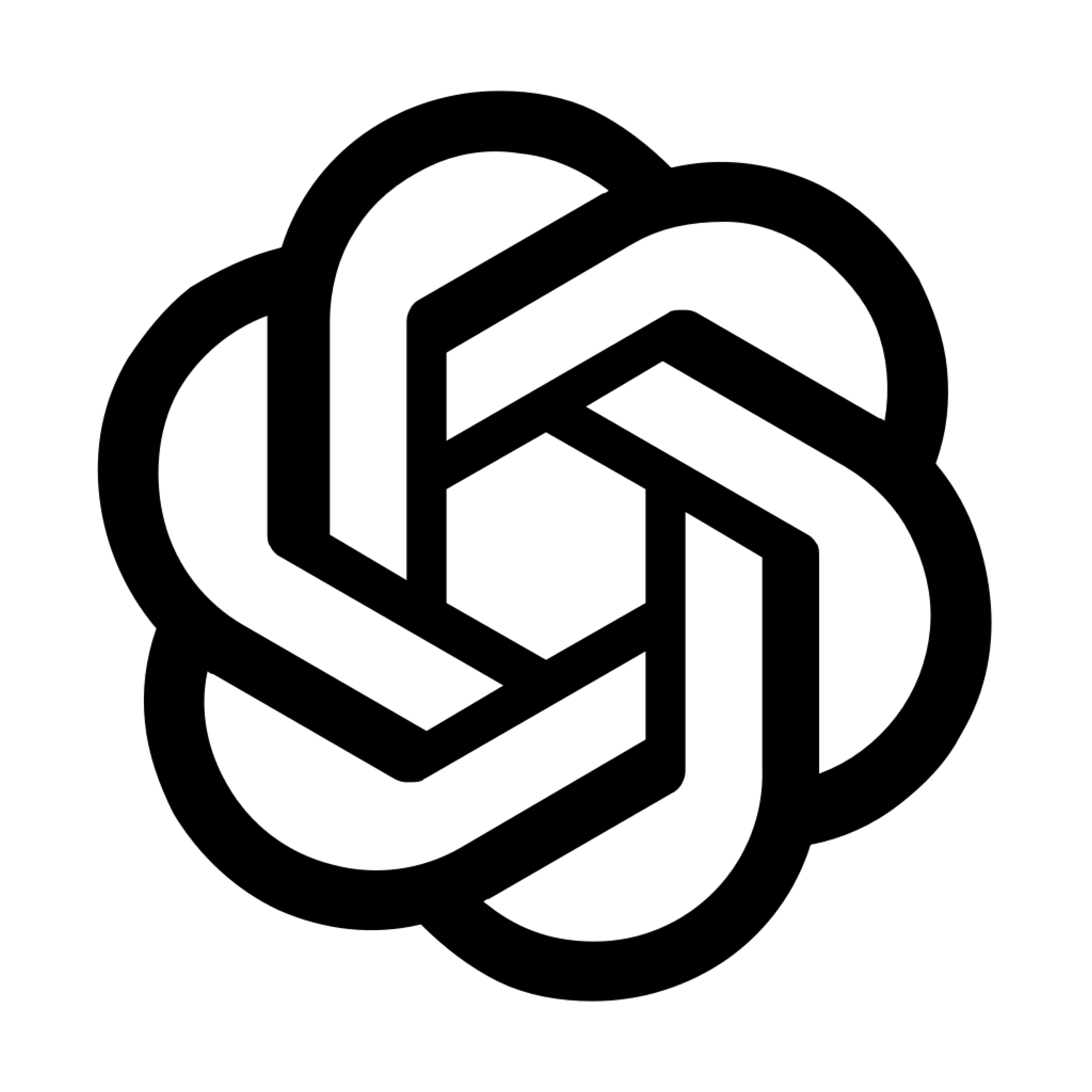}GPT-6 Astra & 60.0 & \textbf{80.0} & 50.0 & 0.0 & \textcolor{gray}{--} & 0.0 & \underline{55.6} & \underline{60.0} & \textbf{50.0} & \textbf{70.0} & \textbf{70.0} & \textcolor{gray}{--} & \textbf{100.0} & \textbf{100.0} & \textcolor{gray}{--} & 55.6 & 55.6 & \textcolor{gray}{--} & \cellcolor[HTML]{E5F0D9}\underline{61.0} & \cellcolor[HTML]{DFEFDC}\underline{71.0} & \cellcolor[HTML]{EAF0D7}\underline{44.0} \\
\osmodelicon{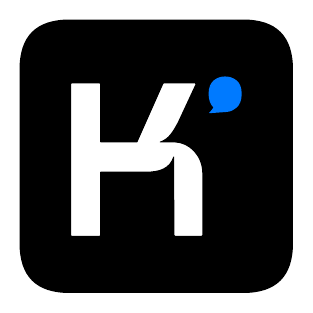}Kimi K3.1 Pro & 60.0 & \underline{60.0} & \underline{60.0} & 0.0 & \textcolor{gray}{--} & 0.0 & 22.2 & 30.0 & 12.5 & \underline{50.0} & \underline{50.0} & \textcolor{gray}{--} & 80.0 & 80.0 & \textcolor{gray}{--} & 44.4 & 44.4 & \textcolor{gray}{--} & \cellcolor[HTML]{F6F1D1}49.5 & \cellcolor[HTML]{EEF1D5}61.0 & \cellcolor[HTML]{FEECC9}26.0 \\
\osmodelicon{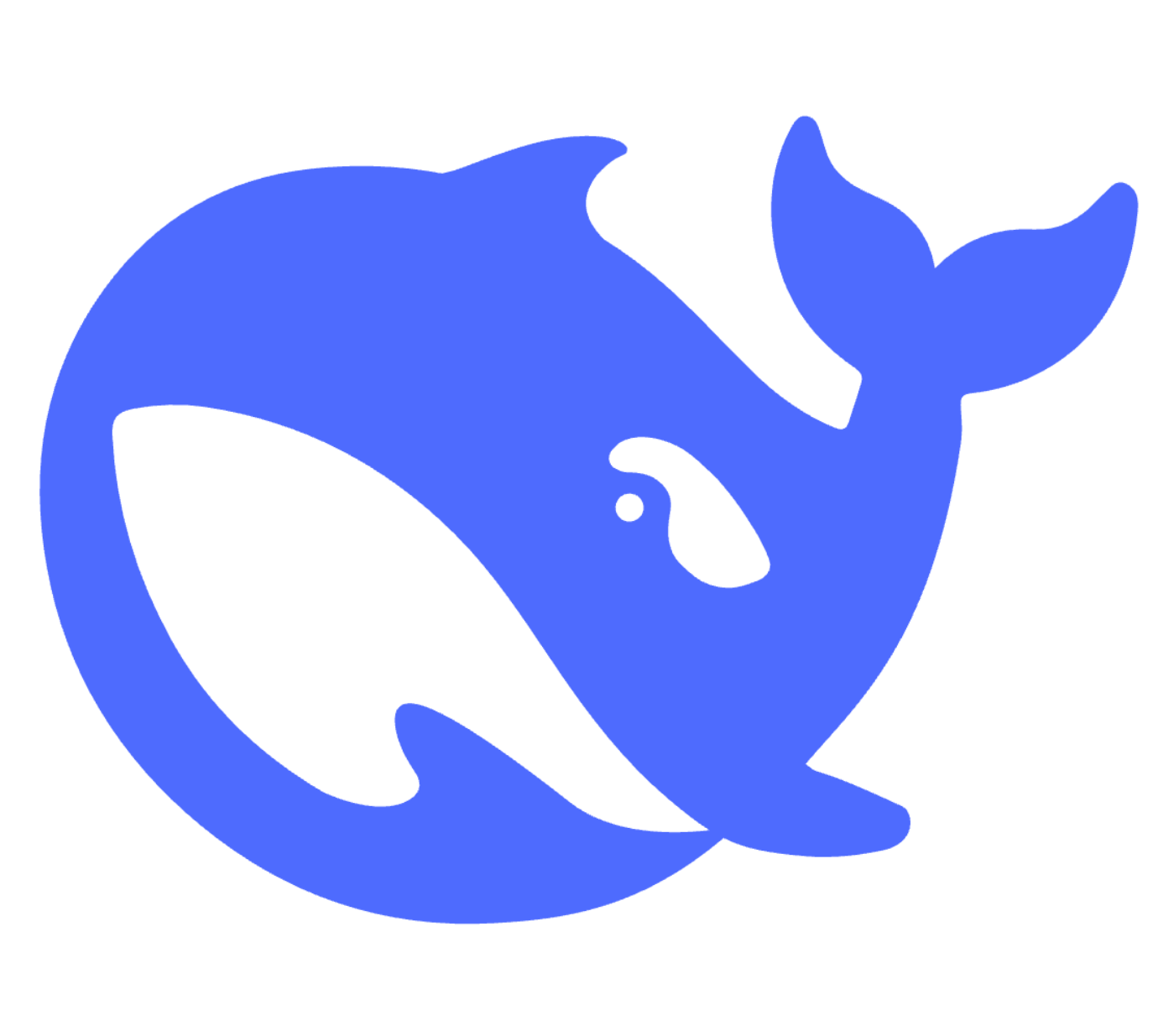}DeepSeek-V4.1-Flash & \underline{66.7} & \textbf{80.0} & \underline{60.0} & 0.0 & \textcolor{gray}{--} & 0.0 & 33.3 & 50.0 & 12.5 & 40.0 & 40.0 & \textcolor{gray}{--} & 60.0 & 60.0 & \textcolor{gray}{--} & \underline{66.7} & \underline{66.7} & \textcolor{gray}{--} & \cellcolor[HTML]{FCF2CD}44.9 & \cellcolor[HTML]{F5F1D1}56.0 & \cellcolor[HTML]{FCE3C6}22.0 \\
\osmodelicon{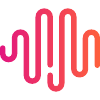}MiniMax M3.1 Flash & 26.7 & \underline{60.0} & 10.0 & 0.0 & \textcolor{gray}{--} & 0.0 & 0.0 & 0.0 & 0.0 & 20.0 & 20.0 & \textcolor{gray}{--} & 80.0 & 80.0 & \textcolor{gray}{--} & 11.1 & 11.1 & \textcolor{gray}{--} & \cellcolor[HTML]{F6C9BC}25.0 & \cellcolor[HTML]{F8D1BF}35.0 & \cellcolor[HTML]{F3BCB7}4.0 \\
\osmodelicon{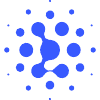}GLM 5.3 Flash & 26.7 & 40.0 & 20.0 & 0.0 & \textcolor{gray}{--} & 0.0 & 16.7 & 0.0 & \underline{37.5} & 0.0 & 0.0 & \textcolor{gray}{--} & 60.0 & 60.0 & \textcolor{gray}{--} & 0.0 & 0.0 & \textcolor{gray}{--} & \cellcolor[HTML]{F5C4BA}22.5 & \cellcolor[HTML]{F3BDB7}26.0 & \cellcolor[HTML]{F9D6C1}16.0 \\
\osmodelicon{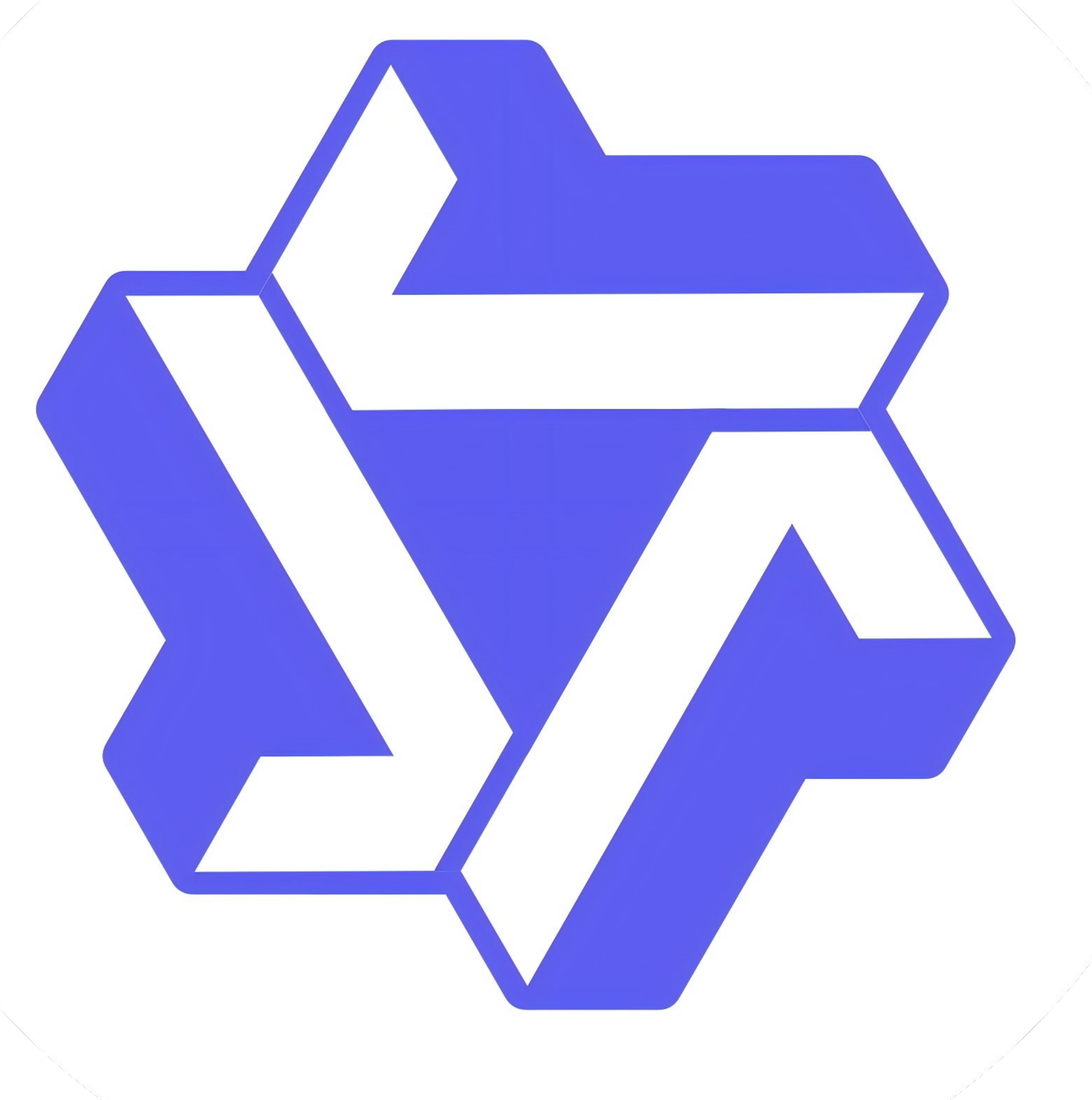}Qwen 3.8 27B & 20.0 & \underline{60.0} & 0.0 & \underline{10.0} & \textcolor{gray}{--} & \underline{10.0} & 0.0 & 0.0 & 0.0 & 0.0 & 0.0 & \textcolor{gray}{--} & 80.0 & 80.0 & \textcolor{gray}{--} & 11.1 & 11.1 & \textcolor{gray}{--} & \cellcolor[HTML]{F2B8B5}17.2 & \cellcolor[HTML]{F2B8B5}24.0 & \cellcolor[HTML]{F2B8B5}2.0 \\
\bottomrule
\end{tabular}%
}
\resizebox{\textwidth}{!}{%
\begin{tabular*}{\wd\mainresultsbox}{@{\extracolsep{\fill}}lcccccccccccccccccc}
\toprule
\multirow{3}{*}{\textbf{Model / Agent}} & \multicolumn{18}{c}{\textbf{Resolution Rate (\%) $\uparrow$}} \\
\cmidrule(l){2-19}
 & \multicolumn{3}{c}{\textbf{\begin{tabular}[b]{@{}c@{}}Minecraft\end{tabular}}} & \multicolumn{3}{c}{\textbf{\begin{tabular}[b]{@{}c@{}}Stardew\\Valley\end{tabular}}} & \multicolumn{3}{c}{\textbf{\begin{tabular}[b]{@{}c@{}}Garry's\\Mod\end{tabular}}} & \multicolumn{3}{c}{\textbf{\begin{tabular}[b]{@{}c@{}}Don't Starve\\Together\end{tabular}}} & \multicolumn{3}{c}{\textbf{\begin{tabular}[b]{@{}c@{}}Terraria\end{tabular}}} & \multicolumn{3}{c}{\textbf{\begin{tabular}[b]{@{}c@{}}RimWorld\end{tabular}}} \\
\cmidrule(lr){2-4} \cmidrule(lr){5-7} \cmidrule(lr){8-10} \cmidrule(lr){11-13} \cmidrule(lr){14-16} \cmidrule(lr){17-19}
 & All & Rep. & Con. & All & Rep. & Con. & All & Rep. & Con. & All & Rep. & Con. & All & Rep. & Con. & All & Rep. & Con. \\
\midrule
\osmodelicon{claude-logo.pdf}Claude Opus 5.5 & \underline{93.3} & \textbf{100.0} & \underline{80.0} & \textbf{81.8} & \textbf{88.9} & \textbf{50.0} & \underline{47.1} & \underline{66.7} & 0.0 & \underline{73.3} & \underline{60.0} & \textbf{80.0} & \textbf{60.0} & \textbf{60.0} & \textcolor{gray}{--} & \textbf{60.0} & \textbf{60.0} & \textcolor{gray}{--} \\
\osmodelicon{openai.pdf}GPT-6 Astra & \textbf{100.0} & \textbf{100.0} & \textbf{100.0} & 63.6 & \underline{77.8} & 0.0 & \underline{47.1} & \underline{66.7} & 0.0 & \textbf{80.0} & \textbf{80.0} & \textbf{80.0} & \underline{50.0} & \underline{50.0} & \textcolor{gray}{--} & \underline{50.0} & \underline{50.0} & \textcolor{gray}{--} \\
\osmodelicon{kimi.pdf}Kimi K3.1 Pro & 80.0 & \underline{90.0} & 60.0 & 54.5 & 66.7 & 0.0 & \textbf{52.9} & \textbf{75.0} & 0.0 & 40.0 & \underline{60.0} & \underline{30.0} & \underline{50.0} & \underline{50.0} & \textcolor{gray}{--} & \textbf{60.0} & \textbf{60.0} & \textcolor{gray}{--} \\
\osmodelicon{deepseek.pdf}DeepSeek-V4.1-Flash & 73.3 & \underline{90.0} & 40.0 & \underline{72.7} & \textbf{88.9} & 0.0 & 29.4 & 41.7 & 0.0 & 26.7 & 40.0 & 20.0 & 20.0 & 20.0 & \textcolor{gray}{--} & \underline{50.0} & \underline{50.0} & \textcolor{gray}{--} \\
\osmodelicon{minimax.pdf}MiniMax M3.1 Flash & 66.7 & \underline{90.0} & 20.0 & 18.2 & 22.2 & 0.0 & 23.5 & 33.3 & 0.0 & 13.3 & 40.0 & 0.0 & 10.0 & 10.0 & \textcolor{gray}{--} & 30.0 & 30.0 & \textcolor{gray}{--} \\
\osmodelicon{zhipu.pdf}GLM 5.3 Flash & 60.0 & 60.0 & 60.0 & 27.3 & 33.3 & 0.0 & 5.9 & 8.3 & 0.0 & 13.3 & 40.0 & 0.0 & 10.0 & 10.0 & \textcolor{gray}{--} & \underline{50.0} & \underline{50.0} & \textcolor{gray}{--} \\
\osmodelicon{qwen.pdf}Qwen 3.8 27B & 46.7 & 70.0 & 0.0 & 9.1 & 11.1 & 0.0 & 5.9 & 8.3 & 0.0 & 13.3 & 40.0 & 0.0 & 0.0 & 0.0 & \textcolor{gray}{--} & 10.0 & 10.0 & \textcolor{gray}{--} \\
\bottomrule
\end{tabular*}%
}

\vspace{4pt}

\resizebox{\textwidth}{!}{\usebox{\mainresultsbox}}
\end{table*}

\paragraph{Live-runtime tasks are far from solved.}
Claude Opus 5.5 achieves the highest overall resolution rate at 69.0\%, followed by GPT-6 Astra at 61.0\% and Kimi K3.1 Pro at 49.5\%. DeepSeek-V4.1-Flash resolves 44.9\%, and MiniMax M3.1 Flash, GLM 5.3 Flash, and Qwen 3.8 27B stay at or below 25.0\%. Because a task is resolved only when the submission works in the live game together with the third-party mods it depends on and leaves existing functionality intact, these rates show that producing code that works where it runs remains difficult even for frontier models. Construction is harder than repair for every model: Claude Opus 5.5 resolves 75.0\% of the \nrepair{} repair tasks but 56.0\% of the \nconstruct{} construction tasks, GPT-6 Astra 71.0\% and 44.0\%, Kimi K3.1 Pro 61.0\% and 26.0\%, and DeepSeek-V4.1-Flash 56.0\% and 22.0\%, while the three weakest models resolve between one and eight construction tasks (2.0\% to 16.0\%).

\paragraph{Model rankings are largely consistent across games.}
Claude Opus 5.5 achieves the highest rate, alone or tied, in eight of the twelve games. The Spearman correlation between each game's ranking of the seven models and their overall ranking averages 0.81 and is at least 0.79 in ten of the twelve games. The exceptions are Balatro (0.64), where Qwen 3.8 27B and MiniMax M3.1 Flash resolve 80.0\% of the tasks, and Palworld (0.13), where only Claude Opus 5.5 and Qwen 3.8 27B resolve any task.

\paragraph{Game difficulty depends on the model.}
Per-game rates nonetheless vary widely within each model: from 40.0\% on Palworld to 93.3\% on Minecraft for Claude Opus 5.5, from 0.0\% on Palworld to 100.0\% on Minecraft and Balatro for GPT-6 Astra, and from 0.0\% on Palworld to 80.0\% for Kimi K3.1 Pro. Palworld is the hardest game overall: five of the seven models resolve none of its tasks. Since task difficulty is calibrated across games (Section~\ref{sec:construction}), these differences suggest that the modding ecosystems themselves, with their languages, interfaces, and conventions, pose unequal challenges, and that an agent's competence in one ecosystem does not carry over uniformly to others.

\makeatletter
\@ifundefined{stable}{\newenvironment{stable}[1][tbp]{\@float{table}[#1]}{\end@float}}{}
\makeatother
\providecommand{\fitcolumn}[1]{\resizebox{\ifdim\width>\columnwidth\columnwidth\else\width\fi}{!}{#1}}

\section{Analysis}
\label{sec:analysis}
\label{sec:error-analysis}

We now examine why submissions fail and what the live runtime contributes to the verdict.

\begin{figure}[t]
\centering
\includegraphics[width=\linewidth]{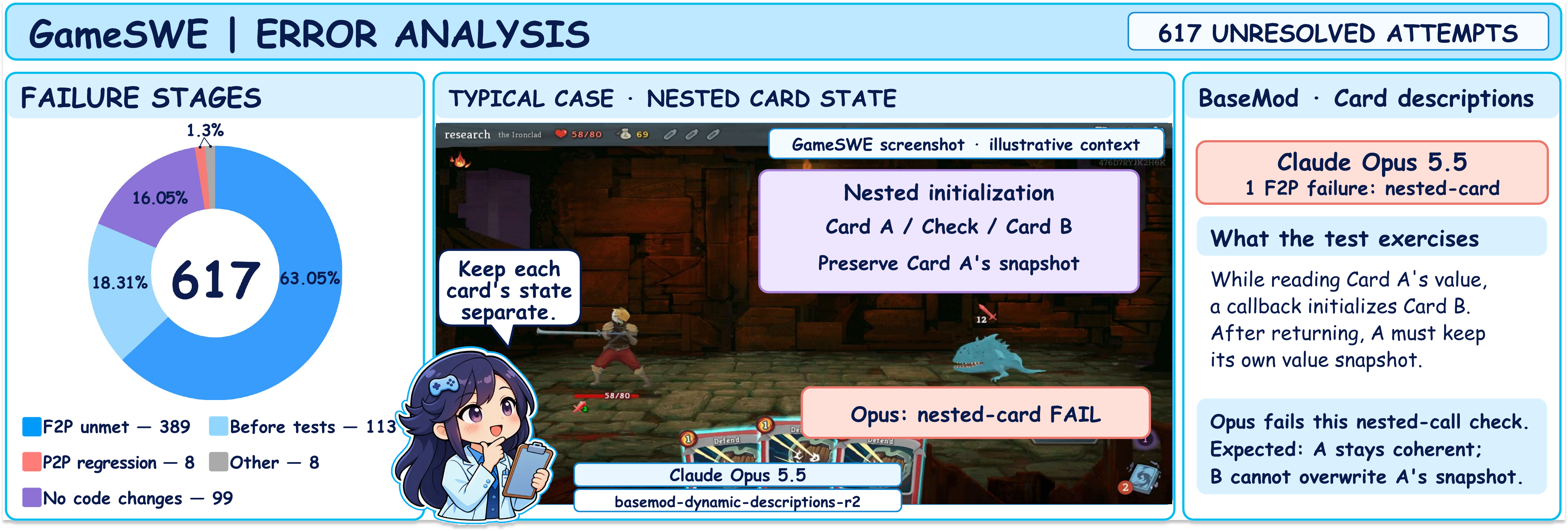}
\caption{\textbf{Failure distribution and a representative runtime-state error..}}
\label{fig:error-dist}
\end{figure}

\subsection{Where submissions fail in the live runtime}
\label{sec:funnel}

We follow each model's attempts through the stages of the harness (Table~\ref{tab:failure-stages}). Of the 1{,}050 attempts, 99 (9.4\%) leave the initial code unchanged. The largest loss comes later: 389 attempts (37.0\%) reach the tests in the running game but fail at least one fail-to-pass test, and 16 pass every audited fail-to-pass test but are still unresolved (Section~\ref{sec:runtime-adds}). In total, 427 of the 617 unresolved attempts (69.2\%) are loaded into the game and fail there, at load time or in its tests, which no check outside the live runtime would reveal.

\subsection{Agents' own checks are a poor proxy for the runtime}
\label{sec:own-checks}

Without access to the game, agents verify their work with syntax checks, text searches, and mock harnesses built on their own assumptions. Of the 617 unresolved attempts, 432 (70.0\%) end with the agent submitting its work as complete, and in at least 191 (31.0\%) the agent's final message states that its own checks or tests pass or that it has verified the work (Table~\ref{tab:failure-stages}).

\subsection{What the live runtime adds to the verdict}
\label{sec:runtime-adds}

To quantify what live-runtime verification contributes, we rescore the same submissions under weaker verification conditions, without new generations. \emph{F2P only} counts a task as resolved when all F2P tests pass, ignoring P2P tests. We then re-run the frozen submissions under two stronger conditions: supplementary P2P tests for the tasks that have none, and a pilot that loads unrelated, independently developed mods beside each submission. Table~\ref{tab:runtime-adds} compares the F2P-only condition with the full harness.

\begin{stable}[!ht]
\centering
\small
\caption{\textbf{Resolution rate (\%) under weaker verification.} The same submissions are rescored; differences show failures that each part of the live-runtime harness detects. Rates average the per-game rates, as in Table~\ref{tab:main_results}.}
\label{tab:runtime-adds}
\begin{tabular}{@{}lcc@{}}
\toprule
Model & F2P only & Full harness\\
\midrule
Claude Opus 5.5 & 72.5 & 69.0\\
GPT-6 Astra & 62.7 & 61.0\\
Kimi K3.1 Pro & 51.9 & 49.5\\
DeepSeek-V4.1-Flash & 47.7 & 44.9\\
MiniMax M3.1 Flash & 25.0 & 25.0\\
GLM 5.3 Flash & 23.3 & 22.5\\
Qwen 3.8 27B & 17.2 & 17.2\\
\bottomrule
\end{tabular}
\end{stable}

Ignoring pass-to-pass tests changes the verdict little: the F2P-only rate exceeds the full-harness rate by at most 3.5 points (Claude Opus 5.5) and equals it for MiniMax M3.1 Flash and Qwen 3.8 27B. 

\paragraph{Supplementary pass-to-pass tests.}
The 57 tasks without P2P tests, 49 constructions and 8 repairs, cannot reveal a submission that implements the request but breaks behavior that already worked. For these tasks we wrote a supplementary suite of 140 P2P tests: 74 check a requirement stated in the task description, such as that a Palworld mod ``must not pollute any other global name'', and 66 check base-game behavior that the mod has no reason to alter, such as saving and reloading the world. A test was admitted only if it passed twice on the initial code (or, where the initial code cannot load, on a documented baseline) and twice on the reference solution, passed on every existing alternative solution, and failed on a copy of the reference with one injected regression. The suite was frozen before any submission was evaluated and runs as a separate study condition, so task versions and Table~\ref{tab:main_results} are unchanged. Of the 399 submissions for these tasks, 258 have a result for every supplementary test and pass them all, 5 fail at least one (Table~\ref{tab:stronger}), and 136 have an unobserved test and do not count as passing. The five also fail F2P tests, and all 109 resolved submissions are among the 258. On the tasks that lacked P2P tests, submissions implementing the request thus rarely break what already worked.

\begin{stable}[!ht]
\centering
\footnotesize
\setlength{\tabcolsep}{3pt}
\caption{\textbf{Supplementary P2P tests and co-loaded mods.} Left: of each model's 57 submissions for the tasks without P2P tests, those observed (all supplementary tests have a result and pass, or one fails), those that fail at least one, and those among them that pass every F2P test. Right: pairs of evaluations without and with three co-loaded mods (Minecraft, Stardew Valley, RimWorld, and the Don't Starve Together constructions) in which all three mods are confirmed loaded, and the submissions resolved in each.}
\label{tab:stronger}
\fitcolumn{%
\begin{tabular}{@{}lccc@{\hspace{8pt}}ccc@{}}
\toprule
& \multicolumn{3}{c}{Supplementary P2P} & \multicolumn{3}{c}{Co-loaded mods}\\
\cmidrule(lr){2-4}\cmidrule(l){5-7}
Model & Observed & Fail & \begin{tabular}[b]{@{}c@{}}Fail,\\F2P pass\end{tabular} & Pairs & \begin{tabular}[b]{@{}c@{}}Res.\\without\end{tabular} & \begin{tabular}[b]{@{}c@{}}Res.\\with\end{tabular}\\
\midrule
Claude Opus 5.5 & 52 & 0 & 0 & 46 & 37 & 37\\
GPT-6 Astra & 49 & 0 & 0 & 41 & 35 & 35\\
Kimi K3.1 Pro & 41 & 0 & 0 & 41 & 27 & 27\\
DeepSeek-V4.1-Flash & 43 & 1 & 0 & 41 & 26 & 26\\
MiniMax M3.1 Flash & 23 & 1 & 0 & 28 & 15 & 15\\
GLM 5.3 Flash & 31 & 0 & 0 & 39 & 17 & 17\\
Qwen 3.8 27B & 24 & 3 & 0 & 33 & 9 & 9\\
\midrule
All & 263 & 5 & 0 & 269 & 166 & 166\\
\bottomrule
\end{tabular}}
\end{stable}

\paragraph{Co-loading independently developed mods.}
The full harness loads third-party mods only where the target mod depends on them, in 20 tasks. To test whether unrelated mods sharing the runtime change the verdict, we ran a pilot in four games that loads three popular mods, pinned by hash, beside every submission. We re-evaluated the 322 submissions for these tasks with and without the mods. In the 269 pairs in which the game's log or an in-game sentinel confirms that all three mods loaded, the outcome and every test result are identical, with one exception: in three Don't Starve Together submissions of one model, Show Me prints an error line that names the submission's mod, and the harness attributes the line to the submission, so their tests inside the game count as unobserved and one failed run becomes unattributed, although the game logs show the same test results.

\subsection{Failures at the boundary between code and host}
\label{sec:taxonomy}

The unresolved submissions that change the code fail in recurring ways, all of which involve the interface between the submitted code and the host game.

\begin{stable}[!ht]
\centering
\footnotesize
\setlength{\tabcolsep}{3pt}
\caption{\textbf{Unresolved attempts by failure stage and by the agent's final report.} The four stage columns partition each model's unresolved attempts: no change to the initial code, failure before any test runs (build, source gate, or load), failure of at least one F2P test, and failure after all audited F2P tests pass. \emph{Submitted} counts attempts that end with the agent submitting its work as complete; \emph{Claims pass} counts those whose final message states that the agent's own checks or tests pass.}
\label{tab:failure-stages}
\fitcolumn{%
\begin{tabular}{@{}lrrrrrrr@{}}
\toprule
& & \multicolumn{4}{c}{Failure stage} & \multicolumn{2}{c}{Final report}\\
\cmidrule(lr){3-6}\cmidrule(l){7-8}
Model & \begin{tabular}[b]{@{}c@{}}Unre-\\solved\end{tabular} & \begin{tabular}[b]{@{}c@{}}No\\change\end{tabular} & \begin{tabular}[b]{@{}c@{}}Before\\tests\end{tabular} & F2P & \begin{tabular}[b]{@{}c@{}}After\\F2P\end{tabular} & \begin{tabular}[b]{@{}c@{}}Sub-\\mitted\end{tabular} & \begin{tabular}[b]{@{}c@{}}Claims\\pass\end{tabular}\\
\midrule
Claude Opus 5.5 & 47 & 1 & 4 & 37 & 5 & 43 & 10\\
GPT-6 Astra & 57 & 0 & 10 & 45 & 2 & 57 & 33\\
Kimi K3.1 Pro & 76 & 1 & 15 & 56 & 4 & 74 & 24\\
DeepSeek-V4.1-Flash & 83 & 2 & 10 & 67 & 4 & 72 & 28\\
MiniMax M3.1 Flash & 113 & 3 & 33 & 77 & 0 & 88 & 18\\
GLM 5.3 Flash & 116 & 70 & 11 & 34 & 1 & 15 & 9\\
Qwen 3.8 27B & 125 & 22 & 30 & 73 & 0 & 83 & 69\\
\midrule
All & 617 & 99 & 113 & 389 & 16 & 432 & 191\\
\bottomrule
\end{tabular}}
\end{stable}

\paragraph{Interfaces that the ecosystem does not provide.}
Agents write plausible code against interfaces they assume rather than those the existing code and libraries provide. Four of GPT-6 Astra's RimWorld submissions do not compile against the pinned RimWorld 1.6 API: they use a member \texttt{AnyColonyThing} that \texttt{Tradeable} does not define, call a method of \texttt{GenPlace} that does not exist (\texttt{PlaceNearMaxRadialCells}), pass a parameter \texttt{checkTemperature} that \texttt{CanEverPlantAt} does not take, and call \texttt{GenerateQualityGeneratingPawn} with three arguments, which no overload accepts. An assumed interface can also compile and still fail: in a Balatro repair task, DeepSeek-V4.1-Flash registers its Jokers only if \texttt{SMODS.Joker} is a function, but the Steamodded loader defines it as a callable table, so none of the three requested Jokers exists in the game.

\paragraph{Host conventions beyond the code.}
A mod must also follow the host's packaging conventions. The harness rejects five of GPT-6 Astra's Garry's Mod construction submissions because the addon contains files outside the layout the task requires, which admits only \texttt{addon.json}, Lua files under \texttt{lua/}, and the task's own instruction files, or in one case the agent's own verification script. Tests that call the mod's functions directly, without loading the mod into the game, would not detect these errors.

\paragraph{Logic not connected to the game's lifecycle.}
Some submissions implement the requested logic but do not connect it to every event through which the game reaches it. This is the most common failure: 389 unresolved attempts load and run in the game but fail at least one fail-to-pass test. GPT-6 Astra's repair of a Don't Starve Together mod passes 18 of its 19 tests but fails the one in which a Glommer statue saved before an option was switched on must still be repairable with Marble. Such failures appear only when the game executes the full sequence of events.

\section{Conclusion}
We introduced GameSWE, a benchmark that evaluates coding agents by executing their code in live game runtimes. Across 150 tasks in 12 games, the strongest agent achieves only 69.0\% resolution, with failures concentrated at the boundary between code and host environments. These failures often escape agents' own checks. As a living benchmark with expanding game coverage, GameSWE provides a reproducible testbed for evaluating whether agent-generated code works in the environments where it runs.

\label{main-text-end}

\clearpage
\bibliography{references}

\clearpage
\appendix
\section{Task Inventory}
\label{app:inventory}

Table~\ref{tab:inventory} lists the \nhosts{} games in \benchmark{}, together with the language of their mods, the runtime in which the harness runs them, and the number of tasks. Tasks derived from the same codebase form a source group, such as Automate for Stardew Valley, DarkRP for Garry's Mod, and Exodus for The Binding of Isaac, and we report results with these groups in mind. Task descriptions keep their original natural language; for example, the Loadout description is written in Chinese.

\begin{table}[!ht]
\centering\small
\caption{\textbf{Games in \benchmark{}.} Runtime is the environment in which the live-runtime harness runs the submission together with the third-party mods.}
\label{tab:inventory}
\begin{tabular}{@{}lllr@{}}
\toprule
Game & Language & Runtime & Tasks\\
\midrule
Minecraft & Java & Paper server & 15\\
Stardew Valley & C\# & Desktop game (SMAPI) & 11\\
Garry's Mod & GLua & Dedicated server & 17\\
Don't Starve Together & Lua & Dedicated server & 15\\
Terraria & C\# & Dedicated server (tModLoader) & 10\\
RimWorld & C\# & Desktop game & 10\\
Mindustry & Java & Headless server & 15\\
Palworld & Lua (UE4SS) & Dedicated server & 10\\
The Binding of Isaac & Lua & Desktop game & 18\\
Left 4 Dead 2 & SourcePawn, Squirrel & Dedicated server & 10\\
Balatro & Lua & Desktop game & 10\\
Slay the Spire & Java & Desktop game (ModTheSpire) & 9\\
\midrule
Total & & & \ntasks{}\\
\bottomrule
\end{tabular}
\end{table}

\section{Supplementary P2P Tests and Co-Loading Pilot}
\label{app:supp}

\paragraph{Supplementary P2P suite.}
The 57 tasks without P2P tests receive a supplementary suite of 140 tests (Table~\ref{tab:supp-games}). Each test is justified by a quoted requirement of the task description or by base-game behavior that the mod has no reason to change. A test was admitted only if it passed twice on the initial code and twice on the reference solution, passed on every existing alternative solution and on every existing control that fails only F2P tests, and failed on a copy of the reference with one injected regression while the F2P tests still passed; tests that did not meet this bar were dropped, not weakened. Where the initial code of a construction task cannot load, a documented baseline replaces it, such as a harness-owned mod that does nothing or the unmodified server. The suite of each game was committed before any submission was evaluated and not changed afterwards. In a re-evaluation, the supplementary tests run in the same evaluation as the original tests, either inside the game's own harness or in a second session of the same frozen image, and never modify the original verdict. They count only when the submission builds, passes the task's static checks, and loads; otherwise all of them are unobserved. A submission therefore ends in one of three states: it \emph{passes} when every supplementary test has a result and passes, \emph{fails} when at least one test fails, and is \emph{not observed} otherwise, that is, when at least one test is unobserved and none fails. \emph{Observed} submissions are those that pass or fail. Garry's Mod and Left 4 Dead 2 submissions were evaluated twice, and the two repetitions are combined as in Appendix~\ref{app:verification}, where a valid failure is never replaced by a passing run: a submission fails if either repetition records a failed supplementary test, and it passes only if every supplementary test has a passing result and neither repetition records a failure. Eleven Palworld evaluations interrupted by a host-wide Docker stall were treated as invalid and re-run once.

\begin{table}[!ht]
\centering
\small
\setlength{\tabcolsep}{4pt}
\caption{\textbf{Supplementary P2P tests by game.} Tasks are repairs/constructions; tests are those checking a requirement of the task description / base-game behavior. Submissions are the frozen E1 submissions of the seven models, split into those that \emph{pass} (every supplementary test has a result and passes), \emph{fail} (at least one test fails), and are \emph{not observed} (at least one test is unobserved and none fails). Observed submissions in Table~\ref{tab:stronger} are those that pass or fail.}
\label{tab:supp-games}
\begin{tabular}{@{}lcccccc@{}}
\toprule
Game & Tasks & Tests & Submissions & Pass & Fail & \begin{tabular}[b]{@{}c@{}}Not\\observed\end{tabular}\\
\midrule
Minecraft & 0/5 & 4/8 & 35 & 25 & 0 & 10\\
Stardew Valley & 1/1 & 4/1 & 14 & 11 & 0 & 3\\
Garry's Mod & 1/5 & 8/8 & 42 & 10 & 0 & 32\\
Don't Starve Together & 0/10 & 1/21 & 70 & 39 & 2 & 29\\
Terraria & 4/0 & 7/3 & 28 & 21 & 1 & 6\\
Mindustry & 0/10 & 17/22 & 70 & 37 & 0 & 33\\
Palworld & 0/10 & 10/0 & 70 & 56 & 0 & 14\\
The Binding of Isaac & 0/8 & 15/3 & 56 & 52 & 0 & 4\\
Left 4 Dead 2 & 2/0 & 8/0 & 14 & 7 & 2 & 5\\
\midrule
All & 8/49 & 74/66 & 399 & 258 & 5 & 136\\
\bottomrule
\end{tabular}
\end{table}

\paragraph{Supplementary P2P results.}
Table~\ref{tab:stronger} in Section~\ref{sec:runtime-adds} gives the per-model counts. Each of the five failures traces to a defect in the submitted code: two Left 4 Dead 2 submissions reject valid configuration values or throw on every activation, a Terraria submission writes a map file whose header omits the type of each entry, and two Don't Starve Together submissions crash while the world is saved. All five also fail F2P tests. In Garry's Mod, 27 of the 35 addon submissions fail the task's static checks and 3 never load; of the remaining 5, 3 pass and 2 do not reach the state that some tests require and are not observed. Because a submission must build and load to be observed, coverage differs across models, from 52 of 57 submissions for Claude Opus 5.5 to 23 for MiniMax M3.1 Flash.

\paragraph{Co-loading pilot.}
Table~\ref{tab:coload} lists the co-loaded mods. Each mod is pinned by its SHA-256 hash, mounted read-only, and verified before every run; a mismatch stops the evaluation rather than running without the mod. Loading is confirmed from the game's own log, and in RimWorld from an in-game sentinel that counts each mod's assemblies and patches. Data Layers and Chests Anywhere are built from a pinned source commit, because their authors publish no release archives. The Don't Starve Together mods come from the Steam Workshop. The Don't Starve Together pilot covers the ten construction tasks. In RimWorld, two candidate mods that broke the reference solution or the game's start-up were replaced before any submission was evaluated. The three changed Don't Starve Together pairs come from one model, whose mod descriptions set no global name. Show Me logs this as an error that names the submission's mod, and the harness attributes the line to the submission, so the tests inside the game count as unobserved; the game logs show the same test results as without the mods.

\begin{table}[!ht]
\centering
\small
\setlength{\tabcolsep}{3.5pt}
\caption{\textbf{Co-loading pilot.} \emph{Pairs} are submissions evaluated with and without the mods; \emph{Loaded} pairs confirm all three mods; \emph{Tested} pairs also observe at least one test inside the game; \emph{Changed} pairs differ in any test result. $^\dagger$Construction tasks only.}
\label{tab:coload}
\begin{tabular}{@{}l>{\raggedright\arraybackslash}p{5.3cm}cccc@{}}
\toprule
Game & Mods & Pairs & Loaded & Tested & Changed\\
\midrule
Minecraft & LuckPerms 5.5.53, VaultUnlocked 2.19.1, PlaceholderAPI 2.12.2 & 105 & 88 & 88 & 0\\
Stardew Valley & Data Layers 1.23.0, Chests Anywhere 1.30.1, Easy Access 2.3.0 & 77 & 64 & 64 & 0\\
RimWorld & Interaction Bubbles 4.2, Pick Up And Haul (2025-08-17 release), Achtung! 4.1.14 & 70 & 53 & 46 & 0\\
Don't Starve Together$^\dagger$ & Global Positions 1.7.6, Show Me 0.58, Health Info 2.2.04 & 70 & 64 & 41 & 3\\
\midrule
All & & 322 & 269 & 239 & 3\\
\bottomrule
\end{tabular}
\end{table}

\section{Statistical Robustness and Task-Level Results}
\label{app:uncertainty}

The model ranking of Table~\ref{tab:main_results} is statistically robust. The seven models form three tiers: Claude Opus 5.5 and GPT-6 Astra; Kimi K3.1 Pro and DeepSeek-V4.1-Flash; and MiniMax M3.1 Flash, GLM 5.3 Flash, and Qwen 3.8 27B. All \UncPairsSignificant{} pairs of models from different tiers differ significantly under both tests of Table~\ref{tab:unc-pairs}: an exact McNemar test on the tasks that only one model of the pair resolves, Holm-adjusted over the \UncPairsTotal{} pairs, and a paired bootstrap interval for the difference in game-averaged rate. Within the top tier, Claude Opus 5.5 leads GPT-6 Astra by 8.0 points, with a paired interval of [2.2, 14.4]. The ranking also holds within games, where its Spearman correlation with the overall ranking averages \UncSpearmanMean{} (Table~\ref{tab:unc-games}), and injected defects, which did not exist publicly before the benchmark, are resolved at least as often as reverted upstream fixes (Table~\ref{tab:unc-origin}).

The intervals resample the \UncGroups{} source groups, so that tasks derived from one codebase, such as the Exodus tasks of The Binding of Isaac or the DarkRP tasks of Garry's Mod, move together. Table~\ref{tab:unc-models} gives each model's pooled rate over the \ntasks{} tasks and its game-averaged rate with 95\% percentile intervals that resample tasks or source groups within each game, or the games themselves. Table~\ref{tab:unc-thirdparty} reports the tasks that load third-party mods, and Table~\ref{tab:task-level} and the two tables that follow it list the outcome of every attempt. The analysis script and the per-attempt records are released with the benchmark.

\providecommand{\fitlinewidth}[1]{\resizebox{\ifdim\width>\linewidth\linewidth\else\width\fi}{!}{#1}}
\begin{table}[!ht]
\centering\footnotesize
\setlength{\tabcolsep}{3pt}
\caption{\textbf{Resolution rates with 95\% bootstrap intervals.} \emph{Pooled} is the share of the \ntasks{} tasks resolved; \emph{Game avg.} is the equally weighted mean of the 12 per-game rates used in Table~\ref{tab:main_results}. Intervals are percentile intervals over \UncB{} resamples: \emph{task} resamples tasks within each game, \emph{source} resamples the \UncGroups{} source groups within each game, so that tasks derived from one codebase move together, and \emph{game} resamples the 12 games.}
\label{tab:unc-models}
\fitlinewidth{\begin{tabular}{@{}lccccc@{}}
\toprule
& & & \multicolumn{3}{c}{Game avg.\ 95\% interval}\\
\cmidrule(l){4-6}
Model & Pooled [source] & Game avg. & task & source & game\\
\midrule
Claude Opus 5.5 & 103/150 = 68.7 [62.7, 74.3] & 69.0 & [61.7, 76.1] & [61.8, 74.2] & [60.2, 77.6]\\
GPT-6 Astra & 93/150 = 62.0 [55.1, 68.5] & 61.0 & [54.0, 67.7] & [52.4, 67.2] & [46.1, 75.0]\\
Kimi K3.1 Pro & 74/150 = 49.3 [43.0, 55.6] & 49.5 & [41.9, 56.8] & [42.2, 57.0] & [36.8, 61.0]\\
DeepSeek-V4.1-Flash & 67/150 = 44.7 [37.0, 52.6] & 44.9 & [37.6, 52.1] & [36.7, 52.3] & [31.6, 57.2]\\
MiniMax M3.1 Flash & 37/150 = 24.7 [18.5, 31.7] & 25.0 & [19.1, 31.2] & [19.5, 33.5] & [12.8, 39.5]\\
GLM 5.3 Flash & 34/150 = 22.7 [16.8, 29.5] & 22.5 & [16.7, 28.3] & [17.2, 29.7] & [10.7, 35.6]\\
Qwen 3.8 27B & 25/150 = 16.7 [11.4, 22.8] & 17.2 & [12.5, 22.3] & [12.7, 25.4] & [6.4, 31.5]\\
\bottomrule
\end{tabular}}
\end{table}

\begin{table}[!ht]
\centering\footnotesize
\setlength{\tabcolsep}{2.5pt}
\caption{\textbf{Paired model comparisons on the same tasks.} Below the diagonal: difference in game-averaged rate (row minus column, points) with a 95\% interval from the paired source-group bootstrap. Above the diagonal: tasks resolved only by the row model / only by the column model, and the exact McNemar $p$-value, Holm-adjusted over the \UncPairsTotal{} pairs.}
\label{tab:unc-pairs}
\fitlinewidth{\begin{tabular}{@{}lccccccc@{}}
\toprule
& Claude & GPT & Kimi & DS & MiniMax & GLM & Qwen\\
\midrule
Claude & -- & \begin{tabular}[c]{@{}c@{}}20/10\\[-1pt]\tiny $p$=0.39\end{tabular} & \begin{tabular}[c]{@{}c@{}}35/6\\[-1pt]\tiny $p$$<$0.001\end{tabular} & \begin{tabular}[c]{@{}c@{}}41/5\\[-1pt]\tiny $p$$<$0.001\end{tabular} & \begin{tabular}[c]{@{}c@{}}70/4\\[-1pt]\tiny $p$$<$0.001\end{tabular} & \begin{tabular}[c]{@{}c@{}}70/1\\[-1pt]\tiny $p$$<$0.001\end{tabular} & \begin{tabular}[c]{@{}c@{}}81/3\\[-1pt]\tiny $p$$<$0.001\end{tabular}\\[2pt]
GPT & \begin{tabular}[c]{@{}c@{}}-8.0\\[-1pt]\tiny[-14.4, -2.2]\end{tabular} & -- & \begin{tabular}[c]{@{}c@{}}27/8\\[-1pt]\tiny $p$=0.011\end{tabular} & \begin{tabular}[c]{@{}c@{}}35/9\\[-1pt]\tiny $p$$<$0.001\end{tabular} & \begin{tabular}[c]{@{}c@{}}57/1\\[-1pt]\tiny $p$$<$0.001\end{tabular} & \begin{tabular}[c]{@{}c@{}}62/3\\[-1pt]\tiny $p$$<$0.001\end{tabular} & \begin{tabular}[c]{@{}c@{}}69/1\\[-1pt]\tiny $p$$<$0.001\end{tabular}\\[2pt]
Kimi & \begin{tabular}[c]{@{}c@{}}-19.5\\[-1pt]\tiny[-25.3, -11.4]\end{tabular} & \begin{tabular}[c]{@{}c@{}}-11.5\\[-1pt]\tiny[-16.7, -4.6]\end{tabular} & -- & \begin{tabular}[c]{@{}c@{}}23/16\\[-1pt]\tiny $p$=0.67\end{tabular} & \begin{tabular}[c]{@{}c@{}}40/3\\[-1pt]\tiny $p$$<$0.001\end{tabular} & \begin{tabular}[c]{@{}c@{}}44/4\\[-1pt]\tiny $p$$<$0.001\end{tabular} & \begin{tabular}[c]{@{}c@{}}54/5\\[-1pt]\tiny $p$$<$0.001\end{tabular}\\[2pt]
DS & \begin{tabular}[c]{@{}c@{}}-24.1\\[-1pt]\tiny[-29.4, -18.0]\end{tabular} & \begin{tabular}[c]{@{}c@{}}-16.1\\[-1pt]\tiny[-22.4, -8.5]\end{tabular} & \begin{tabular}[c]{@{}c@{}}-4.6\\[-1pt]\tiny[-11.8, 1.9]\end{tabular} & -- & \begin{tabular}[c]{@{}c@{}}37/7\\[-1pt]\tiny $p$$<$0.001\end{tabular} & \begin{tabular}[c]{@{}c@{}}39/6\\[-1pt]\tiny $p$$<$0.001\end{tabular} & \begin{tabular}[c]{@{}c@{}}48/6\\[-1pt]\tiny $p$$<$0.001\end{tabular}\\[2pt]
MiniMax & \begin{tabular}[c]{@{}c@{}}-44.0\\[-1pt]\tiny[-49.3, -33.8]\end{tabular} & \begin{tabular}[c]{@{}c@{}}-36.0\\[-1pt]\tiny[-40.9, -26.9]\end{tabular} & \begin{tabular}[c]{@{}c@{}}-24.6\\[-1pt]\tiny[-29.9, -17.0]\end{tabular} & \begin{tabular}[c]{@{}c@{}}-19.9\\[-1pt]\tiny[-25.3, -11.3]\end{tabular} & -- & \begin{tabular}[c]{@{}c@{}}15/12\\[-1pt]\tiny $p$=0.70\end{tabular} & \begin{tabular}[c]{@{}c@{}}17/5\\[-1pt]\tiny $p$=0.085\end{tabular}\\[2pt]
GLM & \begin{tabular}[c]{@{}c@{}}-46.5\\[-1pt]\tiny[-52.2, -37.4]\end{tabular} & \begin{tabular}[c]{@{}c@{}}-38.5\\[-1pt]\tiny[-44.2, -29.7]\end{tabular} & \begin{tabular}[c]{@{}c@{}}-27.0\\[-1pt]\tiny[-33.3, -19.6]\end{tabular} & \begin{tabular}[c]{@{}c@{}}-22.4\\[-1pt]\tiny[-28.3, -14.4]\end{tabular} & \begin{tabular}[c]{@{}c@{}}-2.5\\[-1pt]\tiny[-9.6, 3.6]\end{tabular} & -- & \begin{tabular}[c]{@{}c@{}}19/10\\[-1pt]\tiny $p$=0.41\end{tabular}\\[2pt]
Qwen & \begin{tabular}[c]{@{}c@{}}-51.8\\[-1pt]\tiny[-57.2, -41.0]\end{tabular} & \begin{tabular}[c]{@{}c@{}}-43.8\\[-1pt]\tiny[-49.2, -33.7]\end{tabular} & \begin{tabular}[c]{@{}c@{}}-32.3\\[-1pt]\tiny[-38.4, -23.5]\end{tabular} & \begin{tabular}[c]{@{}c@{}}-27.7\\[-1pt]\tiny[-33.3, -18.6]\end{tabular} & \begin{tabular}[c]{@{}c@{}}-7.8\\[-1pt]\tiny[-12.1, -3.8]\end{tabular} & \begin{tabular}[c]{@{}c@{}}-5.3\\[-1pt]\tiny[-11.2, 1.9]\end{tabular} & --\\[2pt]
\bottomrule
\end{tabular}}
\end{table}

\begin{table}[!ht]
\centering\footnotesize
\setlength{\tabcolsep}{4pt}
\caption{\textbf{Per-game rates and ranking agreement.} \emph{Rate} pools all seven models' attempts on the game's tasks, with a 95\% interval from resampling the game's source groups. $\rho$ is the Spearman correlation between the game's ranking of the seven models and their overall ranking, with a 95\% source-group bootstrap interval.}
\label{tab:unc-games}
\fitlinewidth{\begin{tabular}{@{}lccccc@{}}
\toprule
Game & Tasks & Groups & Rate (\%) & $\rho$ & 95\% interval of $\rho$\\
\midrule
Minecraft & 15 & 5 & 74.3 [59.5, 98.2] & 0.96 & [0.41, 0.96]\\
Stardew Valley & 11 & 6 & 46.8 [7.1, 63.3] & 0.86 & [0.53, 0.94]\\
Garry's Mod & 17 & 10 & 30.3 [15.6, 48.6] & 0.87 & [0.74, 0.95]\\
Don't Starve Together & 15 & 5 & 37.1 [26.9, 80.0] & 0.93 & [0.19, 0.98]\\
Terraria & 10 & 10 & 28.6 [12.9, 44.3] & 0.98 & [0.78, 0.98]\\
RimWorld & 10 & 9 & 44.3 [23.4, 68.3] & 0.79 & [0.41, 0.95]\\
Mindustry & 15 & 2 & 47.6 [38.6, 65.7] & 0.87 & [0.73, 0.93]\\
Palworld & 10 & 1 & 7.1 [7.1, 7.1] & 0.13 & [0.13, 0.27]\\
The Binding of Isaac & 18 & 2 & 27.0 [23.2, 30.0] & 0.90 & [0.73, 0.96]\\
Left 4 Dead 2 & 10 & 5 & 35.7 [22.4, 51.4] & 0.98 & [0.87, 0.98]\\
Balatro & 10 & 1 & 78.6 [78.6, 78.6] & 0.64 & [0.54, 0.73]\\
Slay the Spire & 9 & 4 & 38.1 [23.2, 55.8] & 0.83 & [0.73, 0.95]\\
\midrule
Mean $\rho$ & & & & \UncSpearmanMean{} & \UncSpearmanMeanCI{}\\
\bottomrule
\end{tabular}}
\end{table}

\begin{table}[!ht]
\centering\footnotesize
\setlength{\tabcolsep}{3pt}
\caption{\textbf{Resolved repair tasks by origin.} Injected defects did not exist publicly before the benchmark; over all models they are resolved in 63.5\% of attempts, compared with 58.7\% for reverted upstream fixes.}
\label{tab:unc-origin}
\fitlinewidth{\begin{tabular}{@{}lcccc@{}}
\toprule
Model & \begin{tabular}[b]{@{}c@{}}Upstream fix reverted\\($n$=45)\end{tabular} & \begin{tabular}[b]{@{}c@{}}Latent upstream\\($n$=21)\end{tabular} & \begin{tabular}[b]{@{}c@{}}Reverted + injected\\($n$=16)\end{tabular} & \begin{tabular}[b]{@{}c@{}}Injected\\($n$=18)\end{tabular}\\
\midrule
Claude Opus 5.5 & 35/45 (77.8) & 15/21 (71.4) & 11/16 (68.8) & 14/18 (77.8)\\
GPT-6 Astra & 34/45 (75.6) & 11/21 (52.4) & 11/16 (68.8) & 15/18 (83.3)\\
Kimi K3.1 Pro & 30/45 (66.7) & 9/21 (42.9) & 8/16 (50.0) & 14/18 (77.8)\\
DeepSeek-V4.1-Flash & 31/45 (68.9) & 7/21 (33.3) & 7/16 (43.8) & 11/18 (61.1)\\
MiniMax M3.1 Flash & 22/45 (48.9) & 2/21 (9.5) & 0/16 (0.0) & 11/18 (61.1)\\
GLM 5.3 Flash & 18/45 (40.0) & 1/21 (4.8) & 0/16 (0.0) & 7/18 (38.9)\\
Qwen 3.8 27B & 15/45 (33.3) & 0/21 (0.0) & 1/16 (6.2) & 8/18 (44.4)\\
\midrule
All models & 185/315 (58.7) & 45/147 (30.6) & 38/112 (33.9) & 80/126 (63.5)\\
\bottomrule
\end{tabular}}
\end{table}

\begin{table}[!ht]
\centering\footnotesize
\setlength{\tabcolsep}{3pt}
\caption{\textbf{Resolved tasks that load third-party mods.} \emph{Loads mods}: the tasks whose environment $M$ is non-empty; \emph{P2P via mods}: those among them whose P2P tests exercise code paths through these mods; \emph{No mods}: the remaining 130 tasks whose environment $M$ is empty.}
\label{tab:unc-thirdparty}
\fitlinewidth{\begin{tabular}{@{}lccc@{}}
\toprule
Model & \begin{tabular}[b]{@{}c@{}}Loads mods\\($n$=20)\end{tabular} & \begin{tabular}[b]{@{}c@{}}P2P via mods\\($n$=7)\end{tabular} & \begin{tabular}[b]{@{}c@{}}No mods\\($n$=130)\end{tabular}\\
\midrule
Claude Opus 5.5 & 14/20 (70.0) & 5/7 (71.4) & 89/130 (68.5)\\
GPT-6 Astra & 11/20 (55.0) & 3/7 (42.9) & 82/130 (63.1)\\
Kimi K3.1 Pro & 10/20 (50.0) & 2/7 (28.6) & 64/130 (49.2)\\
DeepSeek-V4.1-Flash & 11/20 (55.0) & 4/7 (57.1) & 56/130 (43.1)\\
MiniMax M3.1 Flash & 4/20 (20.0) & 1/7 (14.3) & 33/130 (25.4)\\
GLM 5.3 Flash & 5/20 (25.0) & 0/7 (0.0) & 29/130 (22.3)\\
Qwen 3.8 27B & 1/20 (5.0) & 0/7 (0.0) & 24/130 (18.5)\\
\midrule
All models & 56/140 (40.0) & 15/49 (30.6) & 377/910 (41.4)\\
\bottomrule
\end{tabular}}
\end{table}

\begin{table}[p]
\centering\tiny
\setlength{\tabcolsep}{2pt}
\renewcommand{\arraystretch}{0.95}
\caption{\textbf{Task-level results.} One row per task, grouped by game: kind (R repair, C construction), origin of a repair (H reverted upstream fix, L latent upstream behavior against an authored contract, I injected defect, H+I both), source group, and each model's outcome (\cmark\ resolved, \textperiodcentered\ unresolved); $^{M}$ marks a task that loads third-party mods.}
\label{tab:task-level}
\fitlinewidth{\begin{tabular}{@{}lcclccccccc@{}}
\toprule
Task & Kind & Orig. & Source group & Claude & GPT & Kimi & DS & MiniMax & GLM & Qwen\\
\midrule
\multicolumn{11}{@{}l}{\textit{Minecraft}}\\
\texttt{betterteams-description-tag-injection} & R & H & BetterTeams & \cmark & \cmark & \cmark & \cmark & \cmark & \textperiodcentered & \cmark\\
\texttt{betterteams-help-page-bounds} & R & H & BetterTeams & \cmark & \cmark & \cmark & \cmark & \cmark & \cmark & \cmark\\
\texttt{chatty-bad-channel-disables-plugin} & R & H & Chatty & \cmark & \cmark & \cmark & \cmark & \cmark & \cmark & \cmark\\
\texttt{chatty-failed-migration-dead-plugin} & R & H & Chatty & \cmark & \cmark & \textperiodcentered & \cmark & \cmark & \textperiodcentered & \cmark\\
\texttt{chatty-reload-ignores-chat-commands} & R & H & Chatty & \cmark & \cmark & \cmark & \textperiodcentered & \textperiodcentered & \textperiodcentered & \textperiodcentered\\
\texttt{chatty-reload-leaks-threads} & R & H & Chatty & \cmark & \cmark & \cmark & \cmark & \cmark & \cmark & \textperiodcentered\\
\texttt{chatty-v2-migration-null-spy-format} & R & H & Chatty & \cmark & \cmark & \cmark & \cmark & \cmark & \textperiodcentered & \textperiodcentered\\
\texttt{multiverse-core-case-insensitive-world-names} & R & H & Multiverse-Core & \cmark & \cmark & \cmark & \cmark & \cmark & \cmark & \cmark\\
\texttt{multiverse-core-world-exists-by-key} & R & H & Multiverse-Core & \cmark & \cmark & \cmark & \cmark & \cmark & \cmark & \cmark\\
\texttt{prometheus-exporter-players-total-main-thread} & R & H & minecraft-prometheus-exporter & \cmark & \cmark & \cmark & \cmark & \cmark & \cmark & \cmark\\
\texttt{blocktrail-block-log-rollback} & C & -- & Minecraft constructions & \textperiodcentered & \cmark & \textperiodcentered & \cmark & \textperiodcentered & \textperiodcentered & \textperiodcentered\\
\texttt{coinledger-bank-ledger} & C & -- & Minecraft constructions & \cmark & \cmark & \textperiodcentered & \cmark & \textperiodcentered & \cmark & \textperiodcentered\\
\texttt{recipesmith-config-recipes} & C & -- & Minecraft constructions & \cmark & \cmark & \cmark & \textperiodcentered & \textperiodcentered & \textperiodcentered & \textperiodcentered\\
\texttt{towncrier-timed-announcements} & C & -- & Minecraft constructions & \cmark & \cmark & \cmark & \textperiodcentered & \cmark & \cmark & \textperiodcentered\\
\texttt{wardstone-region-protection} & C & -- & Minecraft constructions & \cmark & \cmark & \cmark & \textperiodcentered & \textperiodcentered & \cmark & \textperiodcentered\\
\midrule
\multicolumn{11}{@{}l}{\textit{Stardew Valley}}\\
\texttt{stardew-automate-minifridge-storage} & R & H & Automate & \cmark & \cmark & \cmark & \cmark & \textperiodcentered & \cmark & \textperiodcentered\\
\texttt{stardew-automate-output-count-drift} & R & I & Automate & \cmark & \cmark & \cmark & \cmark & \cmark & \cmark & \textperiodcentered\\
\texttt{stardew-automate-release-regressions} & R & H & Automate & \cmark & \cmark & \textperiodcentered & \cmark & \textperiodcentered & \textperiodcentered & \textperiodcentered\\
\texttt{stardew-automate-release-regressions-v2} & R & H+I & Automate & \cmark & \cmark & \cmark & \cmark & \textperiodcentered & \textperiodcentered & \textperiodcentered\\
\texttt{stardew-automate-release-regressions-v3} & R & H+I & Automate & \cmark & \cmark & \cmark & \cmark & \textperiodcentered & \textperiodcentered & \textperiodcentered\\
\texttt{stardew-automate-zero-stack-output} & R & H & Automate & \cmark & \cmark & \cmark & \cmark & \cmark & \cmark & \cmark\\
\texttt{stardew-cjb-cheats-regressions} & R & H & CJB Cheats Menu & \textperiodcentered & \textperiodcentered & \textperiodcentered & \textperiodcentered & \textperiodcentered & \textperiodcentered & \textperiodcentered\\
\texttt{stardew-cp-release-regressions} & R & H & Content Patcher & \cmark & \textperiodcentered & \textperiodcentered & \cmark & \textperiodcentered & \textperiodcentered & \textperiodcentered\\
\texttt{stardew-tractor-attachment-regressions} & R & H & Tractor Mod & \cmark & \cmark & \cmark & \cmark & \textperiodcentered & \textperiodcentered & \textperiodcentered\\
\texttt{stardew-auto-workshop} & C & -- & Stardew constructions & \textperiodcentered & \textperiodcentered & \textperiodcentered & \textperiodcentered & \textperiodcentered & \textperiodcentered & \textperiodcentered\\
\texttt{stardew-summon-horse-anywhere} & C & -- & Horse Flute Anywhere & \cmark & \textperiodcentered & \textperiodcentered & \textperiodcentered & \textperiodcentered & \textperiodcentered & \textperiodcentered\\
\midrule
\multicolumn{11}{@{}l}{\textit{Garry's Mod}}\\
\texttt{advdupe2-paste-regressions} & R & H & Advanced Duplicator 2 & \cmark & \cmark & \cmark & \textperiodcentered & \textperiodcentered & \textperiodcentered & \textperiodcentered\\
\texttt{cfw-contraption-regressions} & R & H & Contraption Framework & \cmark & \cmark & \cmark & \cmark & \cmark & \cmark & \textperiodcentered\\
\texttt{starfallex-library-regressions} & R & H & StarfallEx & \cmark & \cmark & \cmark & \cmark & \cmark & \textperiodcentered & \textperiodcentered\\
\texttt{wiremod-e2-library-regressions} & R & H & Wiremod & \cmark & \cmark & \cmark & \cmark & \textperiodcentered & \textperiodcentered & \textperiodcentered\\
\texttt{wiremod-gates-chips-regressions} & R & H & Wiremod & \cmark & \cmark & \cmark & \textperiodcentered & \cmark & \textperiodcentered & \cmark\\
\texttt{darkrp-core-library-regressions} & R & I & DarkRP & \cmark & \cmark & \cmark & \cmark & \textperiodcentered & \textperiodcentered & \textperiodcentered\\
\texttt{darkrp-door-economy-regressions} & R & I & DarkRP & \textperiodcentered & \textperiodcentered & \textperiodcentered & \textperiodcentered & \textperiodcentered & \textperiodcentered & \textperiodcentered\\
\texttt{darkrp-live-server-regressions} & R & H+I & DarkRP & \textperiodcentered & \textperiodcentered & \textperiodcentered & \textperiodcentered & \textperiodcentered & \textperiodcentered & \textperiodcentered\\
\texttt{darkrp-money-regressions} & R & I & DarkRP & \cmark & \textperiodcentered & \cmark & \textperiodcentered & \textperiodcentered & \textperiodcentered & \textperiodcentered\\
\texttt{darkrp-persistence-regressions} & R & I & DarkRP & \textperiodcentered & \textperiodcentered & \textperiodcentered & \textperiodcentered & \textperiodcentered & \textperiodcentered & \textperiodcentered\\
\texttt{darkrp-registration-regressions} & R & I & DarkRP & \textperiodcentered & \cmark & \cmark & \cmark & \cmark & \textperiodcentered & \textperiodcentered\\
\texttt{darkrp-release-regressions-v3} & R & H+I & DarkRP & \cmark & \cmark & \cmark & \textperiodcentered & \textperiodcentered & \textperiodcentered & \textperiodcentered\\
\texttt{contraption-blueprint-duplicator} & C & -- & Advanced Duplicator 2 (Workshop feature) & \textperiodcentered & \textperiodcentered & \textperiodcentered & \textperiodcentered & \textperiodcentered & \textperiodcentered & \textperiodcentered\\
\texttt{hot-object-round-controller} & C & -- & You Touched it Last (Workshop feature) & \textperiodcentered & \textperiodcentered & \textperiodcentered & \textperiodcentered & \textperiodcentered & \textperiodcentered & \textperiodcentered\\
\texttt{persistent-loadout-profiles} & C & -- & Custom Loadout (Workshop feature) & \textperiodcentered & \textperiodcentered & \textperiodcentered & \textperiodcentered & \textperiodcentered & \textperiodcentered & \textperiodcentered\\
\texttt{prop-ownership-quota-registry} & C & -- & Simple Prop Protection (Workshop feature) & \textperiodcentered & \textperiodcentered & \textperiodcentered & \textperiodcentered & \textperiodcentered & \textperiodcentered & \textperiodcentered\\
\texttt{staged-prop-alignment} & C & -- & Magic Align (Workshop feature) & \textperiodcentered & \textperiodcentered & \textperiodcentered & \textperiodcentered & \textperiodcentered & \textperiodcentered & \textperiodcentered\\
\midrule
\multicolumn{11}{@{}l}{\textit{Don't Starve Together}}\\
\texttt{composting-together-regressions} & R & H & Composting Together & \cmark & \cmark & \cmark & \cmark & \textperiodcentered & \cmark & \textperiodcentered\\
\texttt{dsmmo-release-regressions} & R & H & DsMMO & \cmark & \cmark & \cmark & \cmark & \cmark & \cmark & \cmark\\
\texttt{dst-fixed-bundle-regressions} & R & H & DST Fixed & \textperiodcentered & \cmark & \textperiodcentered & \textperiodcentered & \textperiodcentered & \textperiodcentered & \textperiodcentered\\
\texttt{dst-fixed-module-regressions} & R & H & DST Fixed & \cmark & \textperiodcentered & \textperiodcentered & \textperiodcentered & \textperiodcentered & \textperiodcentered & \textperiodcentered\\
\texttt{glassic-api-regressions} & R & H & Glassic API & \textperiodcentered & \cmark & \cmark & \textperiodcentered & \cmark & \textperiodcentered & \cmark\\
\texttt{camp-supply-ledger} & C & -- & DST constructions & \cmark & \cmark & \cmark & \textperiodcentered & \textperiodcentered & \textperiodcentered & \textperiodcentered\\
\texttt{combat-target-arbiter} & C & -- & DST constructions & \cmark & \textperiodcentered & \textperiodcentered & \textperiodcentered & \textperiodcentered & \textperiodcentered & \textperiodcentered\\
\bottomrule
\end{tabular}}
\end{table}

\begin{table}[p]
\centering\tiny
\setlength{\tabcolsep}{2pt}
\renewcommand{\arraystretch}{0.95}
\caption{\textbf{Task-level results (continued).}}
\label{tab:task-level-2}
\fitlinewidth{\begin{tabular}{@{}lcclccccccc@{}}
\toprule
Task & Kind & Orig. & Source group & Claude & GPT & Kimi & DS & MiniMax & GLM & Qwen\\
\midrule
\multicolumn{11}{@{}l}{\textit{Don't Starve Together}}\\
\texttt{container-transfer-transaction} & C & -- & DST constructions & \cmark & \cmark & \textperiodcentered & \textperiodcentered & \textperiodcentered & \textperiodcentered & \textperiodcentered\\
\texttt{fuel-ration-allocator} & C & -- & DST constructions & \textperiodcentered & \cmark & \textperiodcentered & \textperiodcentered & \textperiodcentered & \textperiodcentered & \textperiodcentered\\
\texttt{loot-roll-arbiter} & C & -- & DST constructions & \cmark & \cmark & \cmark & \textperiodcentered & \textperiodcentered & \textperiodcentered & \textperiodcentered\\
\texttt{rescue-dispatch-queue} & C & -- & DST constructions & \cmark & \cmark & \textperiodcentered & \cmark & \textperiodcentered & \textperiodcentered & \textperiodcentered\\
\texttt{status-effect-stack} & C & -- & DST constructions & \cmark & \cmark & \textperiodcentered & \textperiodcentered & \textperiodcentered & \textperiodcentered & \textperiodcentered\\
\texttt{trade-escrow-exchange} & C & -- & DST constructions & \textperiodcentered & \textperiodcentered & \textperiodcentered & \textperiodcentered & \textperiodcentered & \textperiodcentered & \textperiodcentered\\
\texttt{waystone-route-network} & C & -- & DST constructions & \cmark & \cmark & \cmark & \cmark & \textperiodcentered & \textperiodcentered & \textperiodcentered\\
\texttt{world-rule-ballot} & C & -- & DST constructions & \cmark & \cmark & \textperiodcentered & \textperiodcentered & \textperiodcentered & \textperiodcentered & \textperiodcentered\\
\midrule
\multicolumn{11}{@{}l}{\textit{Terraria}}\\
\texttt{terraria-aequus-carpentry-replication-hard} & R & L & Aequus & \textperiodcentered & \textperiodcentered & \textperiodcentered & \textperiodcentered & \textperiodcentered & \textperiodcentered & \textperiodcentered\\
\texttt{terraria-calamity-overhaul-time-replication-hard} & R & L & Calamity Overhaul & \cmark & \cmark & \cmark & \cmark & \textperiodcentered & \cmark & \textperiodcentered\\
\texttt{terraria-fargos-enchanted-tree-transactions-hard} & R & I & Fargo's Mutant Mod & \cmark & \cmark & \cmark & \cmark & \textperiodcentered & \textperiodcentered & \textperiodcentered\\
\texttt{terraria-fargos-toggle-presets-hard} & R & L & Fargo's Souls Mod & \textperiodcentered & \textperiodcentered & \textperiodcentered & \textperiodcentered & \textperiodcentered & \textperiodcentered & \textperiodcentered\\
\texttt{terraria-improvegame-space-wand-authority-hard} & R & L & Quality of Terraria & \cmark & \cmark & \cmark & \textperiodcentered & \cmark & \textperiodcentered & \textperiodcentered\\
\texttt{terraria-magic-storage-compressed-io-regressions-hard} & R & L & Magic Storage & \cmark & \textperiodcentered & \cmark & \textperiodcentered & \textperiodcentered & \textperiodcentered & \textperiodcentered\\
\texttt{terraria-modliquid-map-persistence-hard} & R & L & ModLiquid Library & \cmark & \cmark & \textperiodcentered & \textperiodcentered & \textperiodcentered & \textperiodcentered & \textperiodcentered\\
\texttt{terraria-overhaul-entity-chunk-lifetime-hard} & R & L & Terraria Overhaul & \cmark & \textperiodcentered & \cmark & \textperiodcentered & \textperiodcentered & \textperiodcentered & \textperiodcentered\\
\texttt{terraria-spirit-branching-quest-replication-hard} & R & L & Spirit Mod & \textperiodcentered & \cmark & \textperiodcentered & \textperiodcentered & \textperiodcentered & \textperiodcentered & \textperiodcentered\\
\texttt{terraria-starlight-ability-state-hard} & R & L & Starlight River & \textperiodcentered & \textperiodcentered & \textperiodcentered & \textperiodcentered & \textperiodcentered & \textperiodcentered & \textperiodcentered\\
\midrule
\multicolumn{11}{@{}l}{\textit{RimWorld}}\\
\texttt{adaptive-storage-slot-limit-restore} & R & H$^{M}$ & Adaptive Storage Framework & \cmark & \cmark & \cmark & \cmark & \cmark & \cmark & \cmark\\
\texttt{bill-autopilot-repeat-mode-ownership} & R & H$^{M}$ & Bill Autopilot & \textperiodcentered & \textperiodcentered & \textperiodcentered & \textperiodcentered & \textperiodcentered & \textperiodcentered & \textperiodcentered\\
\texttt{combat-readiness-colony-wealth} & R & H$^{M}$ & Combat Readiness Check (Continued) & \cmark & \cmark & \cmark & \cmark & \cmark & \cmark & \textperiodcentered\\
\texttt{customize-weapon-ability-tracker} & R & H$^{M}$ & Customize Weapon & \cmark & \cmark & \cmark & \cmark & \cmark & \cmark & \textperiodcentered\\
\texttt{playwright-nonbulk-trade-accounting} & R & H$^{M}$ & Playwright & \textperiodcentered & \textperiodcentered & \textperiodcentered & \textperiodcentered & \textperiodcentered & \textperiodcentered & \textperiodcentered\\
\texttt{project-rimfactory-cultivator-environment} & R & H$^{M}$ & Project RimFactory Revived & \textperiodcentered & \textperiodcentered & \cmark & \textperiodcentered & \textperiodcentered & \textperiodcentered & \textperiodcentered\\
\texttt{r4-small-bench-ingredient-consumption} & R & H$^{M}$ & R$^4$ & \textperiodcentered & \textperiodcentered & \textperiodcentered & \textperiodcentered & \textperiodcentered & \textperiodcentered & \textperiodcentered\\
\texttt{r4-uncraftable-material-basis} & R & H$^{M}$ & R$^4$ & \cmark & \cmark & \cmark & \cmark & \textperiodcentered & \cmark & \textperiodcentered\\
\texttt{research-organized-tab-state} & R & H$^{M}$ & Research: Organized & \cmark & \cmark & \textperiodcentered & \textperiodcentered & \textperiodcentered & \textperiodcentered & \textperiodcentered\\
\texttt{total-control-vanilla-gear-generation} & R & H$^{M}$ & Rimsential - Total Control: Continued & \cmark & \textperiodcentered & \cmark & \cmark & \textperiodcentered & \cmark & \textperiodcentered\\
\midrule
\multicolumn{11}{@{}l}{\textit{Mindustry}}\\
\texttt{data-patch-regressions} & R & H & Mindustry & \textperiodcentered & \textperiodcentered & \cmark & \cmark & \cmark & \textperiodcentered & \textperiodcentered\\
\texttt{defense-and-heat-regressions} & R & H & Mindustry & \cmark & \cmark & \cmark & \cmark & \cmark & \cmark & \cmark\\
\texttt{logic-processor-regressions} & R & H & Mindustry & \cmark & \cmark & \textperiodcentered & \cmark & \cmark & \cmark & \cmark\\
\texttt{payload-and-save-regressions} & R & H & Mindustry & \cmark & \cmark & \cmark & \cmark & \textperiodcentered & \textperiodcentered & \cmark\\
\texttt{server-console-regressions} & R & H & Mindustry & \cmark & \cmark & \textperiodcentered & \textperiodcentered & \textperiodcentered & \textperiodcentered & \textperiodcentered\\
\texttt{blueprint-layout-reconciler} & C & -- & Mindustry constructions & \cmark & \cmark & \cmark & \cmark & \textperiodcentered & \cmark & \textperiodcentered\\
\texttt{build-quota-auditor} & C & -- & Mindustry constructions & \cmark & \cmark & \cmark & \textperiodcentered & \textperiodcentered & \textperiodcentered & \textperiodcentered\\
\texttt{core-failover-coordinator} & C & -- & Mindustry constructions & \cmark & \textperiodcentered & \cmark & \cmark & \textperiodcentered & \textperiodcentered & \textperiodcentered\\
\texttt{item-reservation-ledger} & C & -- & Mindustry constructions & \cmark & \cmark & \cmark & \cmark & \textperiodcentered & \textperiodcentered & \textperiodcentered\\
\texttt{logic-link-registry} & C & -- & Mindustry constructions & \cmark & \cmark & \cmark & \cmark & \textperiodcentered & \textperiodcentered & \textperiodcentered\\
\texttt{power-grid-controller} & C & -- & Mindustry constructions & \cmark & \textperiodcentered & \textperiodcentered & \cmark & \textperiodcentered & \textperiodcentered & \textperiodcentered\\
\texttt{transactional-core-router} & C & -- & Mindustry constructions & \textperiodcentered & \textperiodcentered & \textperiodcentered & \textperiodcentered & \textperiodcentered & \textperiodcentered & \textperiodcentered\\
\texttt{unit-lease-scheduler} & C & -- & Mindustry constructions & \textperiodcentered & \textperiodcentered & \textperiodcentered & \textperiodcentered & \textperiodcentered & \textperiodcentered & \textperiodcentered\\
\texttt{wave-budget-planner} & C & -- & Mindustry constructions & \cmark & \cmark & \cmark & \cmark & \cmark & \cmark & \textperiodcentered\\
\texttt{wave-threat-arbiter} & C & -- & Mindustry constructions & \textperiodcentered & \textperiodcentered & \textperiodcentered & \textperiodcentered & \textperiodcentered & \textperiodcentered & \textperiodcentered\\
\midrule
\multicolumn{11}{@{}l}{\textit{Palworld}}\\
\texttt{base-containment-mincut} & C & -- & Palworld constructions & \cmark & \textperiodcentered & \textperiodcentered & \textperiodcentered & \textperiodcentered & \textperiodcentered & \textperiodcentered\\
\texttt{guild-consistent-checkpoint} & C & -- & Palworld constructions & \textperiodcentered & \textperiodcentered & \textperiodcentered & \textperiodcentered & \textperiodcentered & \textperiodcentered & \textperiodcentered\\
\texttt{guild-draft-canonical-order} & C & -- & Palworld constructions & \cmark & \textperiodcentered & \textperiodcentered & \textperiodcentered & \textperiodcentered & \textperiodcentered & \textperiodcentered\\
\texttt{guild-storage-ledger} & C & -- & Palworld constructions & \cmark & \textperiodcentered & \textperiodcentered & \textperiodcentered & \textperiodcentered & \textperiodcentered & \textperiodcentered\\
\texttt{outbox-receipt-recovery} & C & -- & Palworld constructions & \textperiodcentered & \textperiodcentered & \textperiodcentered & \textperiodcentered & \textperiodcentered & \textperiodcentered & \textperiodcentered\\
\texttt{pal-workforce-optimal-matching} & C & -- & Palworld constructions & \textperiodcentered & \textperiodcentered & \textperiodcentered & \textperiodcentered & \textperiodcentered & \textperiodcentered & \textperiodcentered\\
\texttt{raid-dispatch-queue} & C & -- & Palworld constructions & \textperiodcentered & \textperiodcentered & \textperiodcentered & \textperiodcentered & \textperiodcentered & \textperiodcentered & \cmark\\
\bottomrule
\end{tabular}}
\end{table}

\begin{table}[p]
\centering\tiny
\setlength{\tabcolsep}{2pt}
\renewcommand{\arraystretch}{0.95}
\caption{\textbf{Task-level results (continued).}}
\label{tab:task-level-3}
\fitlinewidth{\begin{tabular}{@{}lcclccccccc@{}}
\toprule
Task & Kind & Orig. & Source group & Claude & GPT & Kimi & DS & MiniMax & GLM & Qwen\\
\midrule
\multicolumn{11}{@{}l}{\textit{Palworld}}\\
\texttt{raid-event-watermark} & C & -- & Palworld constructions & \textperiodcentered & \textperiodcentered & \textperiodcentered & \textperiodcentered & \textperiodcentered & \textperiodcentered & \textperiodcentered\\
\texttt{server-rule-ballot} & C & -- & Palworld constructions & \cmark & \textperiodcentered & \textperiodcentered & \textperiodcentered & \textperiodcentered & \textperiodcentered & \textperiodcentered\\
\texttt{world-epoch-fenced-timers} & C & -- & Palworld constructions & \textperiodcentered & \textperiodcentered & \textperiodcentered & \textperiodcentered & \textperiodcentered & \textperiodcentered & \textperiodcentered\\
\midrule
\multicolumn{11}{@{}l}{\textit{The Binding of Isaac}}\\
\texttt{isaac-exodus-active-item-state} & R & H+I & Exodus & \cmark & \cmark & \textperiodcentered & \textperiodcentered & \textperiodcentered & \textperiodcentered & \textperiodcentered\\
\texttt{isaac-exodus-callback-wiring} & R & H+I & Exodus & \textperiodcentered & \cmark & \cmark & \cmark & \textperiodcentered & \textperiodcentered & \textperiodcentered\\
\texttt{isaac-exodus-mixed-triage} & R & H+I & Exodus & \cmark & \cmark & \textperiodcentered & \textperiodcentered & \textperiodcentered & \textperiodcentered & \textperiodcentered\\
\texttt{isaac-exodus-numeric-arithmetic} & R & H+I & Exodus & \cmark & \textperiodcentered & \textperiodcentered & \cmark & \textperiodcentered & \textperiodcentered & \textperiodcentered\\
\texttt{isaac-exodus-orphaned-effects} & R & H+I & Exodus & \cmark & \cmark & \cmark & \cmark & \textperiodcentered & \textperiodcentered & \textperiodcentered\\
\texttt{isaac-exodus-passive-reshapers} & R & H+I & Exodus & \cmark & \cmark & \textperiodcentered & \textperiodcentered & \textperiodcentered & \textperiodcentered & \textperiodcentered\\
\texttt{isaac-exodus-projectiles-and-targets} & R & H+I & Exodus & \textperiodcentered & \textperiodcentered & \cmark & \textperiodcentered & \textperiodcentered & \textperiodcentered & \textperiodcentered\\
\texttt{isaac-exodus-release-regressions} & R & H & Exodus & \cmark & \textperiodcentered & \textperiodcentered & \cmark & \textperiodcentered & \textperiodcentered & \textperiodcentered\\
\texttt{isaac-exodus-room-floor-state} & R & H+I & Exodus & \cmark & \cmark & \textperiodcentered & \cmark & \textperiodcentered & \textperiodcentered & \textperiodcentered\\
\texttt{isaac-exodus-tear-trajectory} & R & H+I & Exodus & \textperiodcentered & \textperiodcentered & \textperiodcentered & \textperiodcentered & \textperiodcentered & \textperiodcentered & \textperiodcentered\\
\texttt{construct-callback-transaction} & C & -- & Isaac constructions & \cmark & \cmark & \textperiodcentered & \textperiodcentered & \textperiodcentered & \textperiodcentered & \textperiodcentered\\
\texttt{construct-config-navigation} & C & -- & Isaac constructions & \cmark & \cmark & \cmark & \cmark & \textperiodcentered & \cmark & \textperiodcentered\\
\texttt{construct-config-persistence} & C & -- & Isaac constructions & \cmark & \cmark & \textperiodcentered & \textperiodcentered & \textperiodcentered & \textperiodcentered & \textperiodcentered\\
\texttt{construct-entity-cleanup} & C & -- & Isaac constructions & \textperiodcentered & \textperiodcentered & \textperiodcentered & \textperiodcentered & \textperiodcentered & \textperiodcentered & \textperiodcentered\\
\texttt{construct-hud-compositor} & C & -- & Isaac constructions & \cmark & \textperiodcentered & \textperiodcentered & \textperiodcentered & \textperiodcentered & \cmark & \textperiodcentered\\
\texttt{construct-legacy-hud-lifecycle} & C & -- & Isaac constructions & \textperiodcentered & \cmark & \textperiodcentered & \textperiodcentered & \textperiodcentered & \textperiodcentered & \textperiodcentered\\
\texttt{construct-option-rendering} & C & -- & Isaac constructions & \textperiodcentered & \textperiodcentered & \textperiodcentered & \textperiodcentered & \textperiodcentered & \cmark & \textperiodcentered\\
\texttt{construct-room-rng} & C & -- & Isaac constructions & \textperiodcentered & \textperiodcentered & \textperiodcentered & \textperiodcentered & \textperiodcentered & \textperiodcentered & \textperiodcentered\\
\midrule
\multicolumn{11}{@{}l}{\textit{Left 4 Dead 2}}\\
\texttt{l4d2-competitive-release-regressions-easy} & R & H$^{M}$ & L4D2 Competitive Rework & \textperiodcentered & \cmark & \textperiodcentered & \cmark & \textperiodcentered & \textperiodcentered & \textperiodcentered\\
\texttt{l4d2-competitive-score-lifecycle-hard} & R & L$^{M}$ & L4D2 Competitive Rework & \cmark & \textperiodcentered & \textperiodcentered & \textperiodcentered & \textperiodcentered & \textperiodcentered & \textperiodcentered\\
\texttt{l4d2-confogl-cvar-transactions-hard} & R & L$^{M}$ & L4D2 Competitive Rework & \cmark & \cmark & \textperiodcentered & \cmark & \textperiodcentered & \textperiodcentered & \textperiodcentered\\
\texttt{l4d2-infected-bots-release-regressions-medium} & R & H$^{M}$ & L4D Infected Bots & \cmark & \cmark & \cmark & \cmark & \cmark & \textperiodcentered & \textperiodcentered\\
\texttt{l4d2-left4dhooks-api-regressions-medium} & R & H & Left4DHooks & \textperiodcentered & \textperiodcentered & \textperiodcentered & \textperiodcentered & \textperiodcentered & \textperiodcentered & \textperiodcentered\\
\texttt{l4d2-library-map-tank-state-hard} & R & L$^{M}$ & L4D2 Competitive Rework & \cmark & \cmark & \cmark & \textperiodcentered & \textperiodcentered & \textperiodcentered & \textperiodcentered\\
\texttt{l4d2-linge-director-state-hard} & R & L & LinGe VScripts & \cmark & \cmark & \cmark & \cmark & \cmark & \textperiodcentered & \textperiodcentered\\
\texttt{l4d2-linge-release-regressions-hard-v3} & R & H+I & LinGe VScripts & \textperiodcentered & \textperiodcentered & \textperiodcentered & \textperiodcentered & \textperiodcentered & \textperiodcentered & \textperiodcentered\\
\texttt{l4d2-mutant-tanks-release-regressions-hard} & R & H+I & Mutant Tanks & \cmark & \cmark & \cmark & \textperiodcentered & \textperiodcentered & \textperiodcentered & \textperiodcentered\\
\texttt{l4d2-vslib-scheduler-persistence-hard} & R & L & LinGe VScripts & \cmark & \cmark & \cmark & \textperiodcentered & \textperiodcentered & \textperiodcentered & \textperiodcentered\\
\midrule
\multicolumn{11}{@{}l}{\textit{Balatro}}\\
\texttt{bounded-generator-v2} & R & I & Balatro & \cmark & \cmark & \cmark & \textperiodcentered & \cmark & \cmark & \cmark\\
\texttt{deck-removal-consistency-v2} & R & I & Balatro & \cmark & \cmark & \cmark & \cmark & \cmark & \cmark & \cmark\\
\texttt{durable-charge-v2} & R & I & Balatro & \cmark & \cmark & \cmark & \cmark & \cmark & \textperiodcentered & \cmark\\
\texttt{growth-v2} & R & I & Balatro & \cmark & \cmark & \cmark & \cmark & \cmark & \textperiodcentered & \textperiodcentered\\
\texttt{isolated-card-wear-v2} & R & I & Balatro & \cmark & \cmark & \cmark & \cmark & \textperiodcentered & \cmark & \cmark\\
\texttt{live-joker-adjacency-v2} & R & I & Balatro & \cmark & \cmark & \cmark & \cmark & \cmark & \cmark & \cmark\\
\texttt{pack-choice-commit-v2} & R & I & Balatro & \textperiodcentered & \cmark & \textperiodcentered & \textperiodcentered & \cmark & \textperiodcentered & \cmark\\
\texttt{reversible-hand-limit-v1} & R & I & Balatro & \cmark & \cmark & \cmark & \cmark & \cmark & \cmark & \cmark\\
\texttt{seeded-effect-stream-v1} & R & I & Balatro & \cmark & \cmark & \cmark & \textperiodcentered & \cmark & \cmark & \textperiodcentered\\
\texttt{shop-price-coherence-v2} & R & I & Balatro & \cmark & \cmark & \textperiodcentered & \textperiodcentered & \textperiodcentered & \textperiodcentered & \cmark\\
\midrule
\multicolumn{11}{@{}l}{\textit{Slay the Spire}}\\
\texttt{basemod-dynamic-descriptions-r2} & R & L & BaseMod & \textperiodcentered & \textperiodcentered & \textperiodcentered & \textperiodcentered & \textperiodcentered & \textperiodcentered & \textperiodcentered\\
\texttt{basemod-phase-transitions-r1} & R & L$^{M}$ & BaseMod & \cmark & \cmark & \cmark & \cmark & \textperiodcentered & \textperiodcentered & \textperiodcentered\\
\texttt{downfall-socket-flow-r3} & R & L$^{M}$ & Downfall & \cmark & \textperiodcentered & \textperiodcentered & \cmark & \textperiodcentered & \textperiodcentered & \textperiodcentered\\
\texttt{downfall-stasis-lifecycle-r1} & R & L$^{M}$ & Downfall & \cmark & \textperiodcentered & \textperiodcentered & \cmark & \textperiodcentered & \textperiodcentered & \textperiodcentered\\
\texttt{mts-enum-identity-r3} & R & L & ModTheSpire & \cmark & \cmark & \textperiodcentered & \cmark & \textperiodcentered & \textperiodcentered & \textperiodcentered\\
\texttt{mts-load-order-regressions-r2} & R & H+I & ModTheSpire & \cmark & \cmark & \cmark & \cmark & \textperiodcentered & \textperiodcentered & \cmark\\
\texttt{mts-mod-identity-regressions-r2} & R & I & ModTheSpire & \cmark & \cmark & \cmark & \cmark & \cmark & \textperiodcentered & \textperiodcentered\\
\texttt{stslib-block-ledger-r1} & R & L$^{M}$ & StSLib & \cmark & \cmark & \cmark & \textperiodcentered & \textperiodcentered & \textperiodcentered & \textperiodcentered\\
\texttt{stslib-icon-layout-r1} & R & L$^{M}$ & StSLib & \textperiodcentered & \textperiodcentered & \textperiodcentered & \textperiodcentered & \textperiodcentered & \textperiodcentered & \textperiodcentered\\
\bottomrule
\end{tabular}}
\end{table}

\section{Coding of Unresolved Attempts}
\label{app:failure-coding}

All 617 unresolved attempts are coded automatically with the rules below. The scripts and the per-attempt codes are released with the benchmark.

\paragraph{Failure stage.}
An attempt is \emph{no change} if the submission equals the initial code; \emph{before tests} if no F2P or P2P test is observed, subdivided into build or static-check failures (a failed build, required-files, syntax, or layout check, or an adjudication note naming the build) and load failures (the game starts but the mod fails to load or the test suite does not start); \emph{F2P} if at least one audited F2P test fails; and \emph{after F2P} otherwise (for example, a failed P2P test, or a load or check failure after all F2P tests pass). The stages partition the unresolved attempts (Table~\ref{tab:failure-stages}). Within the F2P stage, 180 attempts fail at most two F2P tests while passing at least as many as they fail, 54 fail every F2P test, 25 report a mod error with most tests unobserved, and 130 fall in between.

\paragraph{Build failures.}
For the 89 attempts that stop at the build, we take the first compiler error from the runtime log, or from the adjudication note for RimWorld and Terraria, whose verifiers do not write compiler output to that log, and classify it by its error code and message (Table~\ref{tab:build-errors}).

\begin{table}[!ht]
\centering\footnotesize
\caption{\textbf{First error of the 89 build failures.} Classified by compiler error code and message, for example \texttt{cannot find symbol} (javac), CS0246 and CS1061 (C\#), and error 017 (SourcePawn) as an undefined symbol.}
\label{tab:build-errors}
\begin{tabular}{@{}lr@{}}
\toprule
First error & Attempts\\
\midrule
Symbol, type, or member not defined by the pinned host or library & 45\\
Syntax or semantic error & 11\\
Incompatible types & 9\\
Compilation failed, compiler output not retained & 8\\
Signature or override mismatch & 7\\
Server crashed during the build stage (no compiler error) & 3\\
Evaluation timed out without a verdict & 2\\
Other or unclassified & 4\\
\midrule
All & 89\\
\bottomrule
\end{tabular}
\end{table}

\paragraph{Termination and final report.}
\emph{Submitted} is the agent scaffold's exit status: the agent issued the submission command that the prompt asks for once the implementation is complete (Appendix~\ref{app:prompts}). Other exit statuses are the step limit (94 unresolved attempts), a repeated format error (41), the time limit (31), and no final action (19). \emph{Claims pass} applies only to unresolved attempts that change the code and end with the submission command. The final assistant message is split into sentences, and an attempt is flagged if any sentence matches one of the following patterns (case-insensitive) and contains no negation such as ``could not'', ``unable to'', ``not verified'', ``blocked'', or ``not run'': \emph{all} or \emph{every} followed within 60 characters by \emph{pass}, \emph{succeed}, or \emph{green}; \emph{check}, \emph{test}, \emph{diagnostic}, \emph{verification}, \emph{validation}, \emph{harness}, \emph{syntax}, \emph{compile}, \emph{lint}, or \emph{suite} followed within 40 characters by \emph{pass}, \emph{clean}, \emph{succeed}, \emph{ok}, or \emph{green}; \emph{verified} or \emph{validated}; \emph{passes my/the/all/every/our \ldots{} checks/tests}; or \emph{compiles cleanly/successfully/without}.

\section{Task Examples}
\label{app:task-examples}

This appendix describes the four tasks used as examples in Section~\ref{sec:benchmark}.

\subsection{Automate: returning missing reports through the existing API}

Automate is a Stardew Valley mod that lets machines, such as furnaces, exchange items with nearby storage chests. Another mod, the Data Layers overlay, reads machine states through Automate's API to show players the status of each machine. In the repair task, Automate returns no reports for a location in which no machine is connected to storage. The requested fix adds these reports while keeping the reports for machines that are already connected. The tests query the API immediately after the game loads and expect a nonempty result, then place a chest next to a furnace and expect a report for the connected furnace together with the other reports. A correct fix changes what the existing API returns; adding a separate reporting function would leave the data that the overlay receives unchanged. Data Layers itself is not part of the task's environment $M$: the tests call the API that the overlay reads, so the task is not among the 20 tasks that load third-party mods (Table~\ref{tab:unc-thirdparty}). The co-loading pilot (Appendix~\ref{app:supp}) does load Data Layers beside every Stardew Valley submission, and for the six Automate tasks, the 38 pairs in which all three mods are confirmed loaded have identical test results with and without the mods.

\subsection{Chatty: keeping the command registry consistent after reload}

Chatty is a Minecraft plugin for the Paper server that defines chat commands in its configuration. In the repair task, reloading an edited configuration leaves the server's command registry out of date. After a correct fix, the plugin registers newly configured commands, removes the names of deleted or renamed commands, and keeps unchanged commands and their aliases working. The F2P tests edit the configuration, reload the plugin on the Paper server, and check which commands the server registers for the plugin; the P2P tests check that the console can still dispatch the unchanged commands.

\subsection{Blueprint: restored objects that work with Undo}

Blueprint is a Garry's Mod construction task that starts from an empty addon workspace. The feature request asks for saving and restoring a contraption, a group of game objects together with the physical constraints that connect them, and requires the restored objects to work with the game's ordinary Undo action. The test restores a contraption of three entities and two constraints, invokes Undo, and checks that all restored objects are removed. Because the initial workspace is empty, the task contains F2P tests only.

\subsection{Loadout: equipment profiles across death and restart}
\label{app:loadout}

Loadout is a Garry's Mod construction task whose description is written in Chinese. The requested addon consists of a manifest, a profile store, and an adapter that connects the store to the game server. The store keeps named equipment profiles for each player, supports a preferred profile, assigns revision numbers that only increase, and returns the same result when a request is repeated. Applying a profile removes the player's weapons, clears their ammunition, grants the profile's equipment and ammunition, and selects a weapon. Reapplying a profile does not accumulate pickups, and if a weapon cannot be granted, the adapter stops before granting ammunition.

The saved state contains logical player identities and profile data rather than game objects, so that a profile can be restored for a new player object. The main fixture runs in two dedicated-server processes. In the first, a bot receives a profile with 60 rounds of primary and 10 rounds of secondary ammunition, which combine into a 70-round plan; the bot then picks up 50 additional rounds, is killed, and respawns, and the test checks that its ammunition returns to 70 rounds. After the profile is saved, a second server process binds a new bot to the same logical identity and checks that the 70-round plan is restored. Further fixtures cover disconnecting and reconnecting a player.

\section{Verification Details}
\label{app:verification}

This appendix specifies how the live-runtime harness determines test and task outcomes and how it handles failures unrelated to the submission.

\paragraph{Test outcomes.}
Each test ends in one of three states. It \emph{passes} when its checks succeed, \emph{fails} when a check is violated, and is \emph{unobserved} when execution stops before the test runs, for example because the mod crashes. We keep unobserved tests separate from failed ones, so that a single crash does not appear as many independent failures.

\paragraph{Task outcomes.}
A task is resolved only if the submission builds and loads, every F2P and P2P test passes, and the run terminates cleanly, including the required cleanup. A failed test, a crash caused by the submission, or a failure to build or load leaves the task unresolved. We additionally record the F2P and P2P outcomes separately: each is \emph{passed} when all of its tests pass, \emph{failed} when at least one of its tests fails, and \emph{unknown} otherwise, as when a build failure prevents any test from running.

\paragraph{Failures unrelated to the submission.}
An execution can also fail for reasons outside the submission, such as a missing runtime or a faulty fixture. If no test has failed because of the submission, such an execution is marked invalid and rerun once with the same code, runtime, initial state, seed, and limits. A valid failure is never replaced by a later passing run.

\paragraph{Versions.}
Each task pins the game build, the revision of the target mod, the third-party mods, and the harness. Reported results refer to a specific release of \benchmark{}, and a change to any pinned component produces a new task version.

\section{Benchmark Comparison Details}
\label{app:closest-work}

This appendix describes how we construct the comparison in Table~\ref{tab:closest-work}.

\paragraph{Sources.}
We use the ICLR 2024 version of SWE-bench, version 1 of FeatureBench and GameXpert-Bench, version 2 of GameDevBench and GameEngineBench, the cited arXiv version of GameCraft-Bench, and the pinned software release of GameForgeBench, listed as GameForge in Table~\ref{tab:closest-work}.

\paragraph{Released host and live-runtime verification.}
A benchmark extends a \emph{released host} if its code changes target a shipped program that the task authors do not own, and it is \emph{live-runtime verified} if its tests run the code inside that program. At the cited commit, GameForgeBench contains 17 task packages: 10 Godot projects, 4 Minecraft Paper plugins, 2 C++ subsystems, and 1 browser game. Its Paper plugins extend a released host, and two of them are built and verified on a live Paper server through a protocol client and a verifier plugin, while the other packages run engine projects, standalone probes, or a headless browser; we therefore mark both properties as partial. GameDevBench and GameEngineBench execute code in a game engine, but the engine project is the agent's own development target rather than a released program that the code extends.

\paragraph{Third-party components.}
A benchmark loads \emph{third-party components} if independently developed components share the runtime with the submission during verification. We find no task in the public GameForgeBench release that loads independently developed plugins alongside the candidate.

\paragraph{Evaluation practices of the closest benchmarks.}
SWE-bench combines fail-to-pass tests of the requested change with pass-to-pass tests of previously working functionality~\citep{swebench}; \benchmark{} adopts the same distinction. FeatureBench evaluates feature implementation with and without the surrounding repository and separates partial from full success~\citep{featurebench}. GameEngineBench asks agents to edit C++ in buildable Unreal Engine projects and evaluates the results through tests and an LLM judge~\citep{gameenginebench}. GameXpert-Bench counts a bug as fixed only when both its repair and associated preservation tests pass~\citep{gamexpertbench}. GameForgeBench also checks existing functionality; for example, a Godot status-bar task requests a new visual style while requiring the bars' layout and value computation to stay unchanged~\citep{gameforgebench}.

\section{Extended Related Work}
\label{app:extended-related}

This appendix relates \benchmark{} to interactive testing, task construction, and evaluation validity.

\subsection{Interactive testing and game-playing agents}

PlayCoder couples repository-aware GUI code generation with automated playthroughs and iterative repair~\citep{playcoder}, and GBQA studies bug discovery through interaction with games that contain injected defects~\citep{gbqa}. Both connect code quality to what happens in a running game, a view that \benchmark{} shares. Game environments also serve a different purpose: Voyager builds reusable code skills for exploring Minecraft~\citep{voyager}, and StarDojo evaluates open-ended agent behavior in Stardew Valley~\citep{stardojo}. In these works the agent plays the game, while in \benchmark{} the agent extends it.

\subsection{Constructing executable tasks}

SWE-Gym provides executable repository environments for training agents~\citep{swegym}, and SWE-rebench automates the collection and validation of issue-based tasks~\citep{swerebench}. R2E-Gym generates tests and derives problem statements from code~\citep{r2egym}, SWE-smith combines programmatic and model-generated modifications with pull-request inversion~\citep{swesmith}, SWE-Mirror transfers issues between repositories through separate test and defect generation~\citep{swemirror}, and SWE-Playground generates projects and tasks from interface scaffolds with injected bugs~\citep{sweplayground}. These approaches scale task collection through automation. In \benchmark{}, tests are written by hand and every task is validated in the live game runtime and through human review, trading scale for verification that reflects how mods are actually used.

\subsection{Evaluation validity}

A passing test suite can leave incorrect code undetected. Program-repair research uses additional tests to expose patches that overfit the original suite~\citep{difftgen}, EvalPlus shows that stronger tests reveal incorrect code accepted by weaker ones~\citep{evalplus}, an empirical study finds that some patches accepted by SWE-bench's tests are incorrect~\citep{solvedissues}, and SWE-Bench Pro Verified addresses leakage and inconsistencies between task descriptions and tests~\citep{swebenchproverified}. Live-runtime verification addresses a complementary gap, conditions that the test environment does not reproduce, while the validation in \benchmark{}, which combines reference solutions, the initial code, and human review in the game, reduces the risk of incomplete tests.

\newcommand{\casebodyfont}{\tiny}
\newcommand{\casetitlefont}{\fontsize{7}{8.4}\selectfont}
\newcommand{\casepillfont}{\fontsize{8.5}{10}\selectfont}
\newcommand{\caselogfont}{\fontsize{5.6}{6.6}\selectfont}
\newcommand{\casecapskip}{7pt}
\newcommand{\caseafterskip}{4pt}
\newcommand{\casecaptionsetup}{\captionsetup[figure]{font=footnotesize}}
\definecolor{osgrey}{HTML}{F2F3F5}
\definecolor{osframe}{HTML}{7C7F86}
\definecolor{oscream}{HTML}{FFF8E8}
\definecolor{osblue}{HTML}{D9E4F8}
\definecolor{ospeach}{HTML}{FAD6C8}
\definecolor{ospassbg}{HTML}{E2F1DF}
\definecolor{ospassfg}{HTML}{3F8F3F}
\definecolor{osfailbg}{HTML}{FBE2E0}
\definecolor{osfailfg}{HTML}{D1342F}
\providecommand{\casebodyfont}{\scriptsize}
\providecommand{\casetitlefont}{\footnotesize}
\providecommand{\casepillfont}{\small}
\providecommand{\caselogfont}{\fontsize{6.2}{7.4}\selectfont}
\providecommand{\casecapskip}{9pt}
\providecommand{\caseafterskip}{10pt}
\providecommand{\casecaptionsetup}{}
\tcbset{oscard/.style={enhanced,arc=2mm,boxrule=0.6pt,colframe=osframe,colback=osgrey,coltitle=black,
  drop fuzzy shadow={black!45},left=5pt,right=5pt,top=2pt,bottom=3pt,before skip=2pt,after skip=4pt,
  fontupper=\casebodyfont,halign=flush left,detach title,before upper={{\sffamily\bfseries\casetitlefont\tcbtitle}\par\vspace{1pt}}},
  runin/.style={before upper={{\sffamily\bfseries\casetitlefont\tcbtitle:}~}}}
\newtcolorbox{casecard}[2][]{oscard,title={#2},#1}
\newtcolorbox{caseprompt}[2][]{oscard,colback=white,title={#2},fontupper=\casebodyfont\ttfamily\bfseries,halign=flush left,#1}
\newcommand{\casepill}[2]{\tikz[baseline=(n.base)]\node[fill=#1,rounded corners=4pt,inner xsep=7pt,inner ysep=2.5pt,
  font=\sffamily\casepillfont](n){\strut #2};}
\newcommand{\casechip}[3]{\tikz[baseline=(n.base)]\node[fill=#1,text=#2,rounded corners=1.5pt,inner xsep=3pt,
  inner ysep=1.2pt,font=\sffamily\bfseries\scriptsize](n){\strut #3};}
\newcommand{\casepass}[1]{\casechip{ospassbg}{ospassfg}{#1}}
\newcommand{\casefailchip}[1]{\casechip{osfailbg}{osfailfg}{#1}}
\newcommand{\casehead}[2]{\par\noindent{\raggedright#1\enspace{\sffamily\casepillfont #2}\par}\nobreak\vspace{2pt}}
\newcommand{\casefield}[2]{{\sffamily\bfseries #1:}~#2\par}
\newcommand{\casesub}[1]{\par\vspace{2pt}{\sffamily\bfseries #1}\par\vspace{1pt}}
\newcommand{\caserow}[4]{\par\noindent
  \begin{minipage}[t]{0.30\linewidth}\vspace{0pt}\begin{casecard}[equal height group=#1,before skip=0pt,after skip=0pt]{Task}#2\end{casecard}\end{minipage}\hfill
  \begin{minipage}[t]{0.68\linewidth}\vspace{0pt}\begin{casecard}[colback=oscream,equal height group=#1,before skip=0pt,after skip=0pt]{#3}#4\end{casecard}\end{minipage}\par\vspace{4pt}}
\newcommand{\caseshotlabel}[1]{\par{\casebodyfont\sffamily\bfseries #1}\par\vspace{2pt}}
\newcommand{\caseshot}[1]{{\setlength{\fboxsep}{0pt}\setlength{\fboxrule}{0.4pt}\fcolorbox{osframe}{white}{\includegraphics[width=\dimexpr\linewidth-0.8pt\relax]{#1}}}}
\newcommand{\casecapgap}{\par\vspace{\casecapskip}}
\newcommand{\caselog}[1]{{\caselogfont\ttfamily\raggedright #1\par}}
\newcommand{\caseback}{\par\vspace{1pt}{\centering\scriptsize\hyperlink{app:case-index-list}{Back to the index of case studies}\par}}

\section{Agent Prompts}
\label{app:prompts}
Each attempt receives a system prompt and a single task message (Figure~\ref{fig:prompts}). The system prompt is fixed per task pack. It typically names the game and the kind of task, forbids looking up upstream fixes or benchmark internals, and states that the private evaluator runs once after submission; some packs add host-specific guidance, as in the Don't Starve Together example. The task message wraps the task description in the same template for every attempt of the study. Both are quoted verbatim from the attempt records.
\begin{center}
\casehead{\casepill{osblue}{Prompts given to the agent}}{verbatim}
\begin{caseprompt}{System prompt, Slay the Spire repair tasks}
You are the model under test in a Slay the Spire mod repository-repair benchmark. The workspace holds a real mod's source tree at a fixed commit, carrying regressions reported only as user-visible symptoms. Work only in the supplied project and resolve every reported problem at its cause. Do not inspect or search for upstream fixes, benchmark internals, private tests, reference solutions, validation records, or prior results. A private verifier builds the submission offline and checks it inside the real game, without network access. The benchmark platform will run the private evaluator exactly once after you submit. Do not invoke score-exam, run-attempt, or any benchmark scorer yourself.
\end{caseprompt}
\begin{caseprompt}{System prompt, Don't Starve Together construction tasks}
You are the model under test in a Don't Starve Together engineering benchmark. The workspace contains only the public task contract and an empty starter; implement the requested mod from scratch. Keep plain-Lua state machines independent of DST globals, keep authority on the master simulation, and use the documented base-game API signatures exactly. You cannot see private Lua or dedicated-server assertions. Before submitting, inspect every required file and run \texttt{python -\allowbreak{}m rhenium\_\allowbreak{}dst.\allowbreak{}public\_\allowbreak{}check .\allowbreak{}} for public syntax diagnostics. The benchmark platform will run the private evaluator exactly once after you submit. Do not invoke score-exam, run-attempt, or any benchmark scorer yourself.
\end{caseprompt}
\begin{caseprompt}{Task message (all attempts)}
Implement the complete task below from scratch.\allowbreak{}\\{}\mbox{}\\{}\{task description\}\\{}\mbox{}\\{}Use shell commands to inspect and edit the current workspace.\allowbreak{} Every response must include at least\\{}one bash tool call.\allowbreak{} Directory changes do not persist between calls,\allowbreak{} but every call starts in the\\{}workspace root.\allowbreak{} When the implementation is complete,\allowbreak{} run exactly this command by itself:\allowbreak{}\\{}\mbox{}\\{}~~~~echo COMPLETE\_\allowbreak{}TASK\_\allowbreak{}AND\_\allowbreak{}SUBMIT\_\allowbreak{}FINAL\_\allowbreak{}OUTPUT
\end{caseprompt}
\casecapgap\captionof{figure}{\textbf{Agent prompts.} Two of the per-pack system prompts and the task-message template; \texttt{\{task description\}} is replaced by the task's description, such as the excerpts in Appendix~\ref{app:cases}.}\label{fig:prompts}
\end{center}
\casecaptionsetup
\section{Case Studies}
\label{app:cases}
This appendix examines 22 primary cases from the E1 run behind Table~\ref{tab:main_results}, two in each of 11 of the 12 games, and one additional failure example (Figure~\ref{fig:case-balatro-crash}). In nine of these games the two attempts address the same task, one resolved and one unresolved, so that they share the same task and evaluation environment. Each figure shows an excerpt of the task description, the agent's final report, the verifier's record, and the verdict, followed by our analysis. Task excerpts, agent reports, and verifier records are quoted verbatim from the attempt records; ``\ldots'' marks an elision, and long hexadecimal run identifiers are shortened. F2P and P2P counts refer to the audited fail-to-pass and pass-to-pass tests of the frozen case table; static checks and runtime sentinels are not counted. The records are the verifier's output (check messages, harness logs, and the game's own log), and where marked, a line of the submitted code or the note of the manual review that attributed a run that stopped without a verdict to the submission. The case studies begin with Balatro, The Binding of Isaac, and Stardew Valley, whose figures also show frames captured from the game's virtual display while the verifier ran the same frozen submissions again; these re-runs reproduce the E1 verdicts. The dedicated servers of six of these games draw nothing, and the Slay the Spire and RimWorld verifiers run all of their assertions within a single update, before the tested scene is ever drawn, so frames cannot show those tests. Table~\ref{tab:case-index} indexes the figures by game and outcome, using the failure categories of Section~\ref{sec:taxonomy}, and Figure~\ref{fig:case-seven} compares all seven models on a single task.
\begin{table}[!ht]\centering\scriptsize\setlength{\tabcolsep}{3.5pt}
\caption{\textbf{Index of the case studies.} Each entry is a figure number; rows are games and columns are the outcome of the case (Section~\ref{sec:taxonomy}). Figure~\ref{fig:case-seven} compares all seven models on one task.}
\label{tab:case-index}
\hypertarget{app:case-index-list}{}
\begin{tabular}{@{}lccccc@{}}
\toprule
Game & \begin{tabular}[b]{@{}c@{}}Resolved\end{tabular} & \begin{tabular}[b]{@{}c@{}}Interface the\\ecosystem lacks\end{tabular} & \begin{tabular}[b]{@{}c@{}}Host\\convention\end{tabular} & \begin{tabular}[b]{@{}c@{}}Behavior or\\lifecycle in game\end{tabular} & \begin{tabular}[b]{@{}c@{}}Breaks existing\\behavior\end{tabular}\\
\midrule
Balatro & \ref{fig:case-balatro-ok} & \ref{fig:case-balatro-fail}, \ref{fig:case-balatro-crash} & -- & -- & --\\
The Binding of Isaac & \ref{fig:case-isaac-ok} & -- & -- & \ref{fig:case-isaac-fail} & --\\
Stardew Valley & \ref{fig:case-stardew-ok} & -- & -- & \ref{fig:case-stardew-fail} & --\\
Minecraft & \ref{fig:case-minecraft-ok} & -- & -- & \ref{fig:case-minecraft-fail} & --\\
Garry's Mod & \ref{fig:case-gmod-ok} & -- & \ref{fig:case-gmod-fail} & -- & --\\
Don't Starve Together & \ref{fig:case-dst-ok} & \ref{fig:case-dst-fail} & -- & -- & --\\
Terraria & \ref{fig:case-terraria-ok} & -- & -- & \ref{fig:case-terraria-fail} & --\\
Palworld & \ref{fig:case-palworld-ok} & -- & \ref{fig:case-palworld-fail} & -- & --\\
Left 4 Dead 2 & \ref{fig:case-l4d2-ok} & \ref{fig:case-l4d2-fail} & -- & -- & --\\
Slay the Spire & \ref{fig:case-sts-ok} & -- & -- & -- & \ref{fig:case-sts-fail}\\
RimWorld & \ref{fig:case-rimworld-ok} & \ref{fig:case-rimworld-fail} & -- & -- & --\\
\bottomrule\end{tabular}\end{table}

\begin{center}
\casehead{\casepill{osblue}{All seven models}}{Slay the Spire\,$\cdot$\,\texttt{stslib-block-ledger-r1}\,$\cdot$\,five outcomes}
\begin{tcolorbox}[oscard,colback=white,before upper={}]
\scriptsize
\textsf{\textbf{Task (repair, excerpt).}} \itshape ``Repair block accounting for players and monsters. \ldots{} Block created by a removal callback is a subsequent gain and must not be consumed retroactively by the triggering loss. \ldots{} Preserve public APIs, ordinary partial/complete losses, attack damage and existing removal-based damage reduction.''\upshape\par\vspace{3pt}
\setlength{\tabcolsep}{3pt}
\begin{tabular}{@{}l r l c c l p{5.0cm}@{}}
\toprule
Model & Steps & Build / load & F2P & P2P & Outcome & Final report (excerpt, verbatim)\\
\midrule
GPT-6 Astra & 17 & yes / yes & 20/20 & 6/6 & \cmark\ resolved & \emph{``Compilation and game verification remain blocked here because Maven, Java, and the game libraries are unavailable.''}\\
Claude Opus 5.5 & 22 & yes / yes & 20/20 & 6/6 & \cmark\ resolved & \emph{``I've finished the review. Submitting now.''}\\
Kimi K3.1 Pro & 33 & yes / yes & 20/20 & 6/6 & \cmark\ resolved & \emph{``The only unavailable check is the Maven build because this container has no \texttt{mvn}, Java runtime, compiler, or supplied game jars.''}\\
DeepSeek-V4.1-Flash & 57 & yes / yes & 20/20 & 4/6 & \xmark\ breaks P2P & \emph{``\ldots{} attack damage and removal-based damage reduction are unchanged \ldots{} all scenarios pass.''}\\
MiniMax M3.1 Flash & 56 & yes / crash & -- & -- & \xmark\ unobserved & (no closing summary)\\
GLM 5.3 Flash & -- & no change & 0/20 & 6/6 & \xmark\ no change & (no closing summary)\\
Qwen 3.8 27B & 30 & yes / yes & 4/20 & 0/6 & \xmark\ broad failure & \emph{``I'm confident the implementation is complete and correct. Submitting now.''}\\
\bottomrule
\end{tabular}\par\vspace{3pt}
\textsf{\textbf{Verifier record (verbatim, run identifiers shortened).}}\par
\caselog{Claude Opus 5.\allowbreak{}5:\allowbreak{}~~~~~~BLOCK\_\allowbreak{}SETTLEMENT\_\allowbreak{}SCOUT 1b29.\allowbreak{}.\allowbreak{}.\allowbreak{}\textbar{}\allowbreak{}case,\allowbreak{}id=\allowbreak{}monster-\allowbreak{}attenuate,\allowbreak{}error=\allowbreak{}none,\allowbreak{}block=\allowbreak{}9,\allowbreak{}stacks=\allowbreak{}C:\allowbreak{}9;\allowbreak{},\allowbreak{}otherBlock=\allowbreak{}10,\allowbreak{}otherStacks=\allowbreak{}B:\allowbreak{}10;\allowbreak{},\allowbreak{}hpLost=\allowbreak{}0,\allowbreak{}events=\allowbreak{}A/\allowbreak{}13;\allowbreak{}B/\allowbreak{}0;\allowbreak{}\\{}DeepSeek-\allowbreak{}V4.\allowbreak{}1-\allowbreak{}Flash:\allowbreak{}~~BLOCK\_\allowbreak{}SETTLEMENT\_\allowbreak{}SCOUT 9cc1.\allowbreak{}.\allowbreak{}.\allowbreak{}\textbar{}\allowbreak{}case,\allowbreak{}id=\allowbreak{}monster-\allowbreak{}attenuate,\allowbreak{}error=\allowbreak{}none,\allowbreak{}block=\allowbreak{}0,\allowbreak{}stacks=\allowbreak{},\allowbreak{}otherBlock=\allowbreak{}10,\allowbreak{}otherStacks=\allowbreak{}B:\allowbreak{}10;\allowbreak{},\allowbreak{}hpLost=\allowbreak{}0,\allowbreak{}events=\allowbreak{}A/\allowbreak{}13;\allowbreak{}B/\allowbreak{}0;\allowbreak{}C/\allowbreak{}0;\allowbreak{}\\{}MiniMax M3.\allowbreak{}1 Flash:\allowbreak{}~~ [\allowbreak{}INFO]\allowbreak{} BUILD SUCCESS .\allowbreak{}.\allowbreak{}.\allowbreak{} java.\allowbreak{}lang.\allowbreak{}NullPointerException\\{}~~~~at com.\allowbreak{}evacipated.\allowbreak{}cardcrawl.\allowbreak{}modthespire.\allowbreak{}lib.\allowbreak{}SpireField.\allowbreak{}get(\allowbreak{}SpireField.\allowbreak{}java:\allowbreak{}43)\allowbreak{}\\{}~~~~at com.\allowbreak{}evacipated.\allowbreak{}cardcrawl.\allowbreak{}mod.\allowbreak{}stslib.\allowbreak{}patches.\allowbreak{}BindingPatches\$CaptureBlockBeforeGain.\allowbreak{}capture(\allowbreak{}BindingPatches.\allowbreak{}java:\allowbreak{}196)\allowbreak{}\\{}~~~~at com.\allowbreak{}megacrit.\allowbreak{}cardcrawl.\allowbreak{}core.\allowbreak{}AbstractCreature.\allowbreak{}addBlock(\allowbreak{}AbstractCreature.\allowbreak{}java)\allowbreak{}}
\vspace{2pt}
{\scriptsize\sffamily\bfseries Analysis:}~{\scriptsize The seven submissions for one StSLib repair end in five different ways. The three resolved submissions change only the accounting of block layers. DeepSeek-V4.1-Flash passes every F2P test, but in its rewrite of block loss the 9-point layer \texttt{C} that a removal callback creates is debited to zero by the same hit (event \texttt{C/0}), although the task requires such block to survive; only the two \texttt{attenuate} P2P tests detect this (Figure~\ref{fig:case-sts-fail}). MiniMax M3.1 Flash's submission builds, but its new patch \texttt{CaptureBlockBeforeGain} throws a \texttt{NullPointerException} from \texttt{SpireField.get} on the first \texttt{addBlock} in the running game, so no test is observed; the patch takes effect only once ModTheSpire weaves it into the game. GLM 5.3 Flash leaves the code unchanged, and Qwen 3.8 27B fails 22 of 26 tests, including ordinary block loss. DeepSeek-V4.1-Flash and Qwen 3.8 27B both state in their final reports that the existing behavior is preserved, which is exactly what their failing P2P tests check.}\par
\end{tcolorbox}
\casecapgap\captionof{figure}{\textbf{One task, seven agents.} The E1 attempts of all seven models at the Slay the Spire repair \texttt{stslib-block-ledger-r1}. F2P/P2P give passed/total tests; ``--'' means that no test was observed.}\label{fig:case-seven}
\end{center}

\clearpage
\subsection{Balatro}
\label{app:case-balatro}
\par\noindent\begin{minipage}{\linewidth}
\casehead{\casepill{osblue}{Resolved}}{Balatro\,$\cdot$\,\texttt{pack-\allowbreak{}choice-\allowbreak{}commit-\allowbreak{}v2}\,$\cdot$\,GPT-6 Astra}
\caserow{case-balatro-ok}{%
\casefield{task\_id}{\texttt{pack-\allowbreak{}choice-\allowbreak{}commit-\allowbreak{}v2}}
\casefield{Game}{Balatro}
\casefield{Type}{repair}
\casefield{Outcome}{Resolved}}{Task instruction (repair, excerpt)}{%
The ordinary \texttt{p\_\allowbreak{}pc2\_\allowbreak{}choices} Booster costs \$4 and presents three physical Common Jokers, in order: \texttt{j\_\allowbreak{}pc2\_\allowbreak{}copper}, \texttt{j\_\allowbreak{}pc2\_\allowbreak{}silver}, \texttt{j\_\allowbreak{}pc2\_\allowbreak{}gold}. Choose at most two through normal game actions. Their signing rewards are respectively \$3, \$5 and \$7. ... Opening, creating, displaying or destroying an unchosen preview gives no reward. Each actual selected-and-acquired physical object commits its own reward exactly once.}
\caseshotlabel{In-game capture during the verifier run}
{\centering \begin{minipage}[t]{0.485\linewidth}\centering\caseshot{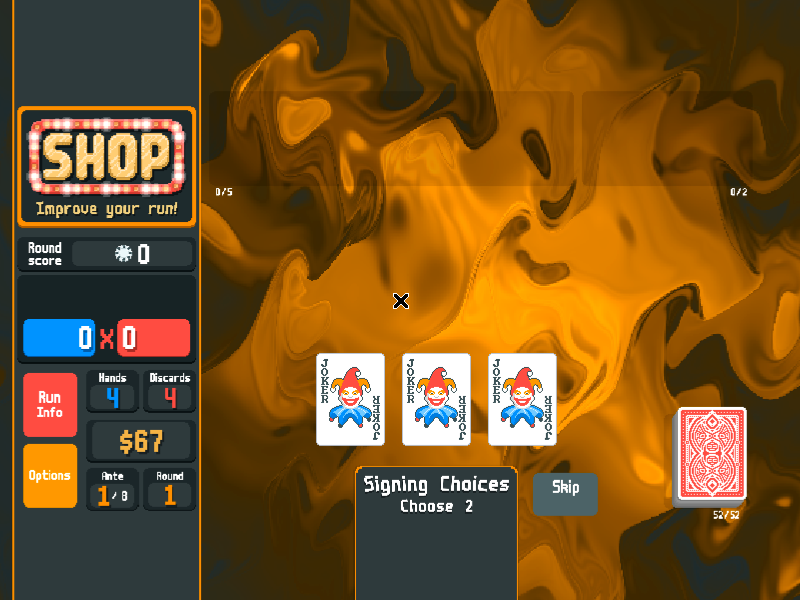}\\[1pt]{\scriptsize Booster opened: the three signing Jokers, \$67.}\end{minipage}\hfill\begin{minipage}[t]{0.485\linewidth}\centering\caseshot{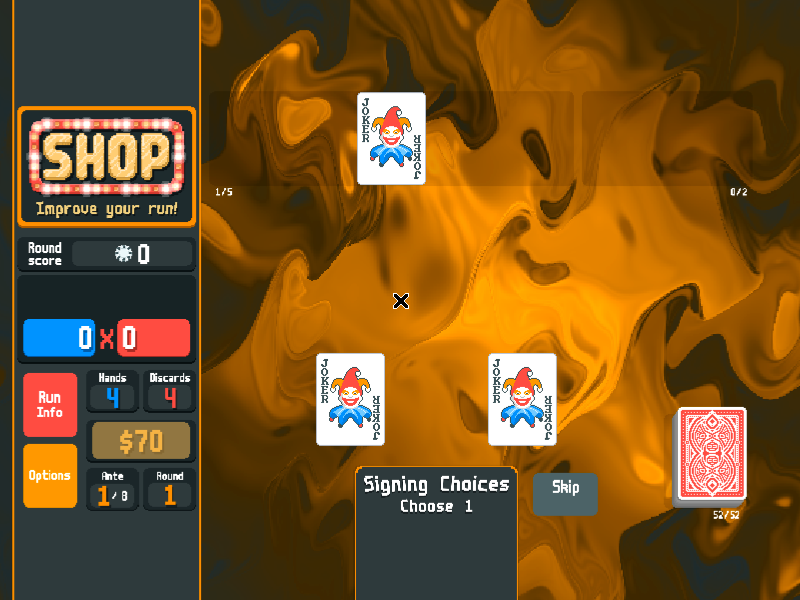}\\[1pt]{\scriptsize First choice taken: one Joker in the Joker row, \$70.}\end{minipage}\par}
\begin{casecard}[colback=white,runin]{Agent's final report (verbatim)}
\itshape Implementation and local checks are complete. All four editable Lua files pass syntax checks, the 329 protocol assertions and 16 choice/skip/settlement scenarios pass, and the inert integration checks pass without manually executing deferred callbacks. The immutable JSON is unchanged. Native event-loop behavior remains for the platform evaluator to verify; no benchmark scorer was invoked.
\end{casecard}
\begin{casecard}[before upper={}]{Run Result}
{\sffamily\bfseries\casetitlefont Run Result}\hspace{1.5em}{\sffamily\bfseries Model:}~GPT-6 Astra\hspace{1.5em}{\sffamily\bfseries Overall status:}~\casepass{RESOLVED}\par
{\sffamily\bfseries Verdict:}~resolved; F2P 4/4 and P2P 57/57 in the running game (9 steps).\par
{\sffamily\bfseries Records (verbatim):}~\caselog{\{"score":\allowbreak{} 1.\allowbreak{}0,\allowbreak{} "complete":\allowbreak{} true,\allowbreak{} "passed":\allowbreak{} true,\allowbreak{} "checks":\allowbreak{} [\allowbreak{}\{"name":\allowbreak{} "check-\allowbreak{}1",\allowbreak{} "passed":\allowbreak{} true\},\allowbreak{} .\allowbreak{}.\allowbreak{}.\allowbreak{},\allowbreak{} \{"name":\allowbreak{} "check-\allowbreak{}68",\allowbreak{} "passed":\allowbreak{} true\}]\allowbreak{}\}}
\end{casecard}
\begin{casecard}[colback=white,runin]{Analysis}
The harness plays three runs of the real game through a remote-control mod: it buys and opens the Booster, chooses or skips Jokers through normal game actions, and checks the game state 68 times during the replay: the 4 F2P and 57 P2P tests, and 7 setup and sentinel checks. The frames come from a re-run of the frozen submission that reproduces the E1 verdict check by check. Choosing the first Joker raises the money from \$67 to \$70, the \$3 signing reward of \texttt{j\_pc2\_copper}, and closing the pack after the second choice leaves \$77.
\end{casecard}
\casecapgap\captionof{figure}{\textbf{Balatro, resolved.} A repair of how Booster choices commit their rewards; frames captured while the verifier plays the submission.}\label{fig:case-balatro-ok}
\caseback
\end{minipage}\par\vspace{\caseafterskip}
\par\noindent\begin{minipage}{\linewidth}
\casehead{\casepill{ospeach}{Error case}}{Balatro\,$\cdot$\,\texttt{pack-\allowbreak{}choice-\allowbreak{}commit-\allowbreak{}v2}\,$\cdot$\,Kimi K3.1 Pro}
\caserow{case-balatro-fail}{%
\casefield{task\_id}{\texttt{pack-\allowbreak{}choice-\allowbreak{}commit-\allowbreak{}v2}}
\casefield{Game}{Balatro}
\casefield{Type}{repair}
\casefield{Outcome}{Interface the ecosystem lacks}}{Task instruction (repair, excerpt)}{%
The ordinary \texttt{p\_\allowbreak{}pc2\_\allowbreak{}choices} Booster costs \$4 and presents three physical Common Jokers, in order: \texttt{j\_\allowbreak{}pc2\_\allowbreak{}copper}, \texttt{j\_\allowbreak{}pc2\_\allowbreak{}silver}, \texttt{j\_\allowbreak{}pc2\_\allowbreak{}gold}. ... The real adapter must route preview creation, acquisition, closure and delayed payout through this ledger.}
\caseshotlabel{In-game capture during the verifier run}
{\centering \begin{minipage}[t]{0.485\linewidth}\centering\caseshot{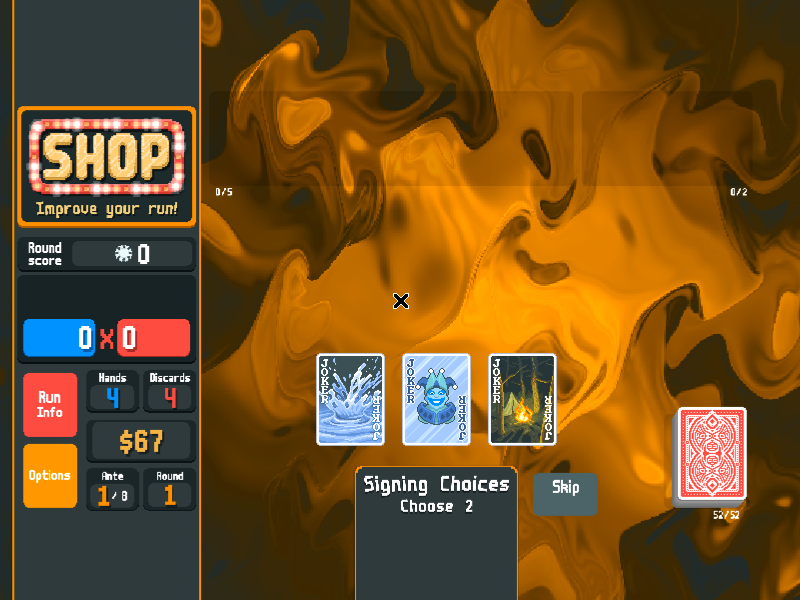}\\[1pt]{\scriptsize Kimi K3.1 Pro: three vanilla Jokers; the run stops.}\end{minipage}\hfill\begin{minipage}[t]{0.485\linewidth}\centering\caseshot{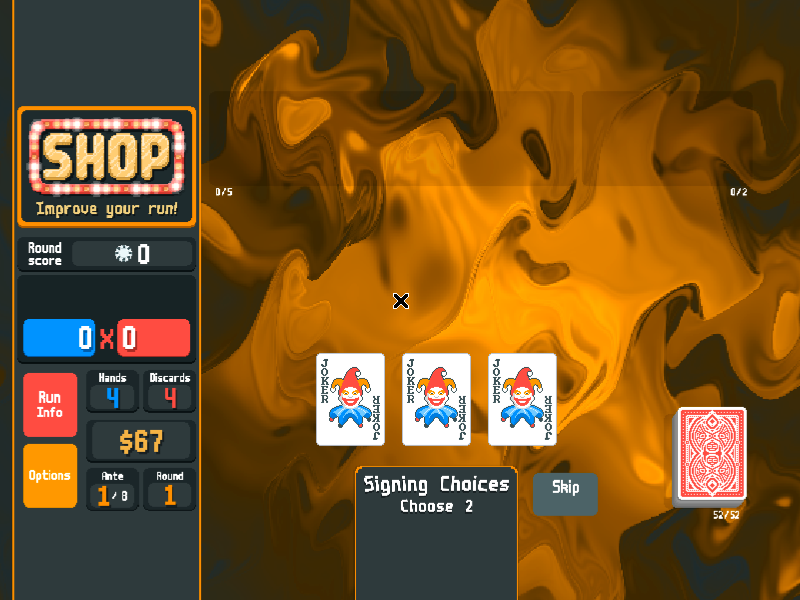}\\[1pt]{\scriptsize Reference solution at the same step.}\end{minipage}\par}
\begin{casecard}[colback=white,runin]{Agent's final report (verbatim)}
\itshape All 177 protocol assertions pass, the native lifecycle simulation passes, all Lua files compile, and the immutable JSON hash is unchanged. I'm doing a final workspace/source audit now; I have not invoked the benchmark scorer.
\end{casecard}
\begin{casecard}[before upper={}]{Run Result}
{\sffamily\bfseries\casetitlefont Run Result}\hspace{1.5em}{\sffamily\bfseries Model:}~Kimi K3.1 Pro\hspace{1.5em}{\sffamily\bfseries Overall status:}~\casefailchip{UNRESOLVED}\par
{\sffamily\bfseries Verdict:}~unresolved; the run stops at the first Booster and is incomplete, so all 4 F2P and 57 P2P tests are recorded as unobserved (13 steps).\par
{\sffamily\bfseries Records (verbatim):}~\caselog{submission starter/\allowbreak{}main.\allowbreak{}lua:\allowbreak{}42:\allowbreak{}   local card =\allowbreak{} create\_\allowbreak{}card(\allowbreak{}'Joker',\allowbreak{} G.\allowbreak{}pack\_\allowbreak{}cards,\allowbreak{} nil,\allowbreak{} nil,\allowbreak{} nil,\allowbreak{} nil,\allowbreak{} nil,\allowbreak{} key,\allowbreak{} 'pc2')\allowbreak{}\\{}\{"score":\allowbreak{} 0.\allowbreak{}0,\allowbreak{} "complete":\allowbreak{} false,\allowbreak{} "passed":\allowbreak{} false,\allowbreak{} .\allowbreak{}.\allowbreak{}.\allowbreak{},\allowbreak{} "category":\allowbreak{} "incomplete",\allowbreak{} "reason":\allowbreak{} "environment\_\allowbreak{}failure",\allowbreak{} .\allowbreak{}.\allowbreak{}.\allowbreak{}\}\\{}adjudication:\allowbreak{} engine classification `test\_\allowbreak{}gap`,\allowbreak{} failure phase `test`,\allowbreak{} error `ProtocolError:\allowbreak{} Required signing preview unavailable`.\allowbreak{} .\allowbreak{}.\allowbreak{}.\allowbreak{} the run stopped at `full\_\allowbreak{}full.\allowbreak{}p1.\allowbreak{}open`,\allowbreak{} 53 checks not executed.\allowbreak{}}
\end{casecard}
\begin{casecard}[colback=white,runin]{Analysis}
Balatro's \texttt{create\_card} takes the key of a forced card as its seventh argument; the submission passes it as the eighth, after an extra \texttt{nil}, where the function expects a seed suffix. The Booster therefore deals random vanilla Jokers (Splash, Blue Joker, and Campfire in this run) instead of the three signing Jokers, and the scenario cannot continue. The agent's own protocol assertions and lifecycle simulation exercise the ledger, not the game's card factory, so all 177 pass. The re-run shown here reproduces the E1 outcome: of the 68 runtime checks in the raw log (the 61 tests and 7 setup and sentinel checks), 15 pass before the run stops and 53 never run, and because the run is incomplete, the verdict records every test as unobserved. The verifier first classified the stop as an environment failure; manual adjudication attributed it to the submission.
\end{casecard}
\casecapgap\captionof{figure}{\textbf{Balatro, unresolved (interface the ecosystem lacks).} The same task as Figure~\ref{fig:case-balatro-ok}: a misplaced argument to the game's card factory fills the pack with the wrong Jokers.}\label{fig:case-balatro-fail}
\caseback
\end{minipage}\par\vspace{\caseafterskip}
Figure~\ref{fig:case-balatro-crash} shows a third Balatro outcome from the same re-runs: Qwen 3.8 27B's submission for the repair \texttt{growth-v2} subscribes to \texttt{G.event\_bus}, which the game does not define, and the game stops at its crash screen while loading the mod, so none of the task's 18 cases runs. The crash text names the submitted file and line.
\begin{center}
\includegraphics[width=0.62\linewidth]{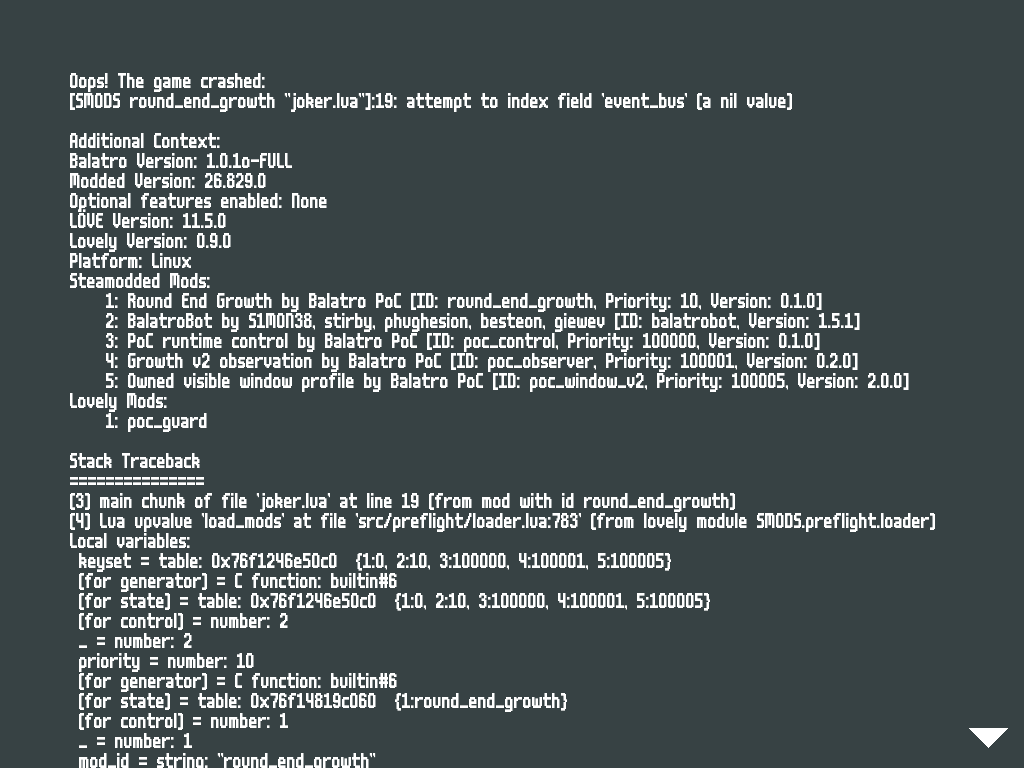}
\casecapgap\captionof{figure}{\textbf{Balatro, unresolved (interface the ecosystem lacks): crash at mod load.} Frame of the game's crash screen while the verifier loads Qwen 3.8 27B's submission for \texttt{growth-v2}; the E1 record shows the same crash.}\label{fig:case-balatro-crash}
\end{center}
\clearpage
\subsection{The Binding of Isaac}
\label{app:case-isaac}
\par\noindent\begin{minipage}{\linewidth}
\casehead{\casepill{osblue}{Resolved}}{The Binding of Isaac\,$\cdot$\,\texttt{isaac-\allowbreak{}exodus-\allowbreak{}release-\allowbreak{}regressions}\,$\cdot$\,Claude Opus 5.5}
\caserow{case-isaac-ok}{%
\casefield{task\_id}{\texttt{isaac-\allowbreak{}exodus-\allowbreak{}release-\allowbreak{}regressions}}
\casefield{Game}{The Binding of Isaac}
\casefield{Type}{repair}
\casefield{Outcome}{Resolved}}{Task instruction (repair, excerpt)}{%
[\#EX-342] Headcase's arcing blood shots become harmless if Headcase dies just after the stomp launches them. Shots already in flight can overlap the player on landing without dealing damage. Their collision must continue to be processed after the firing Headcase is gone. ... Diagnose and fix every report. Preserve behaviour outside the reported paths.}
\caseshotlabel{In-game capture during the verifier run}
{\centering \begin{minipage}[t]{0.485\linewidth}\centering\caseshot{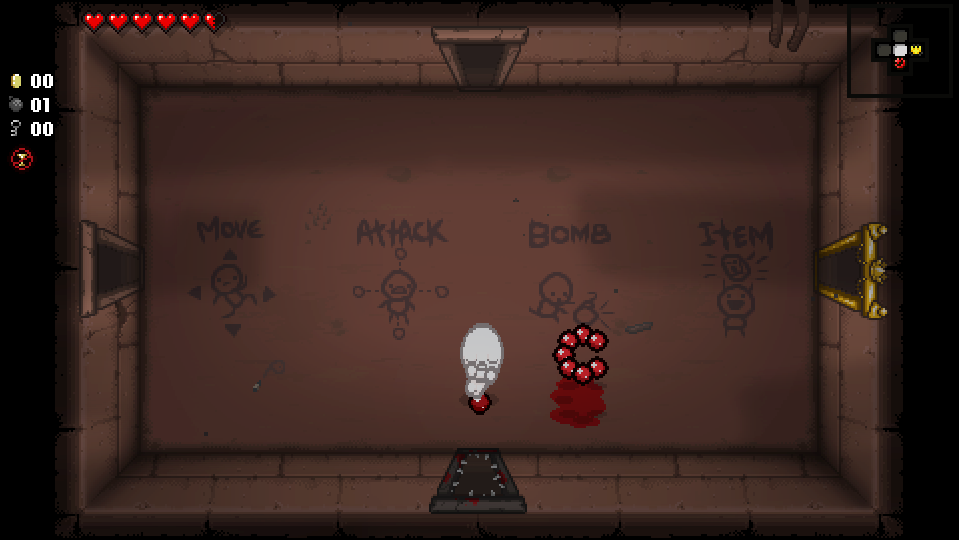}\\[1pt]{\scriptsize The Headcase has died; its shots reach Isaac, who flashes and drops to five and a half hearts.}\end{minipage}\hfill\begin{minipage}[t]{0.485\linewidth}\centering\caseshot{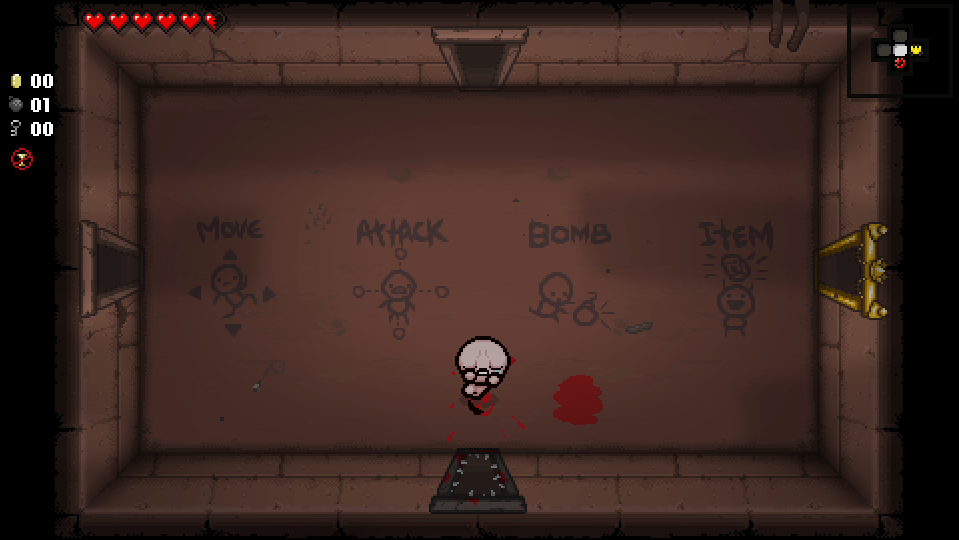}\\[1pt]{\scriptsize 0.3\,s later: Isaac in the hurt pose, five and a half hearts.}\end{minipage}\par}
\begin{casecard}[colback=white,runin]{Agent's final report (verbatim)}
\itshape ... shots are checked every frame, whether or not a Headcase is alive. Each shot can hurt the player at most once.
\end{casecard}
\begin{casecard}[before upper={}]{Run Result}
{\sffamily\bfseries\casetitlefont Run Result}\hspace{1.5em}{\sffamily\bfseries Model:}~Claude Opus 5.5\hspace{1.5em}{\sffamily\bfseries Overall status:}~\casepass{RESOLVED}\par
{\sffamily\bfseries Verdict:}~resolved; F2P 26/26 and P2P 6/6 in the running game, across three game processes (47 steps).\par
{\sffamily\bfseries Records (verbatim):}~\caselog{[\allowbreak{}INFO]\allowbreak{} -\allowbreak{} Lua Debug:\allowbreak{} [\allowbreak{}GCB\_\allowbreak{}CASE]\allowbreak{} PASS headcase\_\allowbreak{}orphan\_\allowbreak{}lob before\_\allowbreak{}red=\allowbreak{}12 after\_\allowbreak{}red=\allowbreak{}11 lob\_\allowbreak{}exists=\allowbreak{}false\\{}[\allowbreak{}INFO]\allowbreak{} -\allowbreak{} Lua Debug:\allowbreak{} [\allowbreak{}GCB\_\allowbreak{}SUMMARY]\allowbreak{} passed=\allowbreak{}30 failed=\allowbreak{}0 total=\allowbreak{}30\\{}runtime\_\allowbreak{}exit=\allowbreak{}0 secret\_\allowbreak{}exit=\allowbreak{}0 james\_\allowbreak{}exit=\allowbreak{}0 main\_\allowbreak{}exit=\allowbreak{}0 passed=\allowbreak{}35 failed=\allowbreak{}0 expected=\allowbreak{}35 summaries=\allowbreak{}3 fatals=\allowbreak{}0}
\end{casecard}
\begin{casecard}[colback=white,runin]{Analysis}
The Exodus repair bundles several bug reports; the suite reproduces each one in the native game. In the Headcase case, the boss launches its arcing shots and is killed before they land; the player then loses one half-heart (\texttt{before\_red=12 after\_red=11}), so the orphaned shot still collides. The frames show this hit in a re-run of the frozen submission; compare the same moment in Figure~\ref{fig:case-isaac-fail}.
\end{casecard}
\casecapgap\captionof{figure}{\textbf{The Binding of Isaac, resolved.} A repair of several Exodus bug reports, including shots that outlive the enemy that fired them, passes every in-game case.}\label{fig:case-isaac-ok}
\caseback
\end{minipage}\par\vspace{\caseafterskip}
\par\noindent\begin{minipage}{\linewidth}
\casehead{\casepill{ospeach}{Error case}}{The Binding of Isaac\,$\cdot$\,\texttt{isaac-\allowbreak{}exodus-\allowbreak{}release-\allowbreak{}regressions}\,$\cdot$\,GPT-6 Astra}
\caserow{case-isaac-fail}{%
\casefield{task\_id}{\texttt{isaac-\allowbreak{}exodus-\allowbreak{}release-\allowbreak{}regressions}}
\casefield{Game}{The Binding of Isaac}
\casefield{Type}{repair}
\casefield{Outcome}{Behavior or lifecycle in game}}{Task instruction (repair, excerpt)}{%
[\#EX-342] Headcase's arcing blood shots become harmless if Headcase dies just after the stomp launches them. Shots already in flight can overlap the player on landing without dealing damage. Their collision must continue to be processed after the firing Headcase is gone.}
\caseshotlabel{In-game capture during the verifier run}
{\centering \begin{minipage}[t]{0.485\linewidth}\centering\caseshot{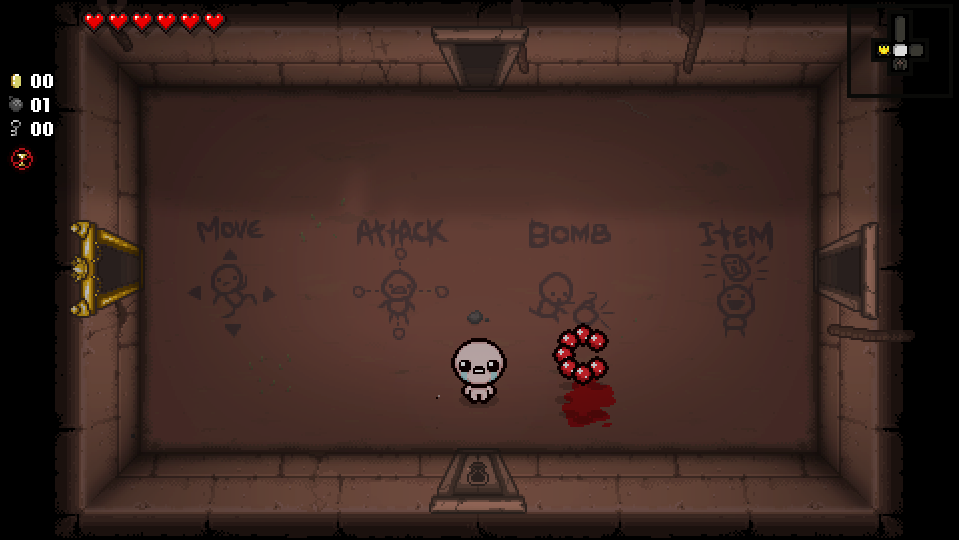}\\[1pt]{\scriptsize A Headcase has just died (blood stain); its eight blood shots are in the air, and Isaac has six hearts.}\end{minipage}\hfill\begin{minipage}[t]{0.485\linewidth}\centering\caseshot{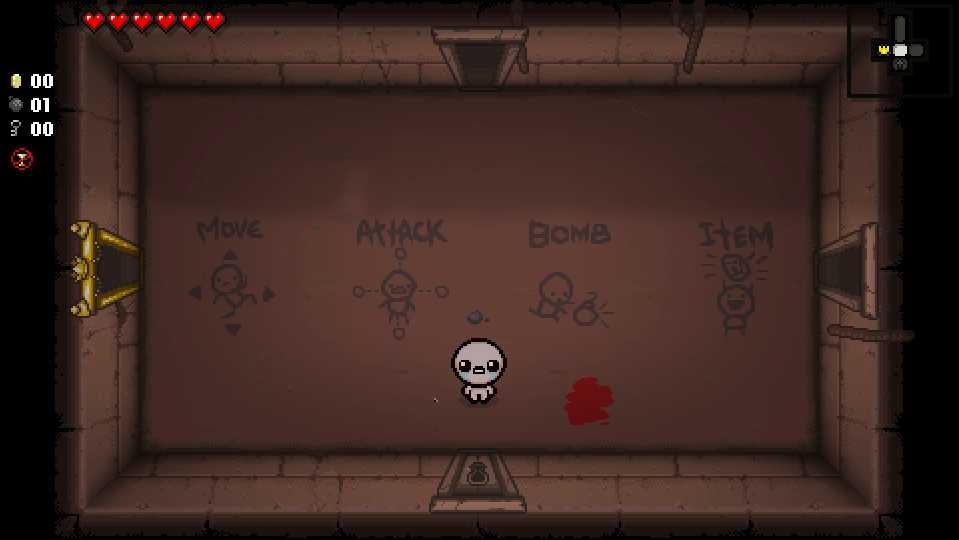}\\[1pt]{\scriptsize 0.2\,s later the shots have landed; Isaac still has six hearts.}\end{minipage}\par}
\begin{casecard}[colback=white,runin]{Agent's final report (verbatim)}
\itshape The fixes and local checks are complete: all 20 mocked regression cases pass, and callback registration and XML checks pass. Native-game verification remains for the supplied evaluator.
\end{casecard}
\begin{casecard}[before upper={}]{Run Result}
{\sffamily\bfseries\casetitlefont Run Result}\hspace{1.5em}{\sffamily\bfseries Model:}~GPT-6 Astra\hspace{1.5em}{\sffamily\bfseries Overall status:}~\casefailchip{UNRESOLVED}\par
{\sffamily\bfseries Verdict:}~unresolved; F2P 25/26 and P2P 6/6 (20 steps).\par
{\sffamily\bfseries Records (verbatim):}~\caselog{[\allowbreak{}INFO]\allowbreak{} -\allowbreak{} Lua Debug:\allowbreak{} [\allowbreak{}GCB\_\allowbreak{}CASE]\allowbreak{} FAIL headcase\_\allowbreak{}orphan\_\allowbreak{}lob before\_\allowbreak{}red=\allowbreak{}12 after\_\allowbreak{}red=\allowbreak{}12 lob\_\allowbreak{}exists=\allowbreak{}false\\{}[\allowbreak{}INFO]\allowbreak{} -\allowbreak{} Lua Debug:\allowbreak{} [\allowbreak{}GCB\_\allowbreak{}SUMMARY]\allowbreak{} passed=\allowbreak{}29 failed=\allowbreak{}1 total=\allowbreak{}30\\{}runtime\_\allowbreak{}exit=\allowbreak{}0 secret\_\allowbreak{}exit=\allowbreak{}0 james\_\allowbreak{}exit=\allowbreak{}0 main\_\allowbreak{}exit=\allowbreak{}0 passed=\allowbreak{}34 failed=\allowbreak{}1 expected=\allowbreak{}35 summaries=\allowbreak{}2 fatals=\allowbreak{}0}
\end{casecard}
\begin{casecard}[colback=white,runin]{Analysis}
The submission fixes every other report and passes all P2P cases, but in the Headcase case the player's health is unchanged after the orphaned shot lands (\texttt{before\_red=12 after\_red=12}): the shot remains harmless once its source has died, which is precisely the reported bug. The frames show this moment in a re-run that reproduces the E1 verdict: the eight shots appear where the Headcase died and vanish within 0.2\,s, while Isaac keeps all six hearts. The agent's mocked regression suite contains a test for exactly this report (``EX-342: Every lobbed shot remains dangerous after Headcase dies''), which calls the shot's update handler directly on a fake entity. In the submission that handler is registered on \texttt{MC\_POST\_TEAR\_UPDATE}, whereas Claude Opus 5.5's resolved submission checks every lobbed shot once per frame from \texttt{MC\_POST\_UPDATE}. Whether the game dispatches the handler at the moment the shot lands is exactly what the mock cannot show.
\end{casecard}
\casecapgap\captionof{figure}{\textbf{The Binding of Isaac, unresolved (behavior in the game).} The same task as Figure~\ref{fig:case-isaac-ok}: shots launched by a Headcase that then dies still do no damage.}\label{fig:case-isaac-fail}
\caseback
\end{minipage}\par\vspace{\caseafterskip}
\clearpage
\subsection{Stardew Valley}
\label{app:case-stardew}
\par\noindent\begin{minipage}{\linewidth}
\casehead{\casepill{osblue}{Resolved}}{Stardew Valley\,$\cdot$\,\texttt{stardew-\allowbreak{}summon-\allowbreak{}horse-\allowbreak{}anywhere}\,$\cdot$\,Claude Opus 5.5}
\caserow{case-stardew-ok}{%
\casefield{task\_id}{\texttt{stardew-\allowbreak{}summon-\allowbreak{}horse-\allowbreak{}anywhere}}
\casefield{Game}{Stardew Valley}
\casefield{Type}{construction}
\casefield{Outcome}{Resolved}}{Task instruction (construction, excerpt)}{%
In the base game, the horse flute summons your horse to you, but the game refuses to do it when you are indoors ... Your mod must make using the horse flute summon the player's horse to the player's current location even when that location is indoors, bringing the horse next to the player the same way it does outdoors.}
\caseshotlabel{In-game capture during the verifier run}
{\centering \begin{minipage}[t]{0.62\linewidth}\centering\caseshot{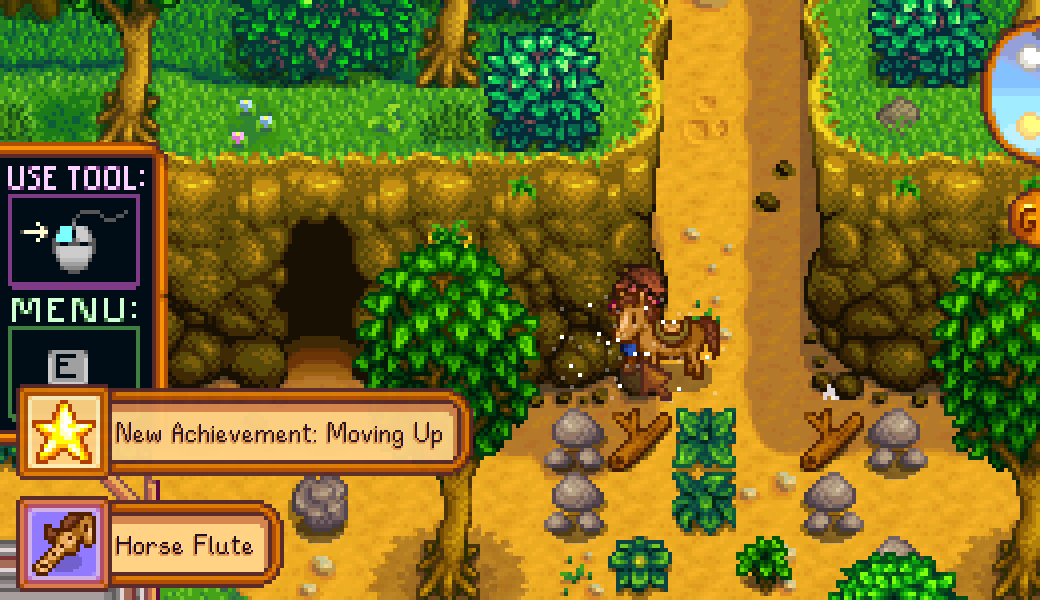}\\[1pt]{\scriptsize Outdoor case: after the flute is used, the horse arrives next to the farmer.}\end{minipage}\par}
\begin{casecard}[colback=white,runin]{Agent's final report (verbatim)}
\itshape Neither the game nor \texttt{dotnet} is installed here, so I can't compile. I'll write code that only uses compile-time types I'm sure exist, and find the game method at runtime with reflection.
\end{casecard}
\begin{casecard}[before upper={}]{Run Result}
{\sffamily\bfseries\casetitlefont Run Result}\hspace{1.5em}{\sffamily\bfseries Model:}~Claude Opus 5.5\hspace{1.5em}{\sffamily\bfseries Overall status:}~\casepass{RESOLVED}\par
{\sffamily\bfseries Verdict:}~resolved; F2P 1/1 and P2P 1/1 in the running game (5 steps).\par
{\sffamily\bfseries Records (verbatim):}~\caselog{headless-\allowbreak{}game:\allowbreak{} all real-\allowbreak{}game assertions passed\\{}passed=\allowbreak{}3 failed=\allowbreak{}0 fatal=\allowbreak{}None expected=\allowbreak{}3}
\end{casecard}
\begin{casecard}[colback=white,runin]{Analysis}
Without a compiler or the game, the agent chose to locate the game method by reflection at run time instead of referencing it at compile time. The harness compiles the mod, loads it into the game with a save that has a horse, and uses the flute outdoors and inside the farmhouse; the horse arrives in both places.
\end{casecard}
\casecapgap\captionof{figure}{\textbf{Stardew Valley, resolved.} A new SMAPI mod lets the horse flute work indoors.}\label{fig:case-stardew-ok}
\caseback
\end{minipage}\par\vspace{\caseafterskip}
\par\noindent\begin{minipage}{\linewidth}
\casehead{\casepill{ospeach}{Error case}}{Stardew Valley\,$\cdot$\,\texttt{stardew-\allowbreak{}summon-\allowbreak{}horse-\allowbreak{}anywhere}\,$\cdot$\,Kimi K3.1 Pro}
\caserow{case-stardew-fail}{%
\casefield{task\_id}{\texttt{stardew-\allowbreak{}summon-\allowbreak{}horse-\allowbreak{}anywhere}}
\casefield{Game}{Stardew Valley}
\casefield{Type}{construction}
\casefield{Outcome}{Behavior or lifecycle in game}}{Task instruction (construction, excerpt)}{%
The grading harness compiles your mod, loads it into the real game headless with a save that has a horse, and uses the flute both outdoors and indoors, checking where the horse ends up.}
\begin{casecard}[colback=white,runin]{Agent's final report (verbatim)}
\itshape The game installation and .NET SDK are not present in this workspace, so I'm implementing a narrow Harmony IL patch: it changes \texttt{IsOutdoors} only while the Horse Flute object itself is being used, leaving every other item and every other summoning condition on the original code path.
\end{casecard}
\begin{casecard}[before upper={}]{Run Result}
{\sffamily\bfseries\casetitlefont Run Result}\hspace{1.5em}{\sffamily\bfseries Model:}~Kimi K3.1 Pro\hspace{1.5em}{\sffamily\bfseries Overall status:}~\casefailchip{UNRESOLVED}\par
{\sffamily\bfseries Verdict:}~unresolved; F2P 0/1, P2P 1/1 (7 steps).\par
{\sffamily\bfseries Records (verbatim):}~\caselog{passed=\allowbreak{}2 failed=\allowbreak{}1 fatal=\allowbreak{}None expected=\allowbreak{}3\\{}  [\allowbreak{}FAIL]\allowbreak{} the flute summons the horse indoors -\allowbreak{}-\allowbreak{} the horse was summoned into the farmhouse:\allowbreak{} expected non-\allowbreak{}null}
\end{casecard}
\begin{casecard}[colback=white,runin]{Analysis}
The mod builds and loads, and the outdoor case passes, but after the flute is used inside the farmhouse the harness finds no horse there. The agent, which had neither the game nor the .NET SDK, could not observe this. GPT-6 Astra's submission for the same task does not compile at all, because it refers to a type \texttt{HorseFlute} that the game does not define (\texttt{error CS0246}).
\end{casecard}
\casecapgap\captionof{figure}{\textbf{Stardew Valley, unresolved (behavior in the game).} The same task as Figure~\ref{fig:case-stardew-ok}: the flute is used indoors, but the horse does not arrive.}\label{fig:case-stardew-fail}
\caseback
\end{minipage}\par\vspace{\caseafterskip}
\clearpage
\subsection{Minecraft}
\label{app:case-minecraft}
\par\noindent\begin{minipage}{\linewidth}
\casehead{\casepill{osblue}{Resolved}}{Minecraft\,$\cdot$\,\texttt{blocktrail-\allowbreak{}block-\allowbreak{}log-\allowbreak{}rollback}\,$\cdot$\,GPT-6 Astra}
\caserow{case-minecraft-ok}{%
\casefield{task\_id}{\texttt{blocktrail-\allowbreak{}block-\allowbreak{}log-\allowbreak{}rollback}}
\casefield{Game}{Minecraft}
\casefield{Type}{construction}
\casefield{Outcome}{Resolved}}{Task instruction (construction, excerpt)}{%
Write a plugin: BlockTrail, a block change log with rollback ... Material alone is not enough: rollback must preserve properties such as stair facing, shape, half and waterlogging. ... A qualifying block is restored to its recorded \texttt{from} material and block data, then marked rolled back (append the \texttt{undo} line); otherwise it is skipped and stays unmarked. ... The reply is \texttt{Rollback of \textless{}actor\textgreater{}:\allowbreak{} \textless{}restored\textgreater{} restored,\allowbreak{} \textless{}skipped\textgreater{} skipped.\allowbreak{}}, or \texttt{Nothing to roll back for \textless{}actor\textgreater{}.\allowbreak{}} when no change qualified.}
\begin{casecard}[colback=white,runin]{Agent's final report (verbatim)}
\itshape The implementation is complete. Static checks passed for the default configuration, command and permission declarations, source formatting and Maven XML. The offline build remains unverified because Java and Maven are unavailable in this workspace.
\end{casecard}
\begin{casecard}[before upper={}]{Run Result}
{\sffamily\bfseries\casetitlefont Run Result}\hspace{1.5em}{\sffamily\bfseries Model:}~GPT-6 Astra\hspace{1.5em}{\sffamily\bfseries Overall status:}~\casepass{RESOLVED}\par
{\sffamily\bfseries Verdict:}~resolved; F2P 26/26 on the live Paper server (11 steps).\par
{\sffamily\bfseries Records (verbatim):}~\caselog{offline-\allowbreak{}build:\allowbreak{} maven build passed\\{}live-\allowbreak{}server:\allowbreak{} 27/\allowbreak{}27 real-\allowbreak{}server cases passed\\{}[\allowbreak{}14:\allowbreak{}49:\allowbreak{}41 INFO]\allowbreak{}:\allowbreak{} [\allowbreak{}GcbHarness]\allowbreak{} [\allowbreak{}STDOUT]\allowbreak{} MINECRAFT\_\allowbreak{}RESULT \{"harness":\allowbreak{}"gcb-\allowbreak{}harness",\allowbreak{}.\allowbreak{}.\allowbreak{}.\allowbreak{},\allowbreak{}"candidate":\allowbreak{}"BlockTrail",\allowbreak{}"passed":\allowbreak{}27,\allowbreak{}"failed":\allowbreak{}0,\allowbreak{}"fatal":\allowbreak{}null,\allowbreak{}.\allowbreak{}.\allowbreak{}.\allowbreak{}\}}
\end{casecard}
\begin{casecard}[colback=white,runin]{Analysis}
The agent never compiled its plugin; its report says so. On the Paper server, the harness places, breaks, and explodes blocks, rolls back changes by actor, area, and time, and checks the exact block data that each rollback restores. The submission passes all of these cases, including the stale-state case that Claude Opus 5.5 fails on the same task (Figure~\ref{fig:case-minecraft-fail}).
\end{casecard}
\casecapgap\captionof{figure}{\textbf{Minecraft, resolved.} A BlockTrail plugin written from a skeleton passes all live-server cases, although the agent could not build it.}\label{fig:case-minecraft-ok}
\caseback
\end{minipage}\par\vspace{\caseafterskip}
\par\noindent\begin{minipage}{\linewidth}
\casehead{\casepill{ospeach}{Error case}}{Minecraft\,$\cdot$\,\texttt{blocktrail-\allowbreak{}block-\allowbreak{}log-\allowbreak{}rollback}\,$\cdot$\,Claude Opus 5.5}
\caserow{case-minecraft-fail}{%
\casefield{task\_id}{\texttt{blocktrail-\allowbreak{}block-\allowbreak{}log-\allowbreak{}rollback}}
\casefield{Game}{Minecraft}
\casefield{Type}{construction}
\casefield{Outcome}{Behavior or lifecycle in game}}{Task instruction (construction, excerpt)}{%
It processes them newest first: a new-format change qualifies only when the current material and full block data equal its recorded \texttt{to} state; a legacy line checks only its material. A qualifying block is restored to its recorded \texttt{from} material and block data, then marked rolled back (append the \texttt{undo} line); otherwise it is skipped and stays unmarked. This preserves stair facing, shape, half and waterlogging, and protects later data-only changes from being overwritten.}
\begin{casecard}[colback=white,runin]{Agent's final report}
None: the agent submitted after six steps, and its last messages contain only commands.
\end{casecard}
\begin{casecard}[before upper={}]{Run Result}
{\sffamily\bfseries\casetitlefont Run Result}\hspace{1.5em}{\sffamily\bfseries Model:}~Claude Opus 5.5\hspace{1.5em}{\sffamily\bfseries Overall status:}~\casefailchip{UNRESOLVED}\par
{\sffamily\bfseries Verdict:}~unresolved; F2P 25/26 on the live Paper server.\par
{\sffamily\bfseries Records (verbatim):}~\caselog{live server failed:\allowbreak{} 26/\allowbreak{}27\\{}\{"name":\allowbreak{} "rollback restores directional block data and skips stale states",\allowbreak{} "passed":\allowbreak{} false,\allowbreak{} "message":\allowbreak{} "/\allowbreak{}trail rollback Wren world 1262 -\allowbreak{}59 0 1 60 (\allowbreak{}expected \textless{}[\allowbreak{}Rollback of Wren:\allowbreak{} 0 restored,\allowbreak{} 1 skipped.\allowbreak{}]\allowbreak{}\textgreater{} but was \textless{}[\allowbreak{}Nothing to roll back for Wren.\allowbreak{}]\allowbreak{}\textgreater{})\allowbreak{}"\}\\{}SUBMISSION\_\allowbreak{}FAILED:\allowbreak{} one or more cases failed}
\end{casecard}
\begin{casecard}[colback=white,runin]{Analysis}
The plugin builds and passes 25 of the 26 F2P cases. The failing case changes a logged block once more, so that its current state no longer equals the recorded \texttt{to} state, and then requests a rollback. The specification counts such a change as skipped (\texttt{0 restored, 1 skipped}); the submission instead reports that nothing qualified. The difference lies in how the plugin orders its filters on state held by the running server, which only a live rollback exercises.
\end{casecard}
\casecapgap\captionof{figure}{\textbf{Minecraft, unresolved (behavior in the game).} The same task as Figure~\ref{fig:case-minecraft-ok}: one stale-state rollback is reported incorrectly.}\label{fig:case-minecraft-fail}
\caseback
\end{minipage}\par\vspace{\caseafterskip}
\clearpage
\subsection{Garry's Mod}
\label{app:case-gmod}
\par\noindent\begin{minipage}{\linewidth}
\casehead{\casepill{osblue}{Resolved}}{Garry's Mod\,$\cdot$\,\texttt{darkrp-\allowbreak{}money-\allowbreak{}regressions}\,$\cdot$\,Kimi K3.1 Pro}
\caserow{case-gmod-ok}{%
\casefield{task\_id}{\texttt{darkrp-\allowbreak{}money-\allowbreak{}regressions}}
\casefield{Game}{Garry's Mod}
\casefield{Type}{repair}
\casefield{Outcome}{Resolved}}{Task instruction (repair, excerpt)}{%
[\#6101] People can buy things they cannot quite pay for. When the price is not a whole number and it lands just above what someone is carrying, the purchase goes through anyway and they are left holding less than nothing. ... [\#6110] Cash that a player put down is no longer treated as cash by anything that asks. Piles the server itself creates are still recognised, so it is specifically the ones that came out of somebody's pocket that have stopped counting.}
\begin{casecard}[colback=white,runin]{Agent's final report (verbatim)}
\itshape All seven regression checks pass, as do the ordinary-value checks for existing behavior. I'm doing the final source diff audit now; no test files were added to the gamemode.
\end{casecard}
\begin{casecard}[before upper={}]{Run Result}
{\sffamily\bfseries\casetitlefont Run Result}\hspace{1.5em}{\sffamily\bfseries Model:}~Kimi K3.1 Pro\hspace{1.5em}{\sffamily\bfseries Overall status:}~\casepass{RESOLVED}\par
{\sffamily\bfseries Verdict:}~resolved; F2P 7/7 and P2P 1/1 on the dedicated server (14 steps).\par
{\sffamily\bfseries Records (verbatim):}~\caselog{headless-\allowbreak{}server:\allowbreak{} all headless GLuaTest assertions passed\\{}entrypoint\_\allowbreak{}exit=\allowbreak{}0 passed=\allowbreak{}9 failed=\allowbreak{}0 expected=\allowbreak{}9 reward=\allowbreak{}1}
\end{casecard}
\begin{casecard}[colback=white,runin]{Analysis}
The reports describe symptoms, not causes, and span three files of the DarkRP gamemode. The submission changes exactly those files, and the GLuaTest fixtures exercise purchases, wallet-change hooks, and dropped money inside the dedicated server with the full gamemode loaded. The agent's own checks covered the same seven reports, and here they agreed with the runtime.
\end{casecard}
\casecapgap\captionof{figure}{\textbf{Garry's Mod, resolved.} Seven money-related regressions in DarkRP are fixed and verified on the dedicated server.}\label{fig:case-gmod-ok}
\caseback
\end{minipage}\par\vspace{\caseafterskip}
\par\noindent\begin{minipage}{\linewidth}
\casehead{\casepill{ospeach}{Error case}}{Garry's Mod\,$\cdot$\,\texttt{persistent-\allowbreak{}loadout-\allowbreak{}profiles}\,$\cdot$\,GPT-6 Astra}
\caserow{case-gmod-fail}{%
\casefield{task\_id}{\texttt{persistent-\allowbreak{}loadout-\allowbreak{}profiles}}
\casefield{Game}{Garry's Mod}
\casefield{Type}{construction}
\casefield{Outcome}{Host convention}}{Task instruction (construction, excerpt)}{%
T3-GOAL-01 Implement a server-authoritative system of named Loadout profiles: save weapon and ammo plans, select a preferred profile, and restore after death, respawn, disconnect-and-rebind, and separate server process restart. ... T3-FILE-01 You must submit: addon.json lua/rhenium\_loadout\_profiles/profiles.lua lua/rhenium\_loadout\_profiles/server.lua}
\begin{casecard}[colback=white,runin]{Agent's final report (verbatim)}
\itshape All local checks pass: 534 assertions, 24,000 replays, and two-process persistence. I'm submitting the implementation now.
\end{casecard}
\begin{casecard}[before upper={}]{Run Result}
{\sffamily\bfseries\casetitlefont Run Result}\hspace{1.5em}{\sffamily\bfseries Model:}~GPT-6 Astra\hspace{1.5em}{\sffamily\bfseries Overall status:}~\casefailchip{UNRESOLVED}\par
{\sffamily\bfseries Verdict:}~unresolved; rejected by the addon gate, so the dedicated-server fixtures never ran.\par
{\sffamily\bfseries Records (verbatim):}~\caselog{changed files:\allowbreak{} README.\allowbreak{}md,\allowbreak{} addon.\allowbreak{}json,\allowbreak{} lua/\allowbreak{}rhenium\_\allowbreak{}loadout\_\allowbreak{}profiles/\allowbreak{}profiles.\allowbreak{}lua,\allowbreak{} lua/\allowbreak{}rhenium\_\allowbreak{}loadout\_\allowbreak{}profiles/\allowbreak{}server.\allowbreak{}lua,\allowbreak{} tests/\allowbreak{}test\_\allowbreak{}profiles.\allowbreak{}lua,\allowbreak{} tests/\allowbreak{}test\_\allowbreak{}restart.\allowbreak{}lua,\allowbreak{} tests/\allowbreak{}test\_\allowbreak{}roundtrip.\allowbreak{}lua,\allowbreak{} tests/\allowbreak{}test\_\allowbreak{}server.\allowbreak{}lua\\{}lua-\allowbreak{}syntax:\allowbreak{} failed  \{"errors":\allowbreak{} [\allowbreak{}\{"message":\allowbreak{} "unsupported submission file:\allowbreak{} README.\allowbreak{}md"\}]\allowbreak{}\}\\{}\ldots\\{}dedicated-\allowbreak{}server:\allowbreak{} 0/\allowbreak{}0 real Dedicated Server groups passed}
\end{casecard}
\begin{casecard}[colback=white,runin]{Analysis}
This is the Loadout task of Appendix~\ref{app:loadout}. Besides its Lua code, the addon contains a \texttt{README.md} and the agent's \texttt{tests/} directory, and the addon gate admits only \texttt{addon.json}, Lua files under \texttt{lua/}, and the task's own instruction files. Garry's Mod's own addon packer, \texttt{gmad}, rejects the same files: its whitelist admits Lua files under \texttt{lua/} but neither a \texttt{README.md} nor files in a top-level \texttt{tests/} directory, so the addon could not be published to the Workshop either. Because of the rejection, the fixtures that apply profiles to bots on two dedicated-server processes never run. The agent's 534 assertions and 24{,}000 replays ran against its own model of the game, which has no packaging rule.
\end{casecard}
\casecapgap\captionof{figure}{\textbf{Garry's Mod, unresolved (host convention).} The addon is rejected because it does not follow the host's packaging layout.}\label{fig:case-gmod-fail}
\caseback
\end{minipage}\par\vspace{\caseafterskip}
\clearpage
\subsection{Don't Starve Together}
\label{app:case-dst}
\par\noindent\begin{minipage}{\linewidth}
\casehead{\casepill{osblue}{Resolved}}{Don't Starve Together\,$\cdot$\,\texttt{container-\allowbreak{}transfer-\allowbreak{}transaction}\,$\cdot$\,GPT-6 Astra}
\caserow{case-dst-ok}{%
\casefield{task\_id}{\texttt{container-\allowbreak{}transfer-\allowbreak{}transaction}}
\casefield{Game}{Don't Starve Together}
\casefield{Type}{construction}
\casefield{Outcome}{Resolved}}{Task instruction (construction, excerpt)}{%
Implement a plain-Lua transfer transaction with an injected adapter and a server world component that operates on base-game container components. Persist intent before removal, use deterministic source slots, preflight destination capacity, ... Create actions with AddAction, AddComponentAction, AddStategraphActionHandler for wilson and wilson\_client, and AddModRPCHandler.}
\begin{casecard}[colback=white,runin]{Agent's final report (verbatim)}
\itshape All required files have been reviewed. Both local test suites pass, and the final public diagnostic reports no Lua syntax errors. The implementation is ready for submission; dedicated-server validation remains with the benchmark platform.
\end{casecard}
\begin{casecard}[before upper={}]{Run Result}
{\sffamily\bfseries\casetitlefont Run Result}\hspace{1.5em}{\sffamily\bfseries Model:}~GPT-6 Astra\hspace{1.5em}{\sffamily\bfseries Overall status:}~\casepass{RESOLVED}\par
{\sffamily\bfseries Verdict:}~resolved; F2P 13/13 (11 steps).\par
{\sffamily\bfseries Records (verbatim):}~\caselog{\ldots\\{}[\allowbreak{}00:\allowbreak{}00:\allowbreak{}00]\allowbreak{}:\allowbreak{} Mod:\allowbreak{} rhenium\_\allowbreak{}a8bf.\allowbreak{}.\allowbreak{}.\allowbreak{} (\allowbreak{}Crash-\allowbreak{}safe Container Transfer)\allowbreak{}	Loading modmain.\allowbreak{}lua\\{}[\allowbreak{}00:\allowbreak{}00:\allowbreak{}02]\allowbreak{}:\allowbreak{} Mod:\allowbreak{} rhenium\_\allowbreak{}a8bf.\allowbreak{}.\allowbreak{}.\allowbreak{} (\allowbreak{}Crash-\allowbreak{}safe Container Transfer)\allowbreak{}	  Registering default mod prefab\\{}headless-\allowbreak{}server:\allowbreak{} All 3 headless server assertions passed}
\end{casecard}
\begin{casecard}[colback=white,runin]{Analysis}
The dedicated server loads \texttt{modmain.lua} in the restricted mod environment, registers the mod's prefab, and runs the in-game transfer assertions against real container components. Compare Figure~\ref{fig:case-dst-fail}, whose mod never reaches this point.
\end{casecard}
\casecapgap\captionof{figure}{\textbf{Don't Starve Together, resolved.} A crash-safe container transfer loads on the dedicated server and passes all in-game assertions.}\label{fig:case-dst-ok}
\caseback
\end{minipage}\par\vspace{\caseafterskip}
\par\noindent\begin{minipage}{\linewidth}
\casehead{\casepill{ospeach}{Error case}}{Don't Starve Together\,$\cdot$\,\texttt{container-\allowbreak{}transfer-\allowbreak{}transaction}\,$\cdot$\,Kimi K3.1 Pro}
\caserow{case-dst-fail}{%
\casefield{task\_id}{\texttt{container-\allowbreak{}transfer-\allowbreak{}transaction}}
\casefield{Game}{Don't Starve Together}
\casefield{Type}{construction}
\casefield{Outcome}{Interface the ecosystem lacks}}{Task instruction (construction, excerpt)}{%
Create actions with AddAction, AddComponentAction, AddStategraphActionHandler for wilson and wilson\_client, and AddModRPCHandler.}
\begin{casecard}[colback=white,runin]{Agent's final report (verbatim)}
\itshape All required files have been inspected, and the Action/RPC wiring harness passes with extra client fields stripped. I'm removing the temporary harness and running the final syntax, line-count, workspace, and public diagnostic checks now.
\end{casecard}
\begin{casecard}[before upper={}]{Run Result}
{\sffamily\bfseries\casetitlefont Run Result}\hspace{1.5em}{\sffamily\bfseries Model:}~Kimi K3.1 Pro\hspace{1.5em}{\sffamily\bfseries Overall status:}~\casefailchip{UNRESOLVED}\par
{\sffamily\bfseries Verdict:}~unresolved; the mod fails to load on the dedicated server.\par
{\sffamily\bfseries Records (verbatim):}~\caselog{\ldots\\{}[\allowbreak{}00:\allowbreak{}00:\allowbreak{}00]\allowbreak{}:\allowbreak{} Mod:\allowbreak{} rhenium\_\allowbreak{}0aaa.\allowbreak{}.\allowbreak{}.\allowbreak{} (\allowbreak{}Crash-\allowbreak{}safe Container Transfer)\allowbreak{}	Loading modmain.\allowbreak{}lua\\{}[\allowbreak{}00:\allowbreak{}00:\allowbreak{}00]\allowbreak{}:\allowbreak{} MOD ERROR:\allowbreak{} rhenium\_\allowbreak{}0aaa.\allowbreak{}.\allowbreak{}.\allowbreak{} (\allowbreak{}Crash-\allowbreak{}safe Container Transfer)\allowbreak{}:\allowbreak{} Mod:\allowbreak{} rhenium\_\allowbreak{}0aaa.\allowbreak{}.\allowbreak{}.\allowbreak{} (\allowbreak{}Crash-\allowbreak{}safe Container Transfer)\allowbreak{}\\{}[\allowbreak{}00:\allowbreak{}00:\allowbreak{}00]\allowbreak{}:\allowbreak{} [\allowbreak{}string ".\allowbreak{}.\allowbreak{}/\allowbreak{}mods/\allowbreak{}rhenium\_\allowbreak{}0aaa.\allowbreak{}.\allowbreak{}.\allowbreak{}/\allowbreak{}modmain.\allowbreak{}lua"]\allowbreak{}:\allowbreak{}70:\allowbreak{} attempt to call global 'ActionHandler' (\allowbreak{}a nil value)\allowbreak{}\\{}LUA ERROR stack traceback:\allowbreak{}}
\end{casecard}
\begin{casecard}[colback=white,runin]{Analysis}
The agent's own wiring harness passes. In the game, \texttt{modmain.lua} runs in a sandbox that does not expose most engine globals; \texttt{ActionHandler} is reachable only through \texttt{GLOBAL}. The call on line 70 therefore fails while the mod loads, and no in-game test runs. A harness that defines \texttt{ActionHandler} globally, as the agent's did, cannot reveal this error.
\end{casecard}
\casecapgap\captionof{figure}{\textbf{Don't Starve Together, unresolved (interface the ecosystem lacks).} The same task as Figure~\ref{fig:case-dst-ok}: the code calls an engine global that the mod environment does not expose.}\label{fig:case-dst-fail}
\caseback
\end{minipage}\par\vspace{\caseafterskip}
\clearpage
\subsection{Terraria}
\label{app:case-terraria}
\par\noindent\begin{minipage}{\linewidth}
\casehead{\casepill{osblue}{Resolved}}{Terraria\,$\cdot$\,\texttt{terraria-\allowbreak{}overhaul-\allowbreak{}entity-\allowbreak{}chunk-\allowbreak{}lifetime-\allowbreak{}hard}\,$\cdot$\,Claude Opus 5.5}
\caserow{case-terraria-ok}{%
\casefield{task\_id}{\texttt{terraria-\allowbreak{}overhaul-\allowbreak{}entity-\allowbreak{}chunk-\allowbreak{}lifetime-\allowbreak{}hard}}
\casefield{Game}{Terraria}
\casefield{Type}{repair}
\casefield{Outcome}{Resolved}}{Task instruction (repair, excerpt)}{%
Chunk creation is transactional across \texttt{OnChunkCreated}. Invoke every creation subscriber once and isolate individual subscriber exceptions. A throwing subscriber does not prevent later subscribers or invalidate an otherwise canonical creation. After notification, return \texttt{true} only if the same live, initialized entity is still the canonical mapping.}
\begin{casecard}[colback=white,runin]{Agent's final report (verbatim)}
\itshape Exceptions from chunk observers are only written to \texttt{Debug.\allowbreak{}WriteLine}. ... Following the system instructions, I did not run the benchmark scorer, so there's no run ID or score.
\end{casecard}
\begin{casecard}[before upper={}]{Run Result}
{\sffamily\bfseries\casetitlefont Run Result}\hspace{1.5em}{\sffamily\bfseries Model:}~Claude Opus 5.5\hspace{1.5em}{\sffamily\bfseries Overall status:}~\casepass{RESOLVED}\par
{\sffamily\bfseries Verdict:}~resolved; F2P 56/56 and P2P 25/25 (19 steps).\par
{\sffamily\bfseries Records (verbatim):}~\caselog{\ldots\\{}real\_\allowbreak{}engine\_\allowbreak{}active\_\allowbreak{}query\_\allowbreak{}snapshot:\allowbreak{} passed;\allowbreak{} real\_\allowbreak{}engine\_\allowbreak{}world\_\allowbreak{}coordinate\_\allowbreak{}entrypoint:\allowbreak{} passed;\allowbreak{} real\_\allowbreak{}engine\_\allowbreak{}creation\_\allowbreak{}observer\_\allowbreak{}transaction:\allowbreak{} passed}
\end{casecard}
\begin{casecard}[colback=white,runin]{Analysis}
The submission passes every test. One real-engine case subscribes a throwing observer and a second observer to \texttt{OnChunkCreated} and checks that the chunk is still created, that both observers run once, and that the chunk stays mapped.
\end{casecard}
\casecapgap\captionof{figure}{\textbf{Terraria, resolved.} A repair of entity-backed chunk lifetime passes every test.}\label{fig:case-terraria-ok}
\caseback
\end{minipage}\par\vspace{\caseafterskip}
\par\noindent\begin{minipage}{\linewidth}
\casehead{\casepill{ospeach}{Error case}}{Terraria\,$\cdot$\,\texttt{terraria-\allowbreak{}overhaul-\allowbreak{}entity-\allowbreak{}chunk-\allowbreak{}lifetime-\allowbreak{}hard}\,$\cdot$\,GPT-6 Astra}
\caserow{case-terraria-fail}{%
\casefield{task\_id}{\texttt{terraria-\allowbreak{}overhaul-\allowbreak{}entity-\allowbreak{}chunk-\allowbreak{}lifetime-\allowbreak{}hard}}
\casefield{Game}{Terraria}
\casefield{Type}{repair}
\casefield{Outcome}{Behavior or lifecycle in game}}{Task instruction (repair, excerpt)}{%
A throwing subscriber does not prevent later subscribers or invalidate an otherwise canonical creation. After notification, return \texttt{true} only if the same live, initialized entity is still the canonical mapping. If an observer unloads the system or destroys/replaces that entity, return \texttt{false} with an invalid output;}
\begin{casecard}[colback=white,runin]{Agent's final report (verbatim)}
\itshape One final edge case: teardown must remove a corrupt mapping by entity identity before notifying observers. I'm adding that guard.
\end{casecard}
\begin{casecard}[before upper={}]{Run Result}
{\sffamily\bfseries\casetitlefont Run Result}\hspace{1.5em}{\sffamily\bfseries Model:}~GPT-6 Astra\hspace{1.5em}{\sffamily\bfseries Overall status:}~\casefailchip{UNRESOLVED}\par
{\sffamily\bfseries Verdict:}~unresolved; F2P 55/56 and P2P 25/25. Score 70/100.\par
{\sffamily\bfseries Records (verbatim):}~\caselog{\ldots\\{}\{"name":\allowbreak{} "real\_\allowbreak{}engine\_\allowbreak{}creation\_\allowbreak{}observer\_\allowbreak{}transaction",\allowbreak{} "passed":\allowbreak{} false\}}
\end{casecard}
\begin{casecard}[colback=white,runin]{Analysis}
Only the real-engine case that runs chunk creation with a throwing observer inside tModLoader fails. That case checks that creation is accepted, that both observers run once, and that the chunk remains the canonical mapping. The agent's final messages describe further edge cases it guarded in source review, without any execution.
\end{casecard}
\casecapgap\captionof{figure}{\textbf{Terraria, unresolved (behavior in the game).} The same task as Figure~\ref{fig:case-terraria-ok}: chunk creation with a throwing observer fails in the real engine.}\label{fig:case-terraria-fail}
\caseback
\end{minipage}\par\vspace{\caseafterskip}
\clearpage
\subsection{Palworld}
\label{app:case-palworld}
\par\noindent\begin{minipage}{\linewidth}
\casehead{\casepill{osblue}{Resolved}}{Palworld\,$\cdot$\,\texttt{guild-\allowbreak{}storage-\allowbreak{}ledger}\,$\cdot$\,Claude Opus 5.5}
\caserow{case-palworld-ok}{%
\casefield{task\_id}{\texttt{guild-\allowbreak{}storage-\allowbreak{}ledger}}
\casefield{Game}{Palworld}
\casefield{Type}{construction}
\casefield{Outcome}{Resolved}}{Task instruction (construction, excerpt)}{%
Build a UE4SS Lua mod from an empty workspace. Implement a plain-Lua ledger module and a dedicated-server service module. ... Package the mod as Info.json with an InstallRule entry of "Type": "Lua", "IsServer": true and a scripts/main.lua entry point loaded by UE4SS.}
\begin{casecard}[colback=white,runin]{Agent's final report}
The agent's last message before submitting describes a smoke test: ``Let me check global hygiene and simulate main.lua with stubbed UE4SS functions (only as a local smoke test, not as real-engine evidence).''
\end{casecard}
\begin{casecard}[before upper={}]{Run Result}
{\sffamily\bfseries\casetitlefont Run Result}\hspace{1.5em}{\sffamily\bfseries Model:}~Claude Opus 5.5\hspace{1.5em}{\sffamily\bfseries Overall status:}~\casepass{RESOLVED}\par
{\sffamily\bfseries Verdict:}~resolved; F2P 15/15 (14 steps).\par
{\sffamily\bfseries Records (verbatim):}~\caselog{Info.\allowbreak{}json:\allowbreak{} \{"ModName":\allowbreak{} "Rhenium Guild Storage Ledger",\allowbreak{} "PackageName":\allowbreak{} "RheniumGuildStorageLedger",\allowbreak{} .\allowbreak{}.\allowbreak{}.\allowbreak{},\allowbreak{} "InstallRule":\allowbreak{} [\allowbreak{}\{"Type":\allowbreak{} "Lua",\allowbreak{} "IsServer":\allowbreak{} true,\allowbreak{} "Targets":\allowbreak{} [\allowbreak{}".\allowbreak{}/\allowbreak{}scripts/\allowbreak{}"]\allowbreak{}\}]\allowbreak{}\}\\{}supply-\allowbreak{}manifest:\allowbreak{} All integration contracts are satisfied\\{}[\allowbreak{}Lua]\allowbreak{} @@RHENIUM\_\allowbreak{}SERVER@@6cd0.\allowbreak{}.\allowbreak{}.\allowbreak{}:\allowbreak{}\{"status":\allowbreak{}"passed",\allowbreak{}"cases":\allowbreak{}[\allowbreak{}\{"name":\allowbreak{}"dedicated server context is available",\allowbreak{}"passed":\allowbreak{}true,\allowbreak{}.\allowbreak{}.\allowbreak{}.\allowbreak{}\},\allowbreak{}\{"name":\allowbreak{}"guild storage service is published",\allowbreak{}"passed":\allowbreak{}true,\allowbreak{}.\allowbreak{}.\allowbreak{}.\allowbreak{}\\{}headless-\allowbreak{}server:\allowbreak{} All 4 headless server assertions passed}
\end{casecard}
\begin{casecard}[colback=white,runin]{Analysis}
Claude Opus 5.5 is one of two models that resolve any Palworld task. Its manifest uses the keys that UE4SS expects for a server-side Lua mod, the dedicated server loads the scripts, and the in-game service is published and exercised. The agent explicitly distinguished its stubbed smoke test from real-engine evidence.
\end{casecard}
\casecapgap\captionof{figure}{\textbf{Palworld, resolved.} A server-authoritative guild storage ledger loads as a UE4SS mod on the dedicated server.}\label{fig:case-palworld-ok}
\caseback
\end{minipage}\par\vspace{\caseafterskip}
\par\noindent\begin{minipage}{\linewidth}
\casehead{\casepill{ospeach}{Error case}}{Palworld\,$\cdot$\,\texttt{guild-\allowbreak{}storage-\allowbreak{}ledger}\,$\cdot$\,GPT-6 Astra}
\caserow{case-palworld-fail}{%
\casefield{task\_id}{\texttt{guild-\allowbreak{}storage-\allowbreak{}ledger}}
\casefield{Game}{Palworld}
\casefield{Type}{construction}
\casefield{Outcome}{Host convention}}{Task instruction (construction, excerpt)}{%
Package the mod as Info.json with an InstallRule entry of "Type": "Lua", "IsServer": true and a scripts/main.lua entry point loaded by UE4SS.}
\begin{casecard}[colback=white,runin]{Agent's final report}
None in prose: the last messages contain only the headings ``Running mandated final public diagnostics after all edits'' and ``Preparing final submission command as exact last action''.
\end{casecard}
\begin{casecard}[before upper={}]{Run Result}
{\sffamily\bfseries\casetitlefont Run Result}\hspace{1.5em}{\sffamily\bfseries Model:}~GPT-6 Astra\hspace{1.5em}{\sffamily\bfseries Overall status:}~\casefailchip{UNRESOLVED}\par
{\sffamily\bfseries Verdict:}~unresolved; rejected by the manifest contract before the mod reaches the dedicated server.\par
{\sffamily\bfseries Records (verbatim):}~\caselog{Info.\allowbreak{}json:\allowbreak{} \{"Name":\allowbreak{} "Rhenium Guild Storage Ledger",\allowbreak{} "Version":\allowbreak{} "1.\allowbreak{}0.\allowbreak{}0",\allowbreak{} "Description":\allowbreak{} "Server-\allowbreak{}authoritative guild supply ledger",\allowbreak{} "InstallRules":\allowbreak{} [\allowbreak{}\{"Type":\allowbreak{} "Lua",\allowbreak{} "IsServer":\allowbreak{} true,\allowbreak{} "Path":\allowbreak{} "scripts/\allowbreak{}main.\allowbreak{}lua"\}]\allowbreak{}\}\\{}\ldots\\{}supply-\allowbreak{}manifest:\allowbreak{} Integration contract patterns are incomplete or forbidden patterns matched}
\end{casecard}
\begin{casecard}[colback=white,runin]{Analysis}
The manifest follows a schema of the agent's own making: \texttt{InstallRules} instead of the \texttt{InstallRule} key named in the task, \texttt{Path} instead of \texttt{Targets}, and no \texttt{ModName} or \texttt{PackageName}. The manifest contract rejects it before any dedicated-server case runs, so the correct ledger never reaches the game.
\end{casecard}
\casecapgap\captionof{figure}{\textbf{Palworld, unresolved (host convention).} The same task as Figure~\ref{fig:case-palworld-ok}: correct logic packaged with an invented manifest schema.}\label{fig:case-palworld-fail}
\caseback
\end{minipage}\par\vspace{\caseafterskip}
\clearpage
\subsection{Left 4 Dead 2}
\label{app:case-l4d2}
\par\noindent\begin{minipage}{\linewidth}
\casehead{\casepill{osblue}{Resolved}}{Left 4 Dead 2\,$\cdot$\,\texttt{l4d2-\allowbreak{}competitive-\allowbreak{}release-\allowbreak{}regressions-\allowbreak{}easy}\,$\cdot$\,GPT-6 Astra}
\caserow{case-l4d2-ok}{%
\casefield{task\_id}{\texttt{l4d2-\allowbreak{}competitive-\allowbreak{}release-\allowbreak{}regressions-\allowbreak{}easy}}
\casefield{Game}{Left 4 Dead 2}
\casefield{Type}{repair}
\casefield{Outcome}{Resolved}}{Task instruction (repair, excerpt)}{%
[\#CR-126] Partial map names during forced changelevel. SourceMod accepts a unique partial map name such as \texttt{c1m2}, but the forced-changelevel hook sends the unresolved text to the engine and the transition stalls. Resolve a valid unique partial name to its full installed-map name before the engine processes the change (for example \texttt{c1m2} to \texttt{c1m2\_\allowbreak{}streets}) ...}
\begin{casecard}[colback=white,runin]{Agent's final report (verbatim)}
\itshape The source review is complete, and the package check passed with no forbidden artifacts. Compilation remains unverified because the pinned compiler is missing.
\end{casecard}
\begin{casecard}[before upper={}]{Run Result}
{\sffamily\bfseries\casetitlefont Run Result}\hspace{1.5em}{\sffamily\bfseries Model:}~GPT-6 Astra\hspace{1.5em}{\sffamily\bfseries Overall status:}~\casepass{RESOLVED}\par
{\sffamily\bfseries Verdict:}~resolved; F2P 7/7 and P2P 10/10 on the dedicated server (8 steps).\par
{\sffamily\bfseries Records (verbatim):}~\caselog{submission-\allowbreak{}shape:\allowbreak{} submission contains the required source-\allowbreak{}only project tree\\{}headless-\allowbreak{}server:\allowbreak{} all 20 real-\allowbreak{}server assertions passed}
\end{casecard}
\begin{casecard}[colback=white,runin]{Analysis}
The three plugins are compiled with the pinned SourcePawn compiler and loaded into a Left 4 Dead 2 dedicated server, where the fixtures trigger forced map changes and the other reported regressions. The agent could not compile and said so; the compiled plugins nevertheless behave as required.
\end{casecard}
\casecapgap\captionof{figure}{\textbf{Left 4 Dead 2, resolved.} Three regressions in competitive server plugins are fixed and verified on the dedicated server.}\label{fig:case-l4d2-ok}
\caseback
\end{minipage}\par\vspace{\caseafterskip}
\par\noindent\begin{minipage}{\linewidth}
\casehead{\casepill{ospeach}{Error case}}{Left 4 Dead 2\,$\cdot$\,\texttt{l4d2-\allowbreak{}competitive-\allowbreak{}release-\allowbreak{}regressions-\allowbreak{}easy}\,$\cdot$\,Kimi K3.1 Pro}
\caserow{case-l4d2-fail}{%
\casefield{task\_id}{\texttt{l4d2-\allowbreak{}competitive-\allowbreak{}release-\allowbreak{}regressions-\allowbreak{}easy}}
\casefield{Game}{Left 4 Dead 2}
\casefield{Type}{repair}
\casefield{Outcome}{Interface the ecosystem lacks}}{Task instruction (repair, excerpt)}{%
Resolve a valid unique partial name to its full installed-map name before the engine processes the change (for example \texttt{c1m2} to \texttt{c1m2\_\allowbreak{}streets}) and use that same full name for the existing Director transition work. Already complete names must keep working.}
\begin{casecard}[colback=white,runin]{Agent's final report}
None: the agent submitted after 18 steps, and its last messages contain only commands.
\end{casecard}
\begin{casecard}[before upper={}]{Run Result}
{\sffamily\bfseries\casetitlefont Run Result}\hspace{1.5em}{\sffamily\bfseries Model:}~Kimi K3.1 Pro\hspace{1.5em}{\sffamily\bfseries Overall status:}~\casefailchip{UNRESOLVED}\par
{\sffamily\bfseries Verdict:}~unresolved; the plugin does not compile, so all 17 tests are unobserved.\par
{\sffamily\bfseries Records (verbatim):}~\caselog{SUBMISSION\_\allowbreak{}FAILED:\allowbreak{} SourcePawn compile failed for l4d2\_\allowbreak{}fix\_\allowbreak{}changelevel.\allowbreak{}sp:\allowbreak{}\\{}/\allowbreak{}app/\allowbreak{}l4d2-\allowbreak{}competitive/\allowbreak{}addons/\allowbreak{}sourcemod/\allowbreak{}scripting/\allowbreak{}l4d2\_\allowbreak{}fix\_\allowbreak{}changelevel.\allowbreak{}sp(\allowbreak{}167)\allowbreak{} :\allowbreak{} error 017:\allowbreak{} undefined symbol "MRES\_\allowbreak{}Changed"\\{}headless-\allowbreak{}server:\allowbreak{} real-\allowbreak{}server assertions did not complete with the frozen case set}
\end{casecard}
\begin{casecard}[colback=white,runin]{Analysis}
The fix for the changelevel hook returns \texttt{MRES\_Changed}, a hook result that the DHooks include does not declare; its result values have other names, such as \texttt{MRES\_ChangedHandled} and \texttt{MRES\_ChangedOverride}. SourcePawn rejects the undefined symbol at compile time, so none of the three plugins is loaded.
\end{casecard}
\casecapgap\captionof{figure}{\textbf{Left 4 Dead 2, unresolved (interface the ecosystem lacks).} The same task as Figure~\ref{fig:case-l4d2-ok}: the plugin returns a hook result that does not exist.}\label{fig:case-l4d2-fail}
\caseback
\end{minipage}\par\vspace{\caseafterskip}
\clearpage
\subsection{Slay the Spire}
\label{app:case-sts}
\par\noindent\begin{minipage}{\linewidth}
\casehead{\casepill{osblue}{Resolved}}{Slay the Spire\,$\cdot$\,\texttt{stslib-\allowbreak{}block-\allowbreak{}ledger-\allowbreak{}r1}\,$\cdot$\,Claude Opus 5.5}
\caserow{case-sts-ok}{%
\casefield{task\_id}{\texttt{stslib-\allowbreak{}block-\allowbreak{}ledger-\allowbreak{}r1}}
\casefield{Game}{Slay the Spire}
\casefield{Type}{repair}
\casefield{Outcome}{Resolved}}{Task instruction (repair, excerpt)}{%
Players report that custom shields can add up to more block than the creature actually has. ... Block created by a removal callback is a subsequent gain and must not be consumed retroactively by the triggering loss. ... Preserve public APIs, ordinary partial/complete losses, attack damage and existing removal-based damage reduction.}
\begin{casecard}[colback=white,runin]{Agent's final report (verbatim)}
\itshape I've finished the review. Submitting now.
\end{casecard}
\begin{casecard}[before upper={}]{Run Result}
{\sffamily\bfseries\casetitlefont Run Result}\hspace{1.5em}{\sffamily\bfseries Model:}~Claude Opus 5.5\hspace{1.5em}{\sffamily\bfseries Overall status:}~\casepass{RESOLVED}\par
{\sffamily\bfseries Verdict:}~resolved; F2P 20/20 and P2P 6/6 in the running game (22 steps).\par
{\sffamily\bfseries Records (verbatim):}~\caselog{native-\allowbreak{}engine:\allowbreak{} all 27 native engine cases passed\\{}BLOCK\_\allowbreak{}SETTLEMENT\_\allowbreak{}SCOUT 1b29.\allowbreak{}.\allowbreak{}.\allowbreak{}\textbar{}\allowbreak{}case,\allowbreak{}id=\allowbreak{}monster-\allowbreak{}attenuate,\allowbreak{}error=\allowbreak{}none,\allowbreak{}block=\allowbreak{}9,\allowbreak{}stacks=\allowbreak{}C:\allowbreak{}9;\allowbreak{},\allowbreak{}otherBlock=\allowbreak{}10,\allowbreak{}otherStacks=\allowbreak{}B:\allowbreak{}10;\allowbreak{},\allowbreak{}hpLost=\allowbreak{}0,\allowbreak{}events=\allowbreak{}A/\allowbreak{}13;\allowbreak{}B/\allowbreak{}0;\allowbreak{}}
\end{casecard}
\begin{casecard}[colback=white,runin]{Analysis}
ModTheSpire loads the rebuilt StSLib into the real game. On the game's first update, a probe mod creates a player and a monster, gives them layered custom block, applies each loss scenario, and prints one line per case; all 26 cases run within one game tick, before any frame of combat is drawn. In \texttt{monster-attenuate}, an 18-damage attack exhausts a layer whose removal callback reduces the remaining damage by 6, and a 9-point layer \texttt{C} appears during the hit; the required result keeps it (\texttt{block=9, stacks=C:9}). Claude Opus 5.5 never compiled its change and closed with a one-line report.
\end{casecard}
\casecapgap\captionof{figure}{\textbf{Slay the Spire, resolved.} A repair of StSLib's custom-block accounting passes all probe cases in the running game.}\label{fig:case-sts-ok}
\caseback
\end{minipage}\par\vspace{\caseafterskip}
\par\noindent\begin{minipage}{\linewidth}
\casehead{\casepill{ospeach}{Error case}}{Slay the Spire\,$\cdot$\,\texttt{stslib-\allowbreak{}block-\allowbreak{}ledger-\allowbreak{}r1}\,$\cdot$\,DeepSeek-V4.1-Flash}
\caserow{case-sts-fail}{%
\casefield{task\_id}{\texttt{stslib-\allowbreak{}block-\allowbreak{}ledger-\allowbreak{}r1}}
\casefield{Game}{Slay the Spire}
\casefield{Type}{repair}
\casefield{Outcome}{Breaks existing behavior}}{Task instruction (repair, excerpt)}{%
Block created by a removal callback is a subsequent gain and must not be consumed retroactively by the triggering loss. ... Preserve public APIs, ordinary partial/complete losses, attack damage and existing removal-based damage reduction.}
\begin{casecard}[colback=white,runin]{Agent's final report (verbatim)}
\itshape ... attack damage and removal-based damage reduction are unchanged, and the duplicated monster start-of-turn block-loss path was folded into the single reentrancy-safe implementation. I validated the algorithms with a faithful model of the new accounting ... all scenarios pass. (This environment has no JDK/Maven/StS jars, so the offline Maven build itself could not be executed here; all sources are Java 8-compatible and syntax-balanced.)
\end{casecard}
\begin{casecard}[before upper={}]{Run Result}
{\sffamily\bfseries\casetitlefont Run Result}\hspace{1.5em}{\sffamily\bfseries Model:}~DeepSeek-V4.1-Flash\hspace{1.5em}{\sffamily\bfseries Overall status:}~\casefailchip{UNRESOLVED}\par
{\sffamily\bfseries Verdict:}~unresolved; F2P 20/20 but P2P 4/6 (57 steps).\par
{\sffamily\bfseries Records (verbatim):}~\caselog{native-\allowbreak{}engine:\allowbreak{} 25/\allowbreak{}27 native engine cases passed\\{}BLOCK\_\allowbreak{}SETTLEMENT\_\allowbreak{}SCOUT 9cc1.\allowbreak{}.\allowbreak{}.\allowbreak{}\textbar{}\allowbreak{}case,\allowbreak{}id=\allowbreak{}monster-\allowbreak{}attenuate,\allowbreak{}error=\allowbreak{}none,\allowbreak{}block=\allowbreak{}0,\allowbreak{}stacks=\allowbreak{},\allowbreak{}otherBlock=\allowbreak{}10,\allowbreak{}otherStacks=\allowbreak{}B:\allowbreak{}10;\allowbreak{},\allowbreak{}hpLost=\allowbreak{}0,\allowbreak{}events=\allowbreak{}A/\allowbreak{}13;\allowbreak{}B/\allowbreak{}0;\allowbreak{}C/\allowbreak{}0;\allowbreak{}\\{}failed:\allowbreak{} case:\allowbreak{}player-\allowbreak{}attenuate (\allowbreak{}preserved)\allowbreak{},\allowbreak{} case:\allowbreak{}monster-\allowbreak{}attenuate (\allowbreak{}preserved)\allowbreak{}}
\end{casecard}
\begin{casecard}[colback=white,runin]{Analysis}
The rewrite fixes every reported symptom (20/20 F2P) but debits the 9-point layer \texttt{C} that appears during the attenuated hit, so the creature ends with no block (event \texttt{C/0}). The task requires such block to survive and requires the existing damage reduction to be preserved; the unmodified code passes both \texttt{attenuate} cases, which are P2P tests, and they are the only tests that detect the regression. The final report asserts the very property that breaks, on the basis of the agent's own model of the accounting, since no JDK was available.
\end{casecard}
\casecapgap\captionof{figure}{\textbf{Slay the Spire, unresolved (breaks existing behavior).} The same task as Figure~\ref{fig:case-sts-ok}: every F2P test passes, but two P2P tests show that existing damage reduction no longer works.}\label{fig:case-sts-fail}
\caseback
\end{minipage}\par\vspace{\caseafterskip}
\clearpage
\subsection{RimWorld}
\label{app:case-rimworld}
\par\noindent\begin{minipage}{\linewidth}
\casehead{\casepill{osblue}{Resolved}}{RimWorld\,$\cdot$\,\texttt{adaptive-\allowbreak{}storage-\allowbreak{}slot-\allowbreak{}limit-\allowbreak{}restore}\,$\cdot$\,Claude Opus 5.5}
\caserow{case-rimworld-ok}{%
\casefield{task\_id}{\texttt{adaptive-\allowbreak{}storage-\allowbreak{}slot-\allowbreak{}limit-\allowbreak{}restore}}
\casefield{Game}{RimWorld}
\casefield{Type}{repair}
\casefield{Outcome}{Resolved}}{Task instruction (repair, excerpt)}{%
Capture the saved logical slot limit during post-initialization and restore that exact backing value through the real CurrentSlotLimit property setter. ... Preserve zero, one-slot, ordinary finite, near-int.MaxValue, multi-cell, already-unlimited, default and repeated-initialization behavior.}
\begin{casecard}[colback=white,runin]{Agent's final report}
None: the agent submitted after nine steps, and its last messages contain only commands.
\end{casecard}
\begin{casecard}[before upper={}]{Run Result}
{\sffamily\bfseries\casetitlefont Run Result}\hspace{1.5em}{\sffamily\bfseries Model:}~Claude Opus 5.5\hspace{1.5em}{\sffamily\bfseries Overall status:}~\casepass{RESOLVED}\par
{\sffamily\bfseries Verdict:}~resolved; F2P 2/2 and P2P 11/11 in the running game (9 steps).\par
{\sffamily\bfseries Records (verbatim):}~\caselog{real-\allowbreak{}game:\allowbreak{} all required real-\allowbreak{}game assertions passed}
\end{casecard}
\begin{casecard}[colback=white,runin]{Analysis}
The harness compiles the one changed file against the pinned RimWorld~1.6 assemblies, starts the real game with Harmony and the submission loaded, and runs the assertions inside the game process once start-up completes. The two F2P cases check that a saved unlimited limit is restored and that the setter is not skipped; the eleven P2P cases cover the zero, one-slot, finite, near-unlimited, multi-cell, unlimited, default, and repeated-initialization limits, the per-cell recomputation, and the initialization bookkeeping and notifications.
\end{casecard}
\casecapgap\captionof{figure}{\textbf{RimWorld, resolved.} A restored storage slot limit survives saving and loading in the real game.}\label{fig:case-rimworld-ok}
\caseback
\end{minipage}\par\vspace{\caseafterskip}
\par\noindent\begin{minipage}{\linewidth}
\casehead{\casepill{ospeach}{Error case}}{RimWorld\,$\cdot$\,\texttt{playwright-\allowbreak{}nonbulk-\allowbreak{}trade-\allowbreak{}accounting}\,$\cdot$\,GPT-6 Astra}
\caserow{case-rimworld-fail}{%
\casefield{task\_id}{\texttt{playwright-\allowbreak{}nonbulk-\allowbreak{}trade-\allowbreak{}accounting}}
\casefield{Game}{RimWorld}
\casefield{Type}{repair}
\casefield{Outcome}{Interface the ecosystem lacks}}{Task instruction (repair, excerpt)}{%
The supplied source is the unmodified upstream base for a bounded RimWorld 1.6 production regression in Playwright. Repair the behavior described below in \texttt{Playwright/\allowbreak{}Source}.}
\begin{casecard}[colback=white,runin]{Agent's final report (verbatim)}
\itshape The patch now uses captured sale data and guards null traders, empty state, and non-positive counts. Source checks passed; compilation and runtime validation remain unverified because this workspace has no C\# compiler or game references.
\end{casecard}
\begin{casecard}[before upper={}]{Run Result}
{\sffamily\bfseries\casetitlefont Run Result}\hspace{1.5em}{\sffamily\bfseries Model:}~GPT-6 Astra\hspace{1.5em}{\sffamily\bfseries Overall status:}~\casefailchip{UNRESOLVED}\par
{\sffamily\bfseries Verdict:}~unresolved; the build fails, so all 12 tests are unobserved (6 steps).\par
{\sffamily\bfseries Records (verbatim):}~\caselog{submission TradeablePatches.\allowbreak{}cs:\allowbreak{}23:\allowbreak{}   (\allowbreak{}\_\allowbreak{}\_\allowbreak{}instance.\allowbreak{}AnyColonyThing =\allowbreak{}=\allowbreak{} null \&\& \_\allowbreak{}\_\allowbreak{}instance.\allowbreak{}AnyTraderThing =\allowbreak{}=\allowbreak{} null)\allowbreak{})\allowbreak{}\\{}RimWorld build failed;\allowbreak{} private evidence was recorded\\{}real-\allowbreak{}game:\allowbreak{} real-\allowbreak{}game assertions did not start\\{}adjudication:\allowbreak{} ".\allowbreak{}.\allowbreak{}.\allowbreak{} The private build log lists C\# compile errors only in the submitted sources under /\allowbreak{}submission/\allowbreak{}Playwright (\allowbreak{}CS1061:\allowbreak{} 'Tradeable' has no definition for 'AnyColonyThing' /\allowbreak{} 'AnyTraderThing')\allowbreak{};\allowbreak{} no candidate assembly was produced .\allowbreak{}.\allowbreak{}.\allowbreak{}"}
\end{casecard}
\begin{casecard}[colback=white,runin]{Analysis}
The patch reads \texttt{AnyColonyThing} and \texttt{AnyTraderThing} from RimWorld's \texttt{Tradeable}, which defines neither member in the pinned RimWorld~1.6 API, so the C\# compiler rejects the submission and the game never starts. The task is a reverted upstream fix (H in Table~\ref{tab:task-level-2}): its initial code is the unmodified upstream commit immediately before the fix, which the task description calls the upstream base. The agent's report acknowledges that it could not compile, yet describes the patch as checked. Three further RimWorld submissions of the same model fail to compile for the same kind of reason (Section~\ref{sec:taxonomy}).
\end{casecard}
\casecapgap\captionof{figure}{\textbf{RimWorld, unresolved (interface the ecosystem lacks).} A trade-accounting repair reads two members that RimWorld's \texttt{Tradeable} does not define.}\label{fig:case-rimworld-fail}
\caseback
\end{minipage}\par\vspace{\caseafterskip}
\FloatBarrier

\end{document}